\documentclass[10pt,twocolumn]{article}
\usepackage{times}
\usepackage{epsfig}
\usepackage{url}

\newcommand{\Section}[1]{\vspace{-8pt}\section{\hskip -1em.~~#1}\vspace{-3pt}} 
\newcommand{\SubSection}[1]{\vspace{-3pt}\subsection{\hskip -1em.~~#1}
     	\vspace{-3pt}}

\begin{document}

\date{}

\title{\Large\bf Better Foreground Segmentation Through Graph Cuts}


\author{\begin{tabular}[t]{c@{\extracolsep{8em}}c} 
Nicholas R. Howe  & Alexandra Deschamps \\
 \\
\multicolumn{2}{c}{Computer Science} \\
\multicolumn{2}{c}{Smith College} \\
\multicolumn{2}{c}{Northampton, MA 01063}
\end{tabular}}

\maketitle

\section*{\centering Abstract}

{\em For many tracking and surveillance applications, background
subtraction provides an effective means of segmenting objects moving
in front of a static background.  Researchers have traditionally used
combinations of morphological operations to remove the noise inherent
in the background-subtracted result.  Such techniques can effectively
isolate foreground objects, but tend to lose fidelity around the
borders of the segmentation, especially for noisy input.  This paper
explores the use of a minimum graph cut algorithm to segment the
foreground, resulting in qualitatively and quantitiatively cleaner
segmentations.  Experiments on both artificial and real data show that
the graph-based method reduces the error around segmented foreground
objects.  A MATLAB code implementation is available at
\url{http://www.cs.smith.edu/~nhowe/research/code/#fgseg}.
}

\Section{Introduction}

Many computer vision applications require the segmentation of
foreground from background as a prelude to further processing.
Although difficult in the general case, the task can be greatly
simplified if the object or objects of interest in the foreground move
across a static background.  Such situations arise or can be
engineered in a wide variety of applications, including security
videos, video-based tracking and motion capture, sports ergonomics,
and human-computer interactions via inexpensive workstation-mounted
cameras.

All of these applications rely on or would benefit from high-quality
foreground segmentation.  Unfortunately, existing methods sometimes
prove unreliable and error-prone.  Furthermore, the results can vary
greatly between successive frames of a video.  This paper introduces a
new way to compute the foreground segmentation that makes fewer errors
and can be temporally stabilized from frame to frame.

Traditionally, researchers solve the foreground segmentation problem
with static background through a procedure called {\em background
subtraction}.  In this approach, the computer builds a model of the
static background, either off-line or updated dynamically after each
frame in the video stream, and then compares the next frame with the
background model on a per-pixel basis.  Pixels that differ
sufficiently from the background model may deemed part of the
foreground, at least in the ideal case.  If the calculation were free
of noise, one could simply fix a threshold and declare anything
sufficiently different from the background to be part of the
foreground.

Unfortunately, a number of confounding factors make perfect background
subtraction unattainable.  Camera noise (particularly in inexpensive
CCD cameras) ensures that even background pixels will not exhibit
constant values from frame to frame, but will instead show a
distribution around some characteristic value.  If the background
contains non-static elements (such as vegetation, cloth, or gravel)
then the variance in the measurements of background pixels may be
quite large.  Moving objects in the foreground may cause shadows and
reflections to fall on background areas, changing their appearance
significantly.  The foreground objects may lack sufficient contrast
with the background areas they obscure, either through deliberate
camouflage or by chance.  In consequence, comparison of a pixel in a
given frame with the background model for that pixel cannot
definitively classify the pixel as either foreground or background
without some potential for error.

Errors at a single pixel may be mitigated by aggregating the results
over some local neighborhood of pixels.  Researchers have
traditionally taken this approach to cleaning up the errors in the
thresholded image.  A combination of morphological operations on the
binary thresholded image removes isolated foreground and background
pixels, generating a better approximation to the silhouettes of the
moving objects in the foreground.  Unfortunately, the same approach
can obliterate details at the edges of the silhouette.

This work departs from the standard practice by using an algorithm
based upon the minimum graph cut to separate the foreground from the
background.  The algorithm presented herein uses information that
would be thrown away by thresholding to construct a graph
incorporating all the differences measured between the current frame
and the background model.  Links in this graph reflect the
connectivity of the pixels in the image, allowing each pixel to affect
those in its local neighborhood.  (Details of the graph construction
appear below.)  Segmenting the graph using a standard graph-cut
algorithm produces a foreground-background segmentation that can
correct local errors without introducing larger global distortions.
Qualitatively, the results using the new technique look cleaner and
more correct; the quantitative tests in Section~\ref{result-sect} show
that the method produces fewer errors than do current practices when
compared to human-segmented ground truth.

The remainder of this paper conducts an in-depth look at the old and
new methods for foreground segmentation using background subtraction.
Section~\ref{alg-sect} describes the two algorithms that
Sections~\ref{synth-sect} and \ref{video-sect} compare experimentally.
Section~\ref{conc-sect} concludes with a discussion of the numerous
ties between this work and other efforts, and some final thoughts.

\Section{Algorithmic Details}
\label{alg-sect}

Morphological operations form the basis of the standard approach for
cleaning up noise after background subtraction and thresholding.  In
particular, the morphological techniques commonly applied consist of
the two basic operations {\em dilation} and {\em erosion} applied in
various combinations.  Dilation expands the foreground of the image,
adding a pixel to the foreground if any of its neighbors within a
specified neighborhood of radius $r$ (called the structuring element)
are already part of the foreground.  Erosion expands the background,
removing a pixel from the foreground if any of its neighbors are
background.  These two operations may be combined; a dilation followed
by an identical erosion is called a {\em closing}, and fills in holes
in the foreground smaller than the neighborhood diameter.  Likewise,
an erosion followed by an identical dilation is called an {\em
opening}, and may be used to eliminate isolated foreground pixels.
Such operations are well studied, and more details may be found in
reference texts \cite{matlab:ipt}.

Noise in the background-subtracted image tends to make some foreground
pixels look like background, and vice versa.  A morphological closing
followed by an opening addresses these sources of error: the closing
fills in the the missing foreground pixels (assuming that enough of
their neighbors are correctly identified), and the opening removes
extraneous foreground pixels surrounded by background.  Care must be
taken in choosing the radius for these operations.  If the radius is
too small, then larger clusters of noisy pixels will remain
uncorrected; if too large, then legitimate detail in the foreground
silhouette will be lost.

With particularly noisy background-subtracted images, mislabeled
background pixels may become so numerous and closely spaced that the
initial closing operation fills in the gaps between them.  Increasing
the threshold $\tau$ for the initial foreground-background
segmentation prevents this undesirable effect by biasing the initial
labeling away from the foreground.  In other words, the higher
threshold causes more foreground pixels to be classified as background
than background pixels as foreground.  Performing the initial closing
operation corrects for this bias by closing the gaps between the
correctly labeled foreground pixels.

\SubSection{Graph Cuts for Foreground Segmentation}

Unlike the morphological approach, the graph-cut algorithm begins by
building a graph based upon the image.  Each pixel $p_{ij}$ in the
image generates a corresponding graph vertex $v_{ij}$.  Two additional
vertices form the source and sink, representing the foreground and the
background respectively.

\begin{figure}
\begin{center}
\epsfxsize=3in\epsfysize=1.94in\epsfbox{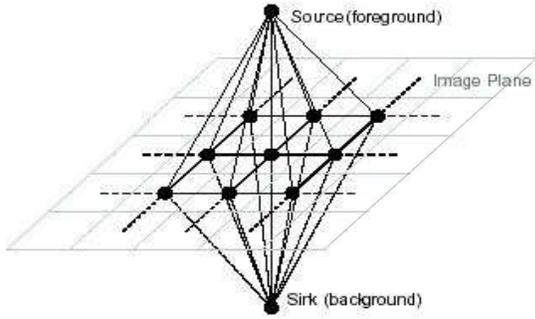}
\end{center}
\caption{Graph construct embedded in image plane.  Each pixel
corresponds to a node, and all pixel nodes are connected to the source
and the sink.}
\label{fig-graph}
\end{figure}

Figure~\ref{fig-graph} illustrates the graph formed for a small $3
\times 3$ portion of the image plane.  A typical vertex in the graph
links to exactly six other nodes: the source and the sink, plus the
the vertices of its four-connected neighbors.  Vertices corresponding
to pixels on the edge of the image will have fewer neighbor links, and
the source and the sink will each connect to all the pixel vertices.
The weights of the links between the pixel vertices and the source $s$
and sink $t$ derive directly from the difference between the current
frame and the background at the corresponding pixel, $\delta_{ij}$:
\begin{equation}
w(s,p_{ij}) = \delta_{ij}
\end{equation}
\begin{equation}
w(p_{ij},t) = 2\tau-\delta_{ij}
\end{equation}
The neighbor links (between pixel vertices) all have identical
weights, equal to $\tau$ times a second parameter $\alpha$ (typically
taking on values close to 1.0).  The parameter $\tau$ in the latter
equation plays an analogous role to the threshold in the morphological
algorithm, corresponding to the level above which the pixel associates
more strongly with the foreground than the background.

The value of $\alpha$ controls how strongly neighboring pixels tend to
group.  If $\alpha$ is low, then neighboring pixels bond weakly and
the end result will look much like that obtained by simply thesholding
the output from the background subtraction.  Conversely, high $\alpha$
causes pixels to bond strongly with neighboring pixels, and the output
will contain larger clusters of homogeneity.  Noisy inputs thus tend
to require larger values of $\alpha$, in order to smooth over the
larger clusters of noisy pixels.

Once constructed, standard methods based upon graph flow will find an
optimal (minimum cost) cut separating the source from the sink.
Andrew Goldberg has kindly made optimized code for this computation
available on the web \cite{goldberg:mincut}.  Each node in the graph
will lie on one side or the other of the optimal cut, remaining
connected solely to the source or to the sink.  The algorithm labels
those nodes still connected to the source as foreground, and those
connected to the sink as background.

If desired, one may construct a graph to represent several frames of
video at once.  In this case, a typical pixel vertex connects to six
neighbors (four spatial plus two temporal), but otherwise the
construction remains the same.  Using multiframe graphs can impose
consistency of the result from one frame to the next, but trial
experiments indicate no benefit in terms of overall accuracy.
Therefore, the remainder of this paper focuses on single-frame graphs.

\Section{Synthetic Results}
\label{synth-sect}

\begin{figure}
\begin{center}
\begin{tabular}{c@{\hspace{0in}}c@{\hspace{0in}}c@{\hspace{0in}}c@{\hspace{0in}}c@{\hspace{0in}}c}
\raisebox{-4ex}[0pt]{(a)} &
\epsfxsize=0.49in\epsfysize=2.94in\rotatebox{-90}{\epsfbox{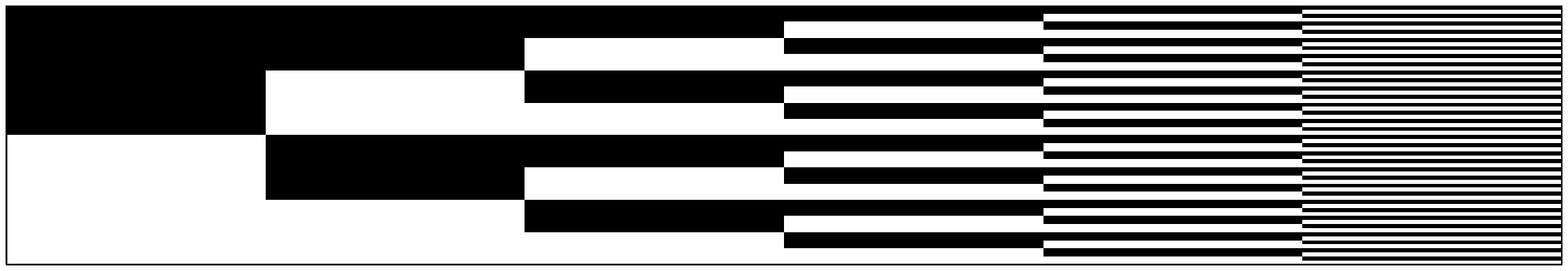}} \\
\raisebox{-4ex}[0pt]{(b)} &
\epsfxsize=0.49in\epsfysize=2.94in\rotatebox{-90}{\epsfbox{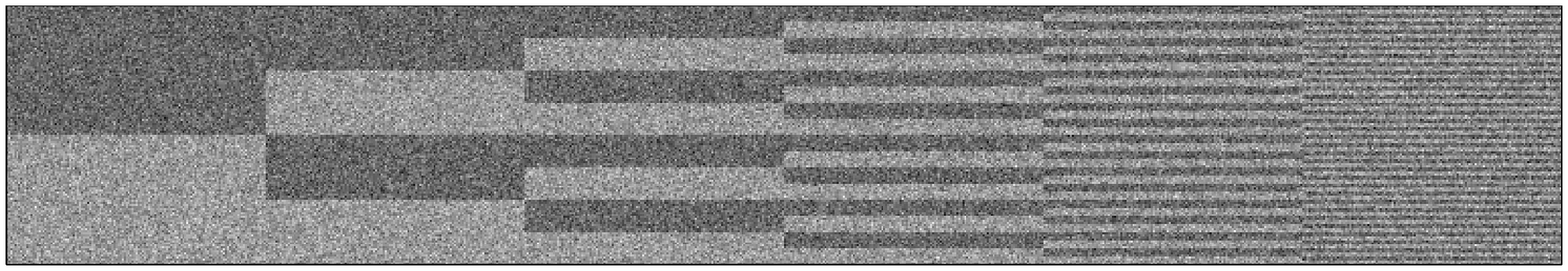}} \\
\raisebox{-4ex}[0pt]{(c)} &
\epsfxsize=0.49in\epsfysize=2.94in\rotatebox{-90}{\epsfbox{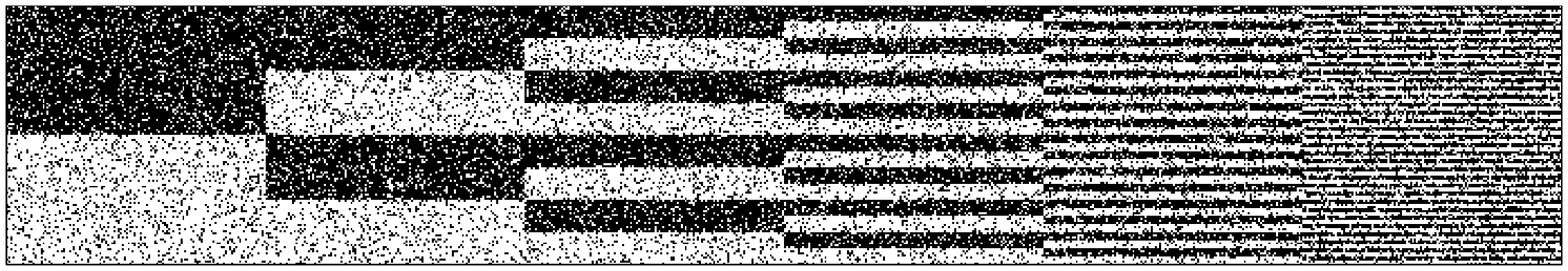}} \\
\raisebox{-4ex}[0pt]{(d)} &
\epsfxsize=0.49in\epsfysize=2.94in\rotatebox{-90}{\epsfbox{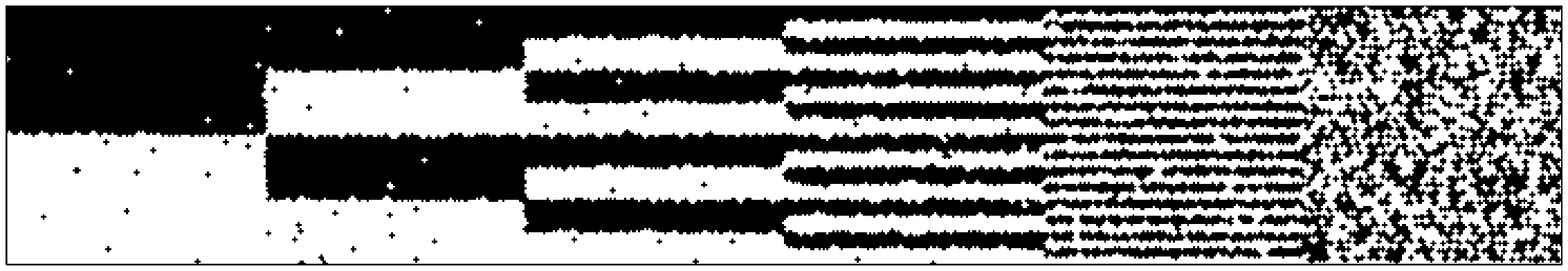}} \\
\raisebox{-4ex}[0pt]{(e)} &
\epsfxsize=0.49in\epsfysize=2.94in\rotatebox{-90}{\epsfbox{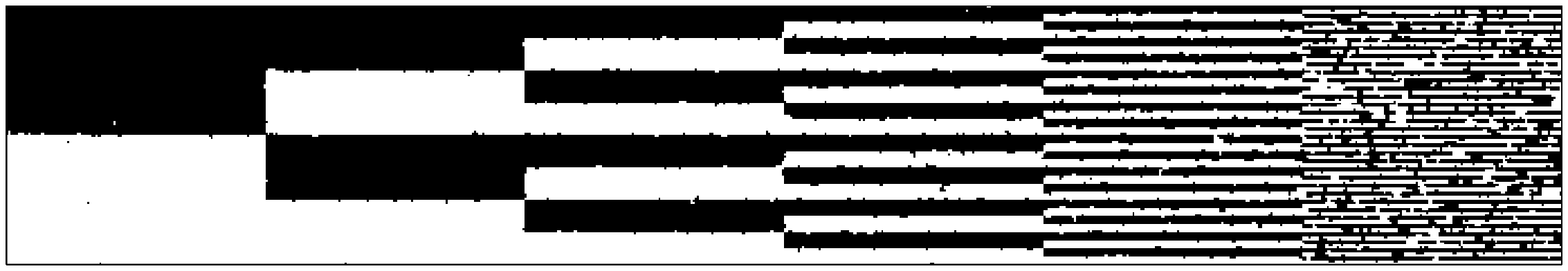}} \\
\raisebox{-4ex}[0pt]{(f)} &
\epsfxsize=0.49in\epsfysize=2.94in\rotatebox{-90}{\epsfbox{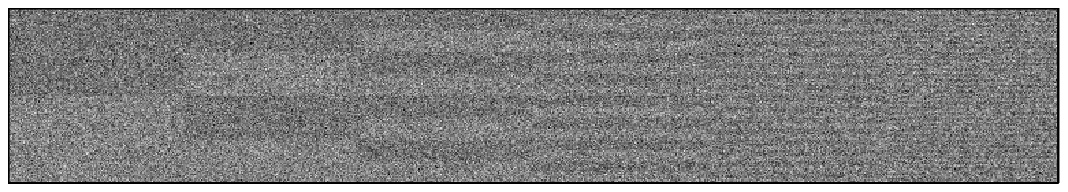}} \\
\raisebox{-4ex}[0pt]{(g)} &
\epsfxsize=0.49in\epsfysize=2.94in\rotatebox{-90}{\epsfbox{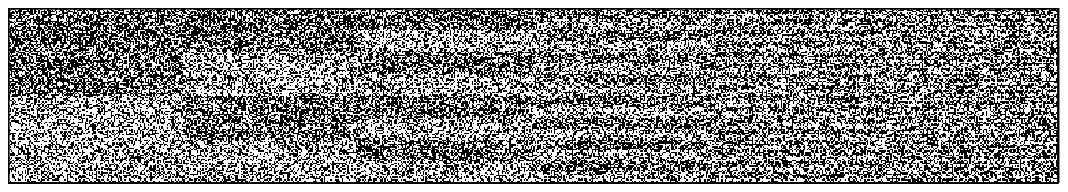}} \\
\raisebox{-4ex}[0pt]{(h)} &
\epsfxsize=0.49in\epsfysize=2.94in\rotatebox{-90}{\epsfbox{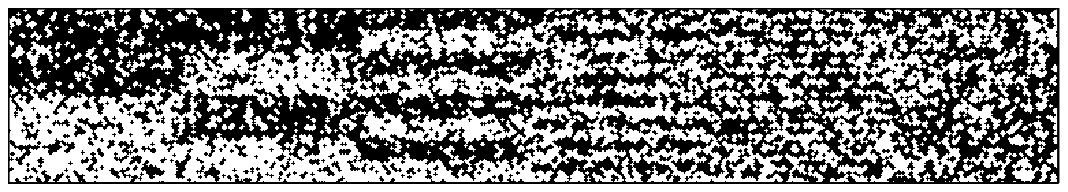}} \\
\raisebox{-4ex}[0pt]{(i)} &
\epsfxsize=0.49in\epsfysize=2.94in\rotatebox{-90}{\epsfbox{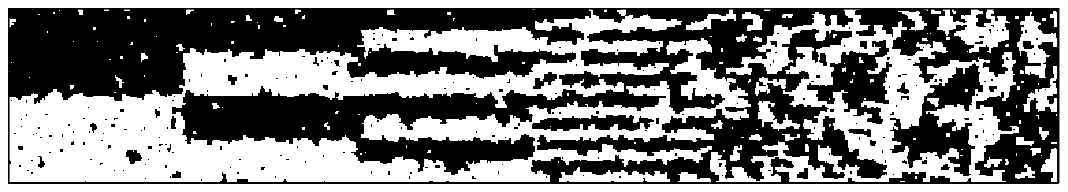}} \\
\end{tabular}
\end{center}
\caption{Segmentation results for synthetic data. (a) Ground truth;
(b-e) SNR = 2.0 results: input signal, control (thresholded),
morphological, graph; (f-i) SNR = 0.5 results: input signal, control
(thresholded), morphological, graph.}
\label{fig-synth}
\end{figure}

A preliminary set of experiments measures the performance of the two
algorithms on artificial data.  Using artificial data allows careful
control of experimental conditions.  Figure~\ref{fig-synth}a shows the
test pattern used, containing gradations in detail from coarse to
fine.  The right-hand portion of the image contains lines one pixel in
width spaced a single pixel apart, and successive portions to the left
double the width of both the lines and the gaps.
Figures~\ref{fig-synth}b and \ref{fig-synth}f show two images used as
input to the algorithms, formed by taking the ground truth and adding
noise.  Noise at each pixel is sampled independently from a normal
distribution of known variance, to generate an input with known
signal-to-noise ratio (SNR).
Figures~\ref{fig-synth}c-\ref{fig-synth}e and
\ref{fig-synth}g-\ref{fig-synth}i shows the results generated for the
two inputes shown in \ref{fig-synth}b and \ref{fig-synth}f.

Table~\ref{tbl-synth} gives the error rate on the best performance of
each algorithm for a number of different SNR values, including those
illustrated in Figure~\ref{fig-synth}.  As a baseline, it also gives
the performance achieved by simply thesholding the input image.  In
addition to the error rates, the parameter values used to achieve the
result also appear (except for $\tau$ on the graph algorithm, which is
always 0.5).  The values given correspond to the best result for a
particular algorithm on a particular input.

\begin{table}
\begin{center}
\caption{Error rates for the test patterns of Figure~\ref{fig-synth}.}
\label{tbl-synth}
\begin{tabular}{|c||c||c|cc||c|c|}
\hline
SNR & Thresh & Morph & $\tau$ & $r$ & Graph & $\alpha$ \\
\hline
4 & 2.3\% & 2.3\% & 0.50 & 0 & 0.0\% & 0.8 \\
2 & 15.8\% & 10.3\% & 0.70 & 1 & 2.8\% & 0.87 \\
1.5 & 22.8\% & 13.5\% & 0.75 & 1 & 5.9\% & 0.95 \\
1 & 30.9\% & 19.6\% & 0.95 & 1 & 11.2\% & 1.4 \\
0.8 & 34.3\% & 23.8\% & 1.05 & 1 & 14.0\% & 1.8 \\
0.65 & 37.3\% & 27.5\% & 1.20 & 1 & 16.7\% & 2.1 \\
0.5 & 40.1\% & 31.4\% & 1.45 & 1 & 19.8\% & 2.5 \\
\hline
\end{tabular}
\end{center}
\end{table}

The graph cut algorithm performs markedly better on the synthetic data
than either the control or the morphological method at all
signal-to-noise ratios.  The error rate for the morphological
techniques is always better than that for the control, except at the
highest SNR (where they tie).

Examination of the graphical output in Figure~\ref{fig-synth} shows
that at SNR = 2.0 both trial algorithms do fairly well, but the graph
cut result displays cleaner edges and captures more of the fine detail
at the highest resolution.  The morphological result loses the details
of the highest-resolution section at the right of the test image.  At
the SNR = 0.5, where the outlines of the input test pattern can barely
be discerned by human eyes, the graph cut result still looks
reasonable for the areas with coarser detail.  Both algorithms lose
the details in the two sections of highest resolution.  Interestingly,
increasing $\alpha$ in the graph cut algorithm can further improve
results on the low-resolution segments, albeit at the expense of
further degradation at high and medium resolution.


\Section{Experiments With Real Images}
\label{video-sect}

The main set of experiments explores the performance of the algorithms
on video taken under real world conditions.  The three video clips
used cover a range in quality and subject: from color clips shot both
indoors and outdoors with fixed cameras, to a low-quality grayscale
video of a ballet dancer with a panning camera and compression
artifacts.  The former represent relatively easy conditions, while the
latter presents a stiffer challenge for background subtraction.
Table~\ref{tbl-clips} gives more details on each clip.

\begin{table}
\begin{center}
\caption{Details on the video clips used for testing.}
\label{tbl-clips}
\begin{tabular}{|@{}ll|}
\hline
\raisebox{-10pt}{\epsfxsize=1.2in\epsfysize=0.9in\epsfbox{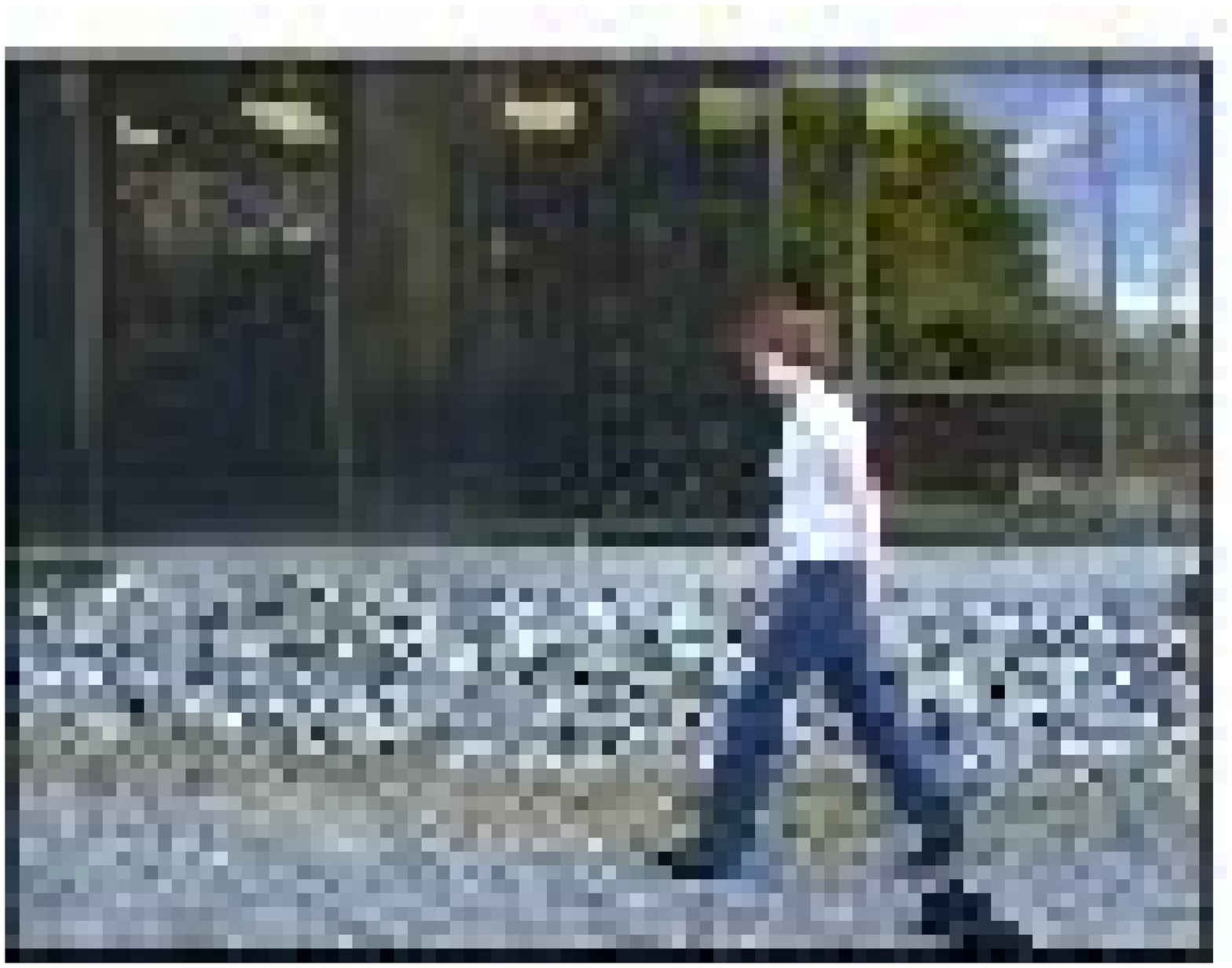}}
& \raisebox{5ex}[0pt]{\parbox{1.8in}{Clip \#2: Outdoor clip, regular
motion, some reflection off glass.  Fixed camera.  124 frames.}} \\
\hline \hline
\raisebox{-10pt}{\epsfxsize=1.2in\epsfysize=0.9in\epsfbox{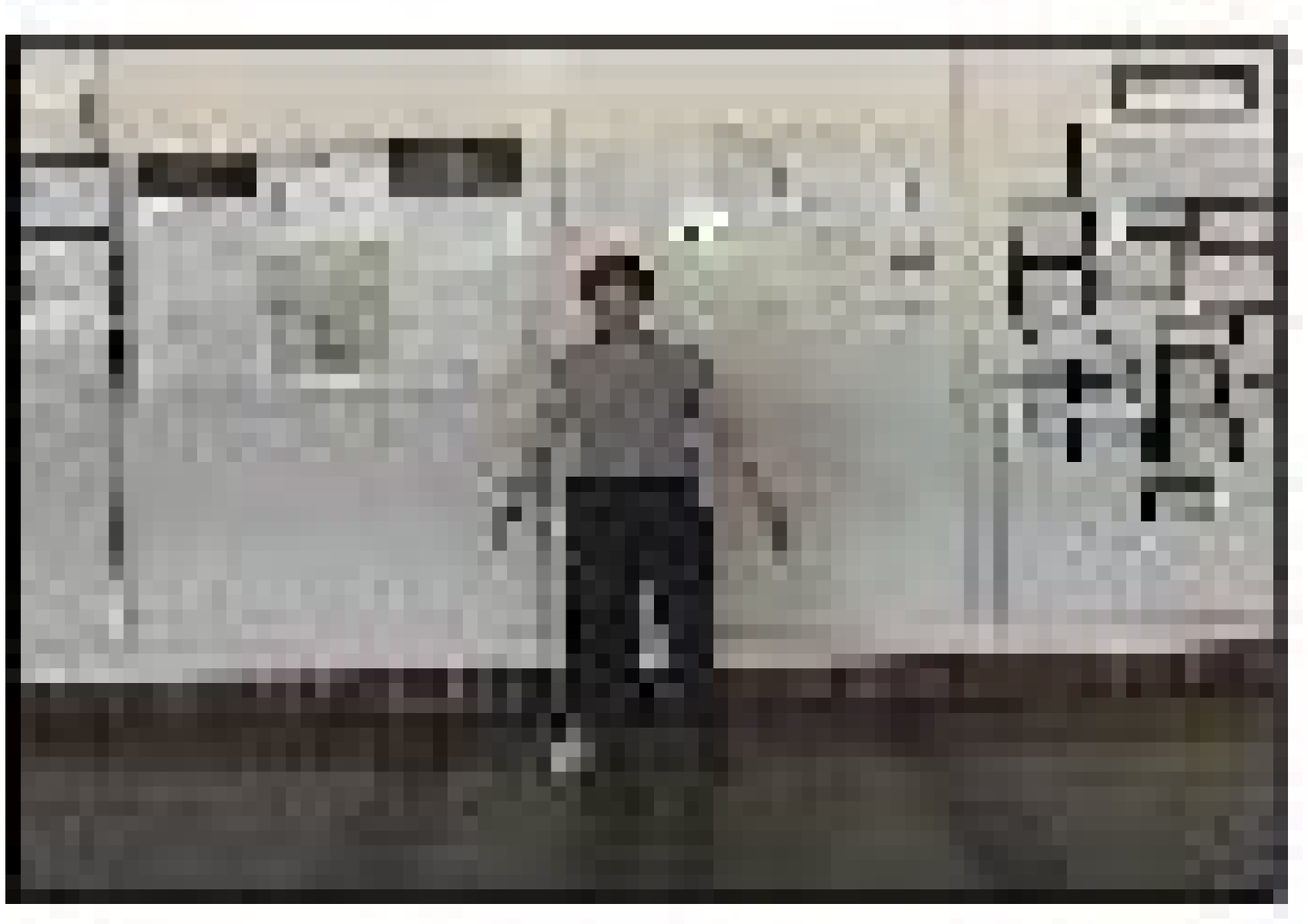}}
& \raisebox{5ex}[0pt]{\parbox{1.8in}{Clip \#1: Indoor clip with some
shadowing, reflections off the floor.  Some low contrast portions.
Fixed camera.  160 frames.}} \\ \hline \hline
\raisebox{-10pt}{\epsfxsize=1.2in\epsfysize=0.9in\epsfbox{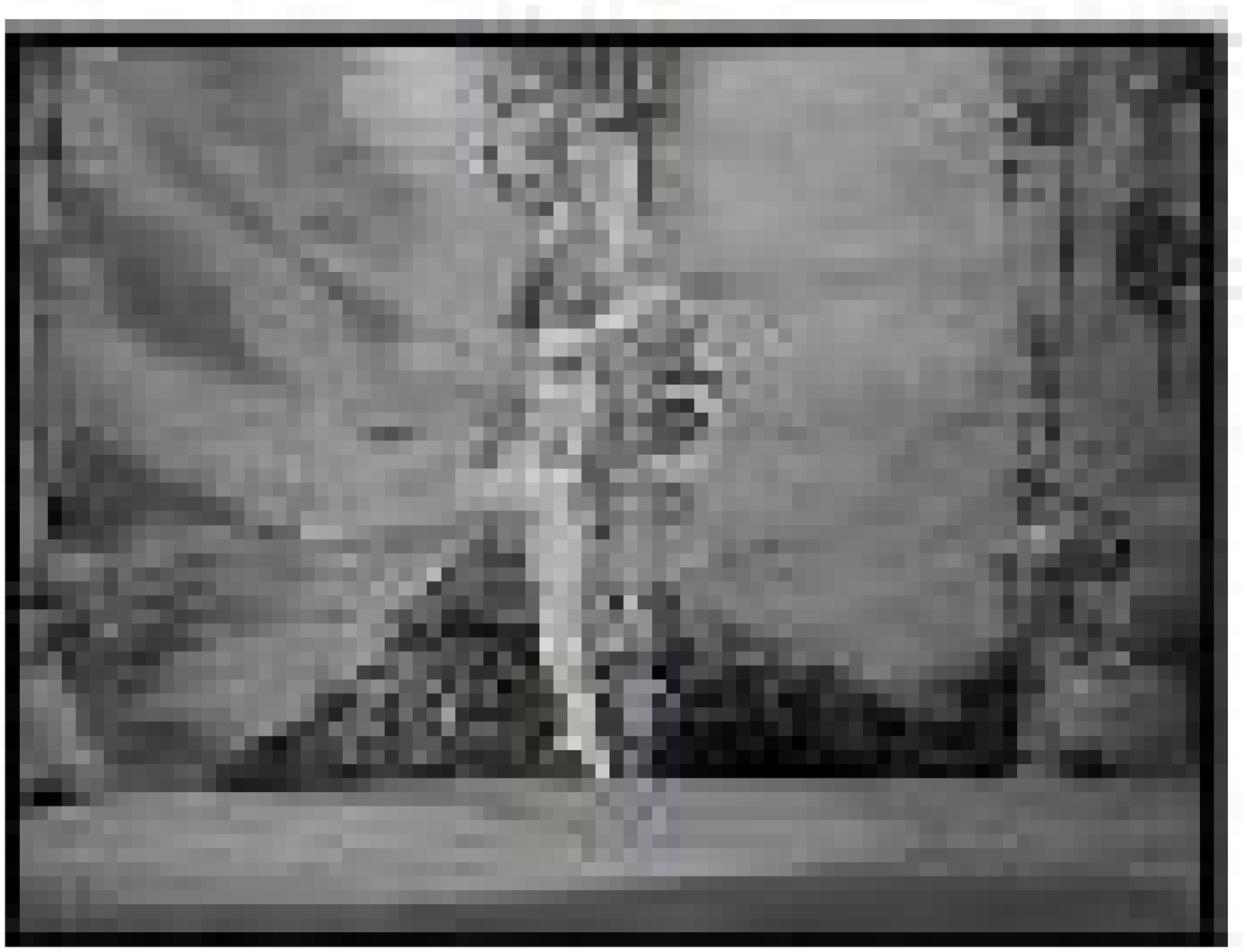}}
& \raisebox{5ex}[0pt]{\parbox{1.8in}{Clip \#3: Grayscale MPEG video of
a dancer.  Low contrast.  Panning camera.  99 frames.}} \\ \hline
\end{tabular}
\end{center}
\end{table}


Each clip must undergo extensive preprocessing before reaching the
stage where the algorithms on trial may be applied.  Although some may
object to the processing choices made here (particularly the use of
static background models), changes in these choices would amount only
to a change of the input to both trial algorithms, akin to choosing
different videos for the tests.  Neither of the approaches under
consideration precludes algorithmic improvements to the preprocessing
stages, but more complicated preprocessing introduces potentially
confounding design decisions.  Therefore, for simplicity and
replicability, the experiments avoid sophisticated preprocessing where
possible.  For the first two clips, the experiments eschew the
dynamically updated background models used by most current systems
(e.g., $W^4$ \cite{haritaoglu:w4}) in favor of a static background
model computed over the entire clip.  (A static background model is
also comparable to a dynamic model that has been allowed to
equilibrate.)

To prepare a video for background subtraction, a background model is
built using crudely robust statistical techniques.  The model builder
takes the pixel color from every fourth frame, and throws out the data
above and below a pair of thresholds (say the 25th and the 75th
percentiles).  From the remaining numbers it estimates the mean and
variance of each pixel's color, assuming a normal distribution.  This
approach provides effective robustness to occlusions of the background
on a small fraction of the video frames.  (More complicated pixel
modeling based upon mixtures of Gaussians can be used, but again this
would only confound the comparison of the two algorithms under trial.)

\begin{figure}
\begin{center}
\epsfxsize=3.01in\epsfysize=1.36in\epsfbox{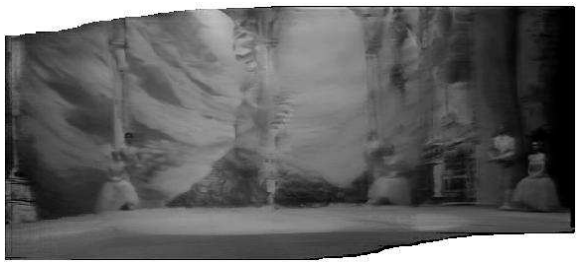}
\end{center}
\caption{Registered median background model for Clip \#3, {\em
Dancer}.  The camera pans from right to left, following the dancer.}
\label{fig-barybg}
\end{figure}

For the third clip, the camera movement necessitates special
treatment.  Before building the background model described above, each
frame must be registered with a canvas representing the entire
background.  A least-squares fit based on frame-to-frame optical flow
generates an approximate set of initial registrations, expressed as
affine transforms.  The computer then builds up the canvas frame by
frame, using function minimization to find the affine transform
yielding the smallest disparities between the new frame and the median
values on the existing canvas.  The pixel values for the frame then
get interpolated onto the canvas, yielding new median values.
Figure~\ref{fig-barybg} shows the result of this process.

With the completed model (and registrations, if necessary) in hand,
each frame can be compared with the mean background image.  The
difference at each pixel, normalized by the variance at that pixel,
forms the raw data input to the two finishing algorithms.  Typically
the calculations might be carried out on each of the red, green, and
blue (RGB) components of an image, and the differences in each
component summed together.  However, several modifications to the
process provide necessary tolerance to shadows and lighting changes.
Making the comparisons in the hue-saturation-value (HSV) color space
causes disparities due to shadows to show up primarily in one channel,
namely the pixel value ($V$).  Areas in shadow display lower $V$
component values than the unshadowed background, but are similar in
the other two components.  Therefore, discounting small decreases
(less than 5\% of the total range) in $V$ effectively ensures that
shadowed background areas do not falsely appear to belong to the
foreground.  This method effectively mirrors other recent results
\cite{horprasert:shadow}, although that work differs superficially by
processing images in RGB color space.

\begin{figure}[tb]
\begin{center}
\begin{tabular}{c@{\hspace{2pt}}c}
\epsfxsize=1.48in\epsfysize=1.11in\epsfbox{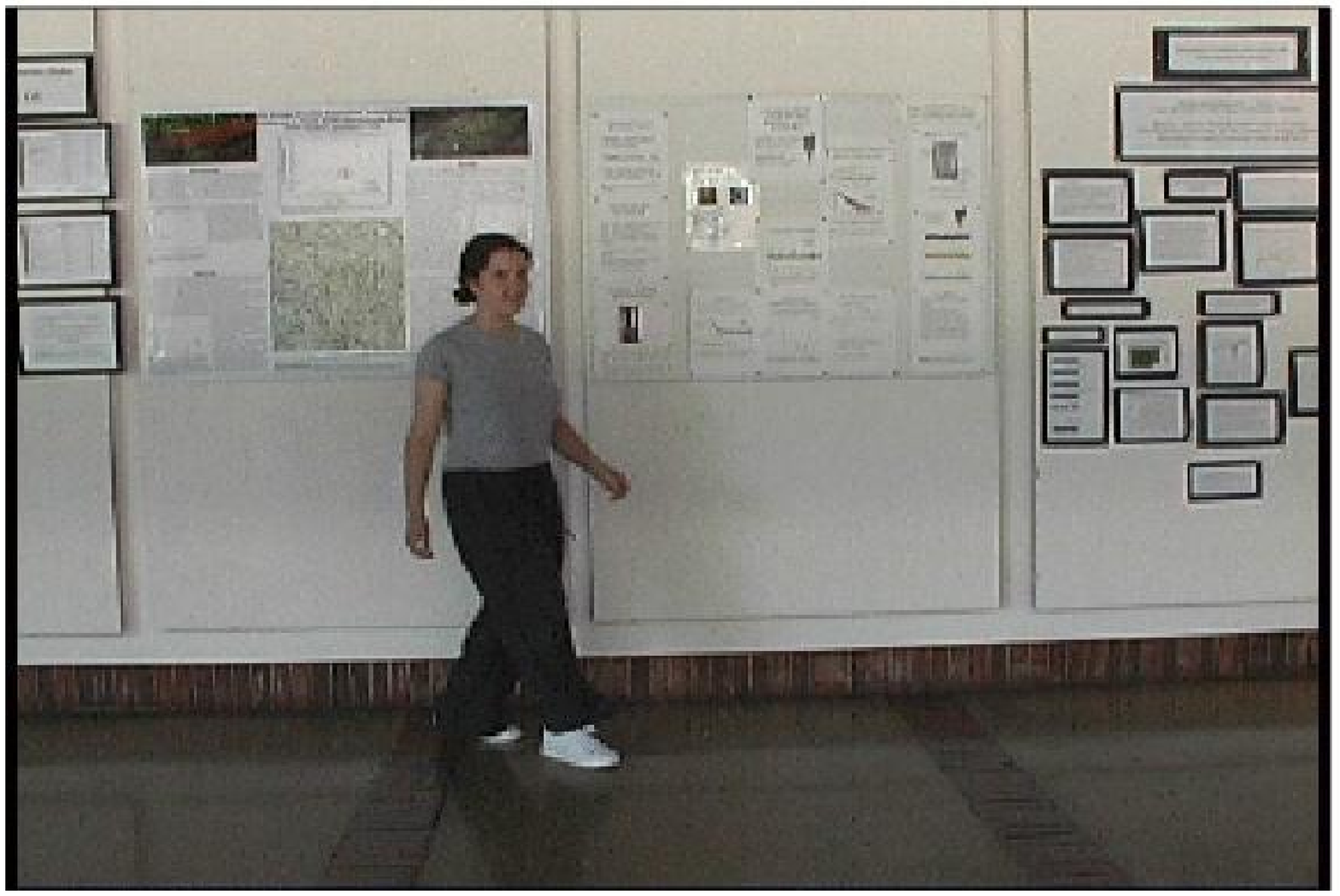} &
\epsfxsize=1.48in\epsfysize=1.11in\epsfbox{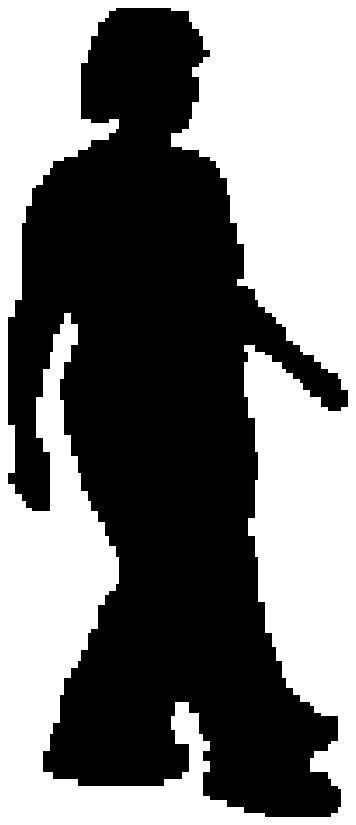} \\
(a) & (b) \\
\epsfxsize=1.48in\epsfysize=1.11in\epsfbox{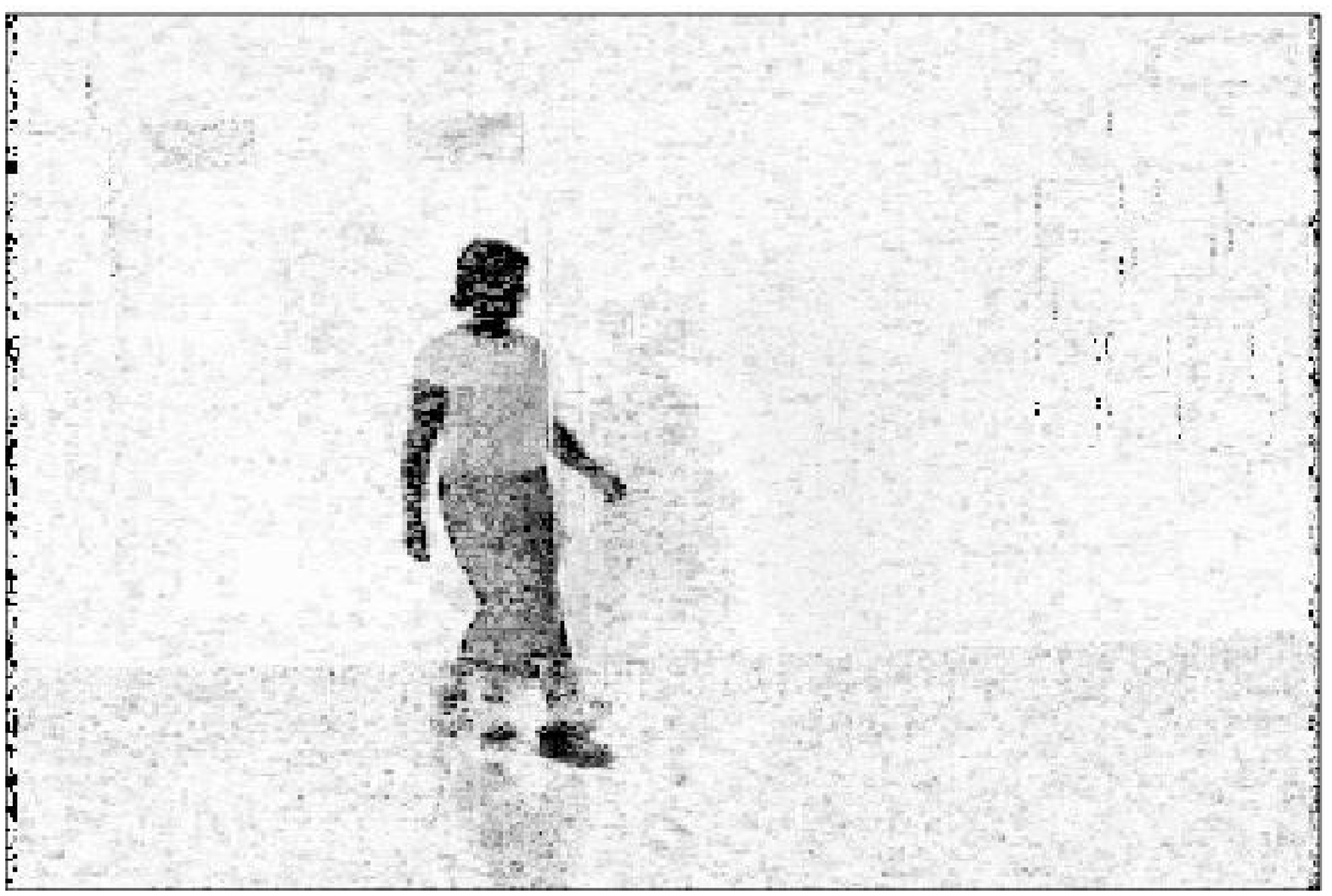} &
\epsfxsize=1.48in\epsfysize=1.11in\epsfbox{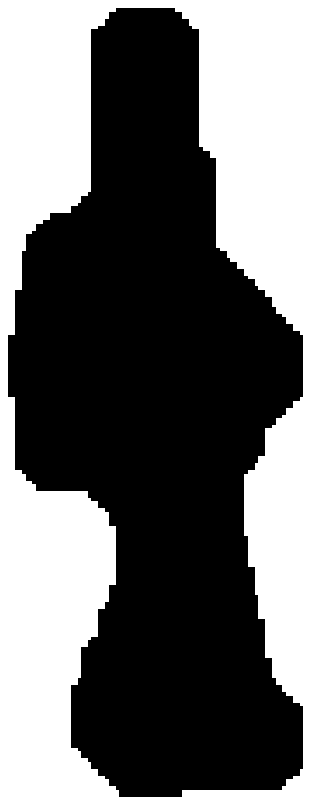} \\
(c) & (d) \\
\epsfxsize=1.48in\epsfysize=1.11in\epsfbox{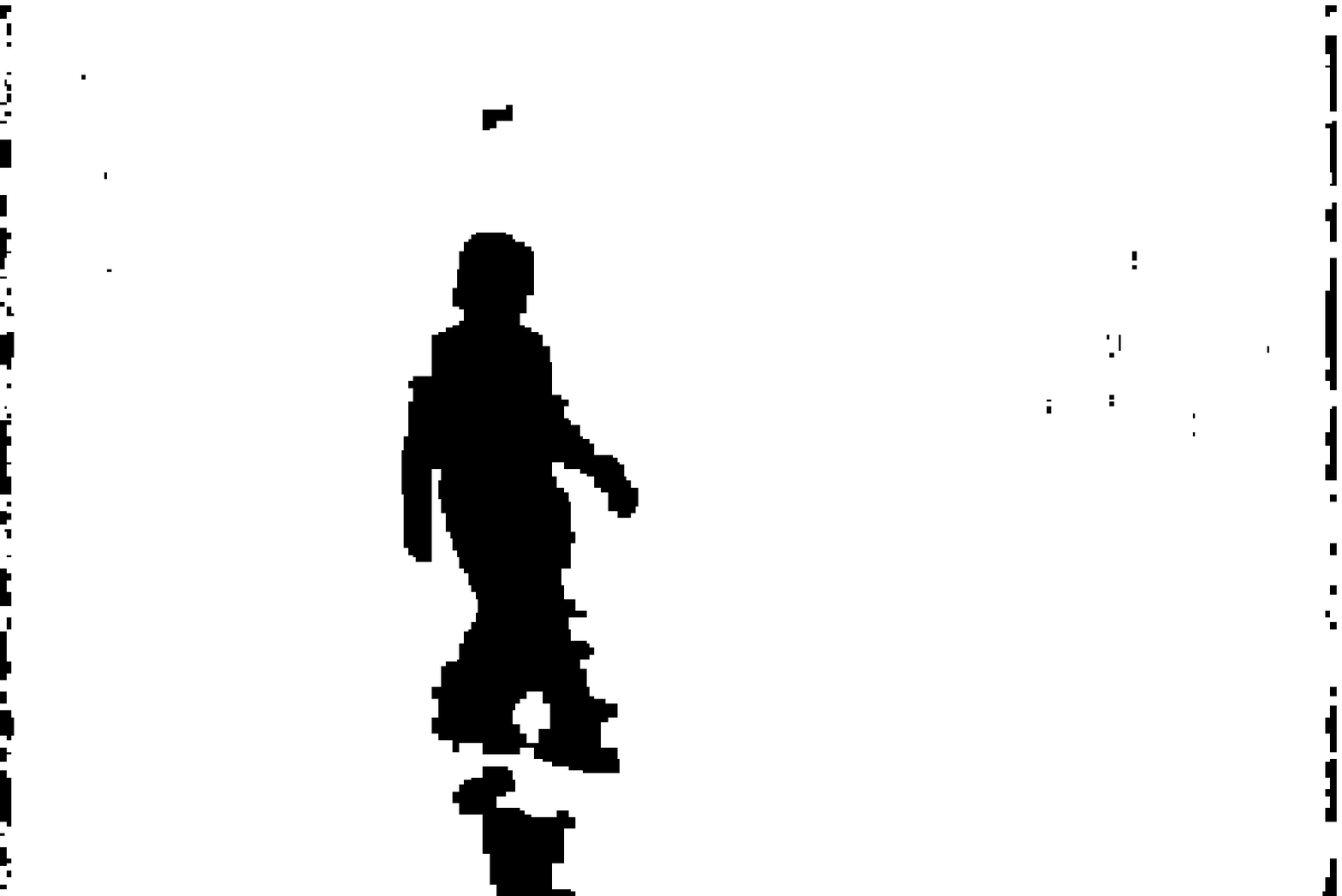} &
\epsfxsize=1.48in\epsfysize=1.11in\epsfbox{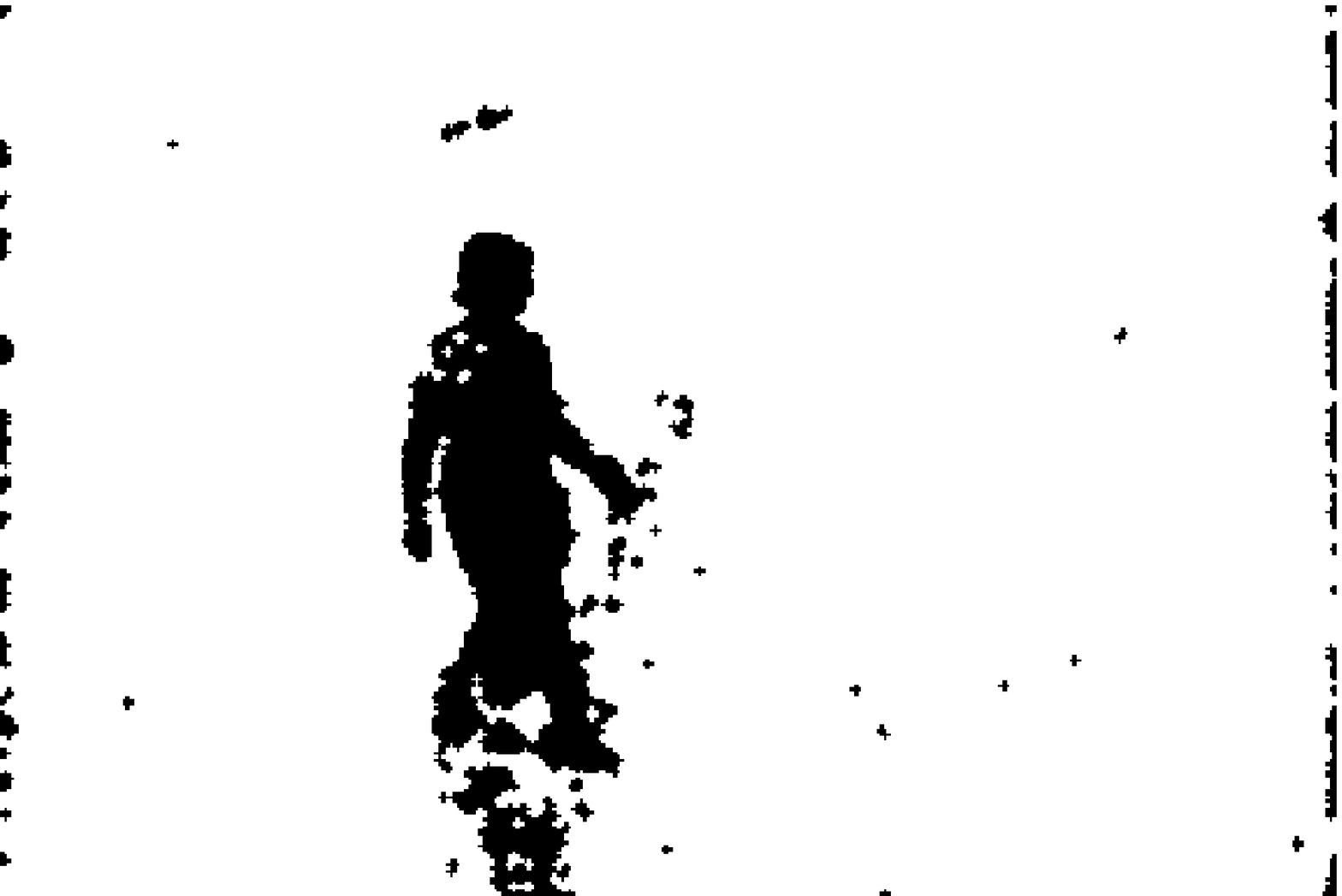} \\
(e) & (f) \\
\end{tabular}
\end{center}
\caption{Effects of different error measurements.  (a) Original
image. (b) Ground truth.  (c) Difference from background. (d) Best
morphological segmentation by whole-image criterion.  (e) Best graph
and (f) morphological segmentations using the connected components
criterion.}
\label{fig-err}
\end{figure}

A suitable criterion must be chosen for grading the segmentation
results.  The total fraction of pixels in the image differing from
ground truth appears an attractive error measure at first glance, but
a closer look reveals flaws in this criterion.  Figure~\ref{fig-err}d
shows a segmentation created using morphological techniques with very
large structuring element (a disk of radius 10).  Although it captures
few details of the subject figure's outline accurately, its error over
the whole frame is 1.25\%.  Figure~\ref{fig-err}e and \ref{fig-err}f
show alternate segmentations that look more faithful to the exact
outlines of the ground truth.  Yet these have higher whole-frame
errors of 2.24\% and 2.59\%, respectively, largely due to noise at the
edges of the frame and other areas separate from the main figure.

In order to reward segmentations similar to Figure~\ref{fig-err}e and
\ref{fig-err}f, the experiments employ a modified error measure
focusing specifically on the moving figure in the clip.  The measure
first computes the connected components of the segmented foreground
and identifies all the components that overlap with ground truth (as
identified by a human operator tracing the figure's outline in
Photoshop).  Pixels in the selected components that do not overlap
with ground truth count as false positives, while ground truth areas
identified as background by the segmentation algorithm count as false
negatives.  Combining the number of false positives and false
negatives, then scaling by the number of pixels in the ground truth
yields the error measurement for the frame.

\SubSection{Real Image Results}
\label{result-sect}

Table~\ref{tbl-videos} shows the error rates of both the morphological
and graph algorithms on the video inputs, after tuning the algorithm
parameters for best results.  (Interestingly, although larger
stucturing elements were tested for the morphological operations,
small radius-one disk-shaped elements give the best results on all
three clips.)  Examining the numbers, one sees that the quality of the
video input forms the largest factor determining the error, with far
more mistakes for either approach on the {\em Dancer} clip.
Nevertheless, the graph-based algorithm performs significantly better
(in a statistical sense) on every clip tested, according to a paired
sample t-test.

\begin{figure}
\begin{center}
\begin{tabular}{|@{}c@{}|@{}c@{}|@{}c@{}|} \hline
\epsfxsize=1in\epsfysize=0.67in\epsfbox{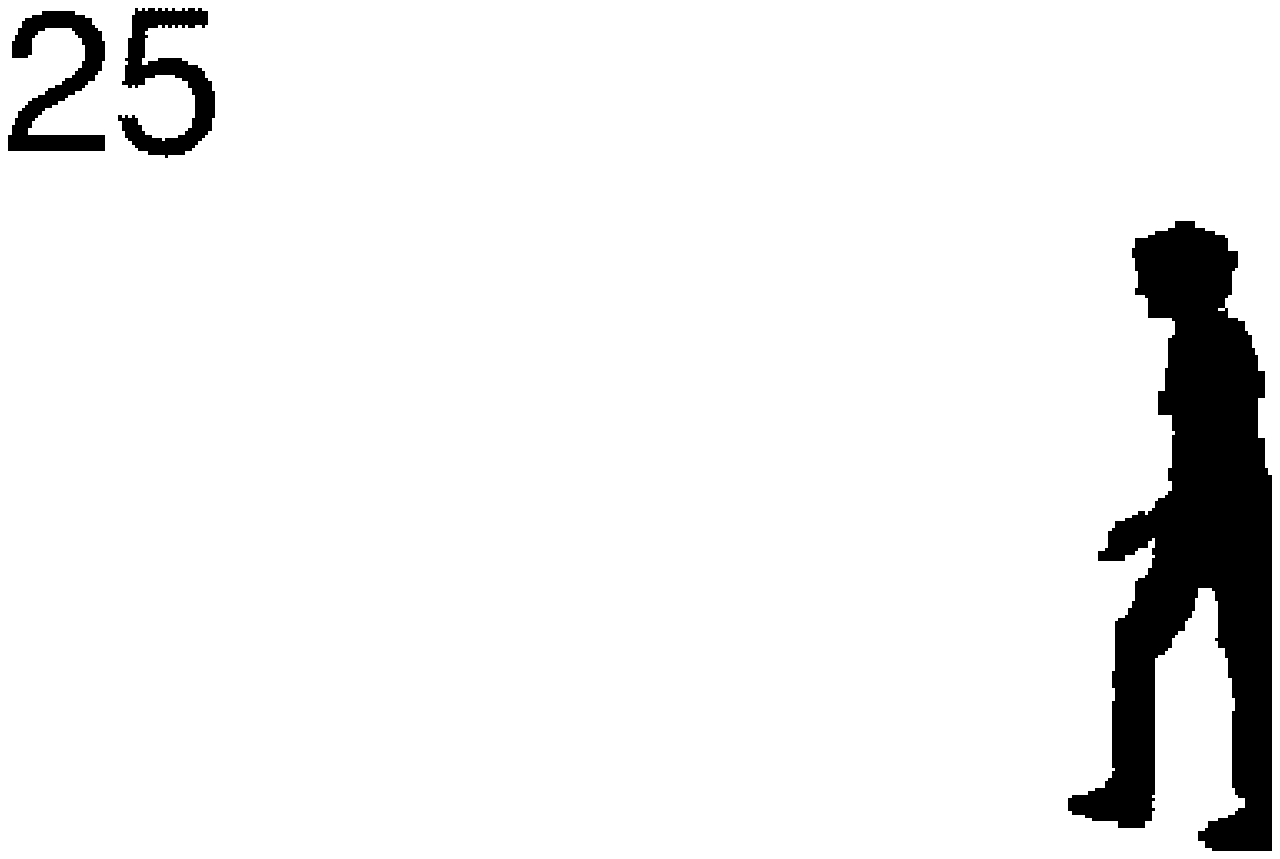} &
\epsfxsize=1in\epsfysize=0.67in\epsfbox{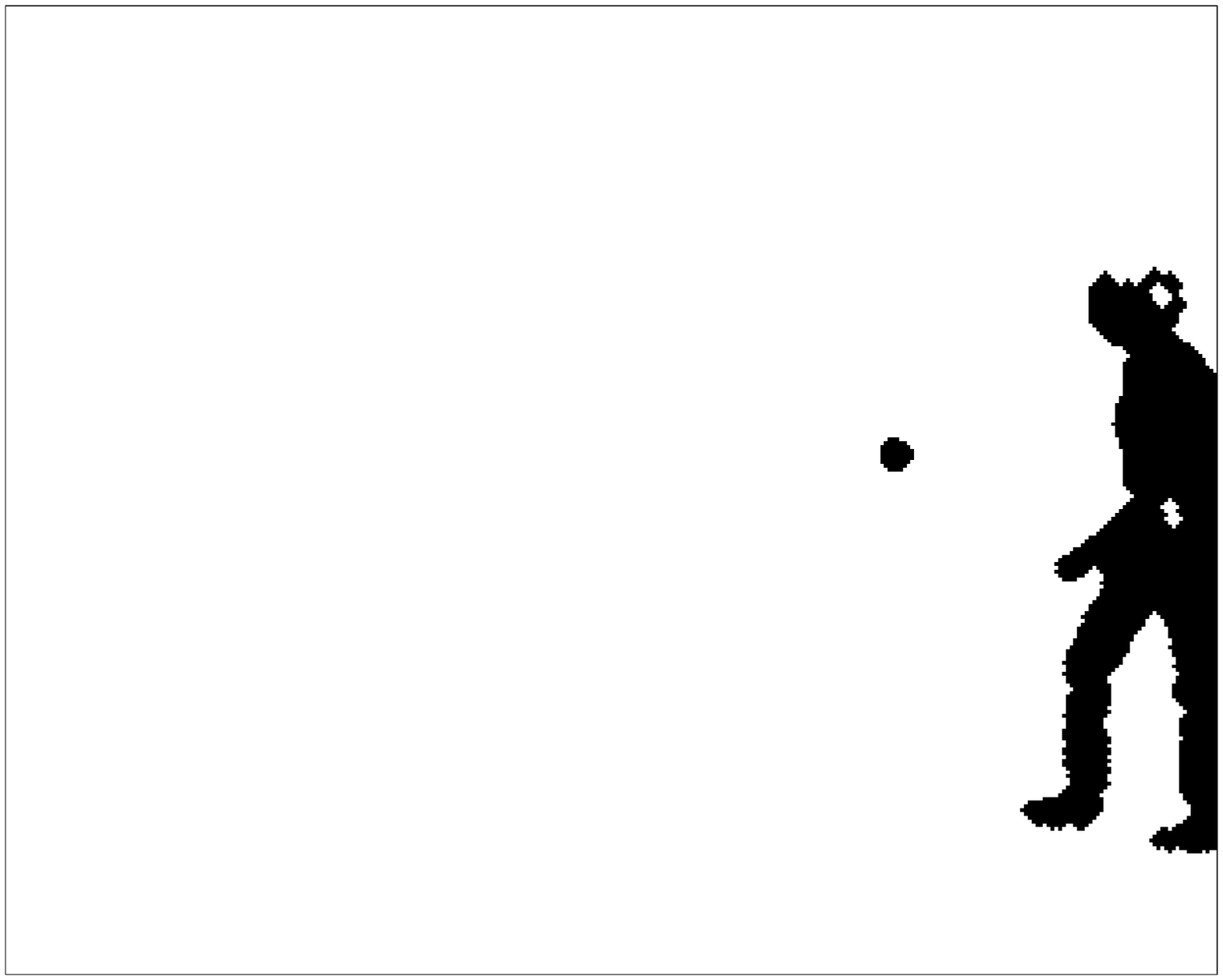} &
\epsfxsize=1in\epsfysize=0.67in\epsfbox{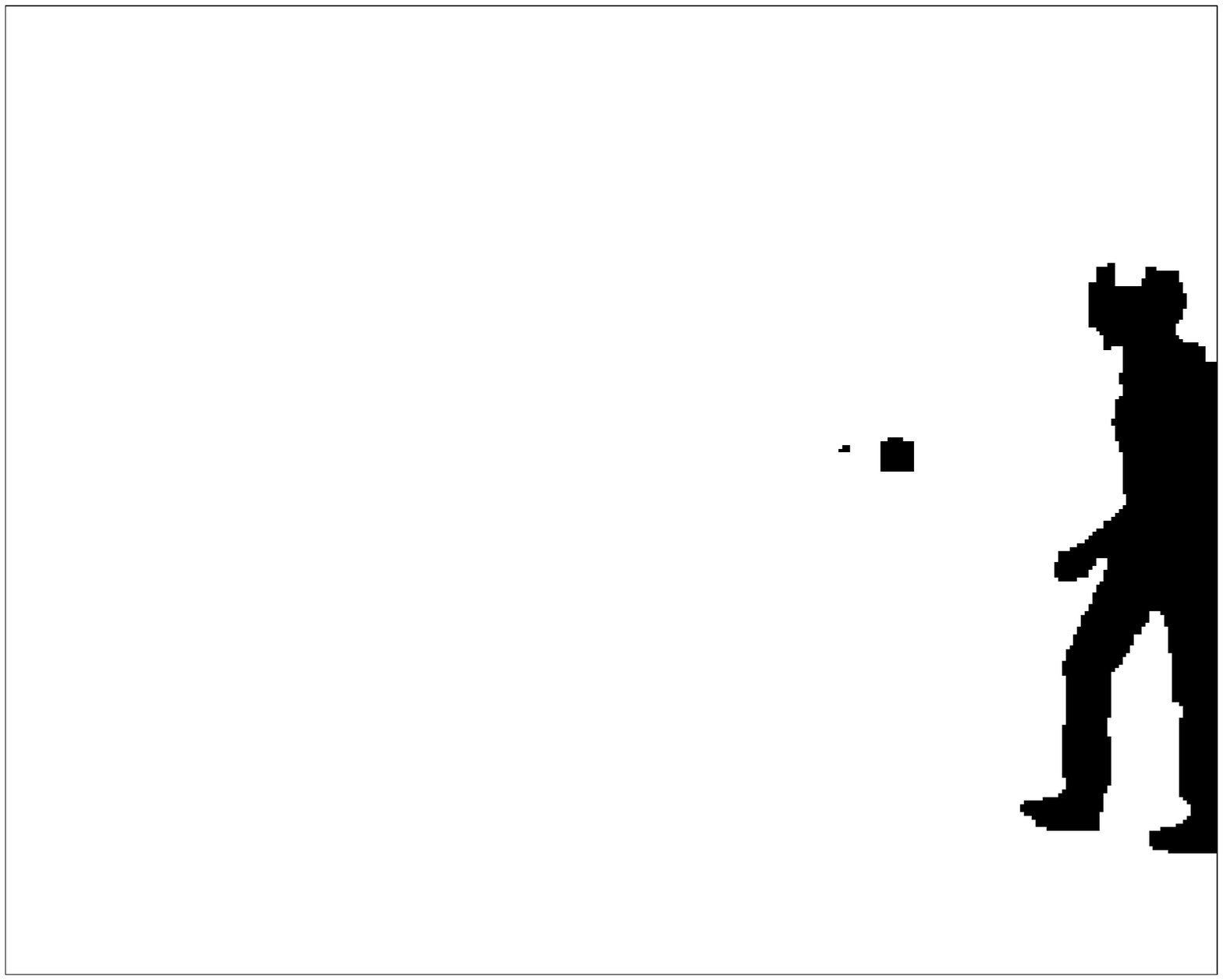} \\ \hline
\epsfxsize=1in\epsfysize=0.67in\epsfbox{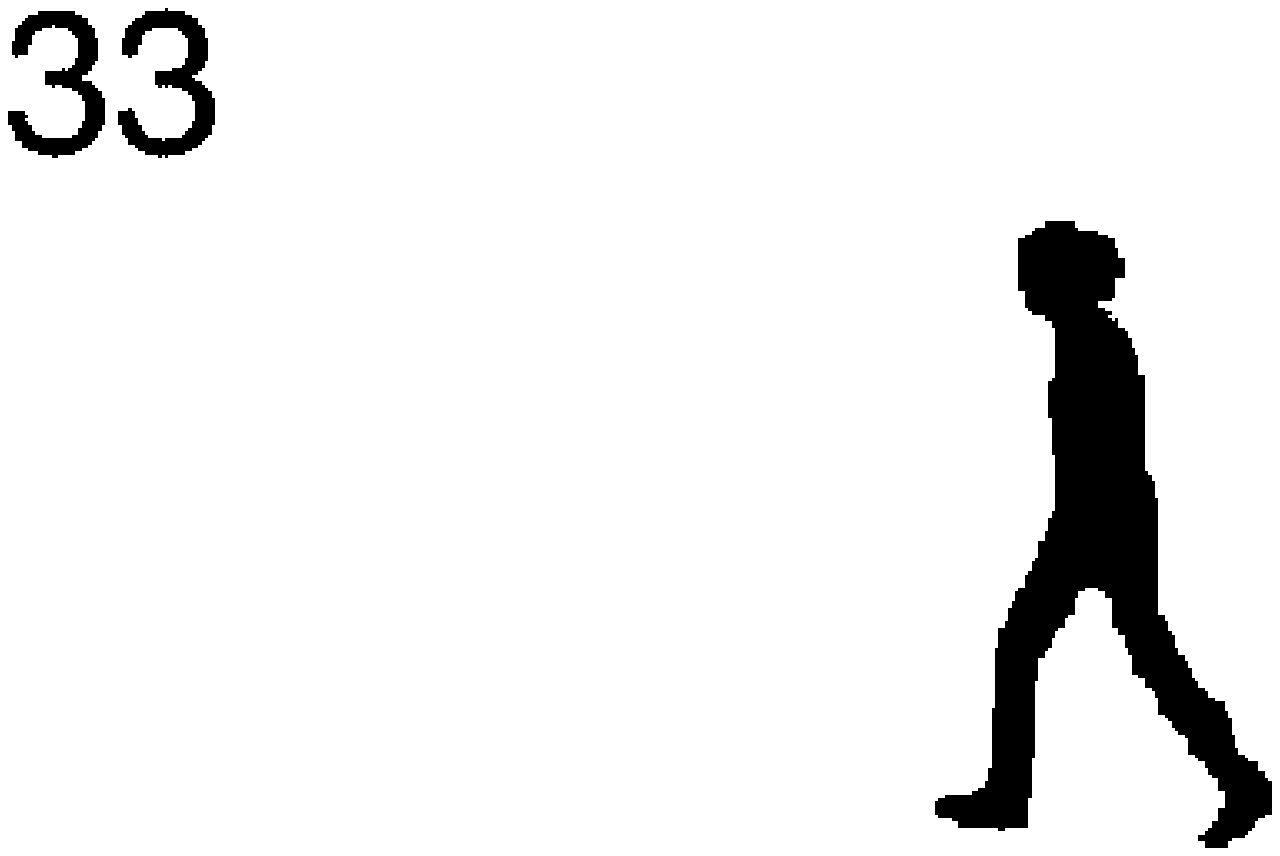} &
\epsfxsize=1in\epsfysize=0.67in\epsfbox{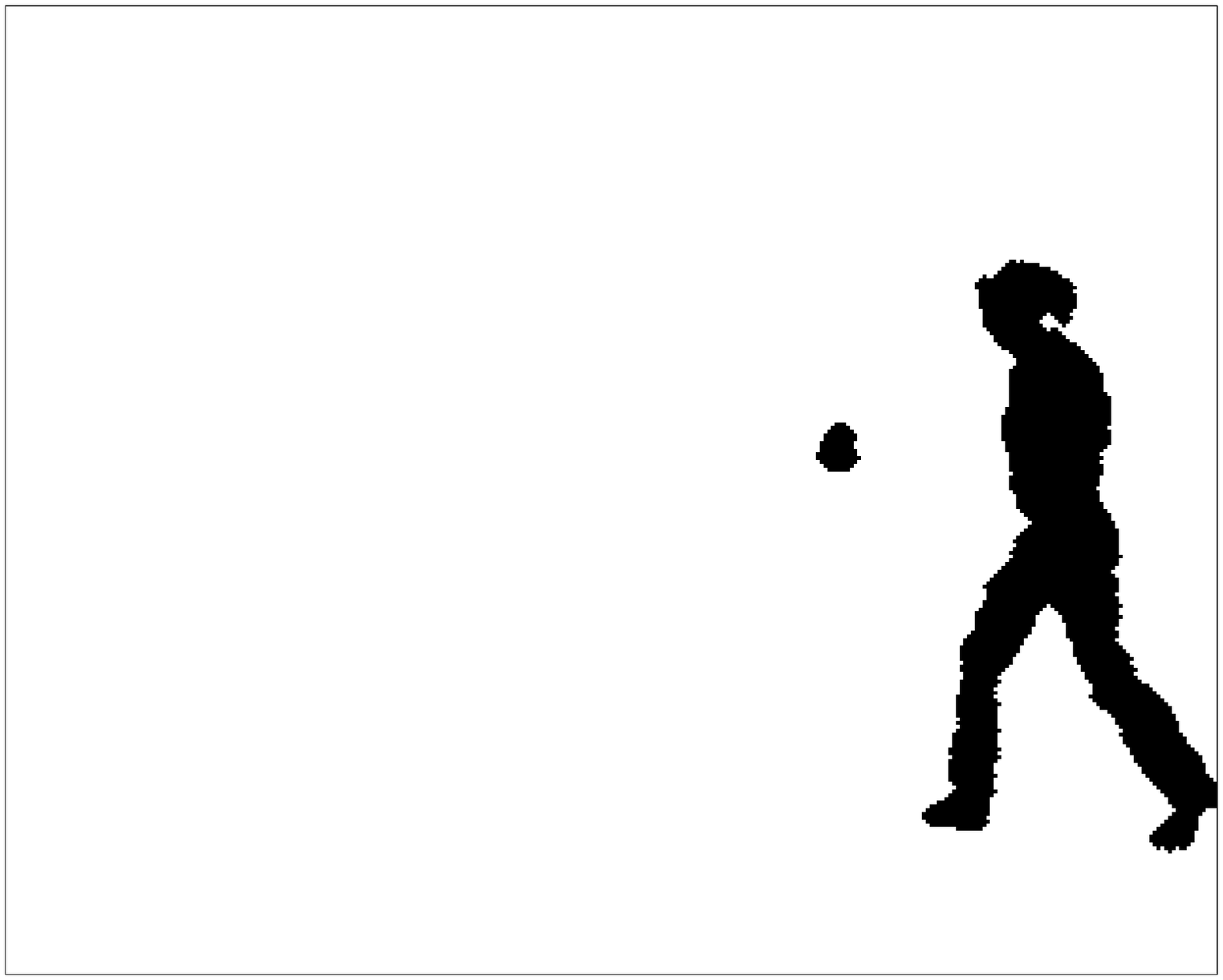} &
\epsfxsize=1in\epsfysize=0.67in\epsfbox{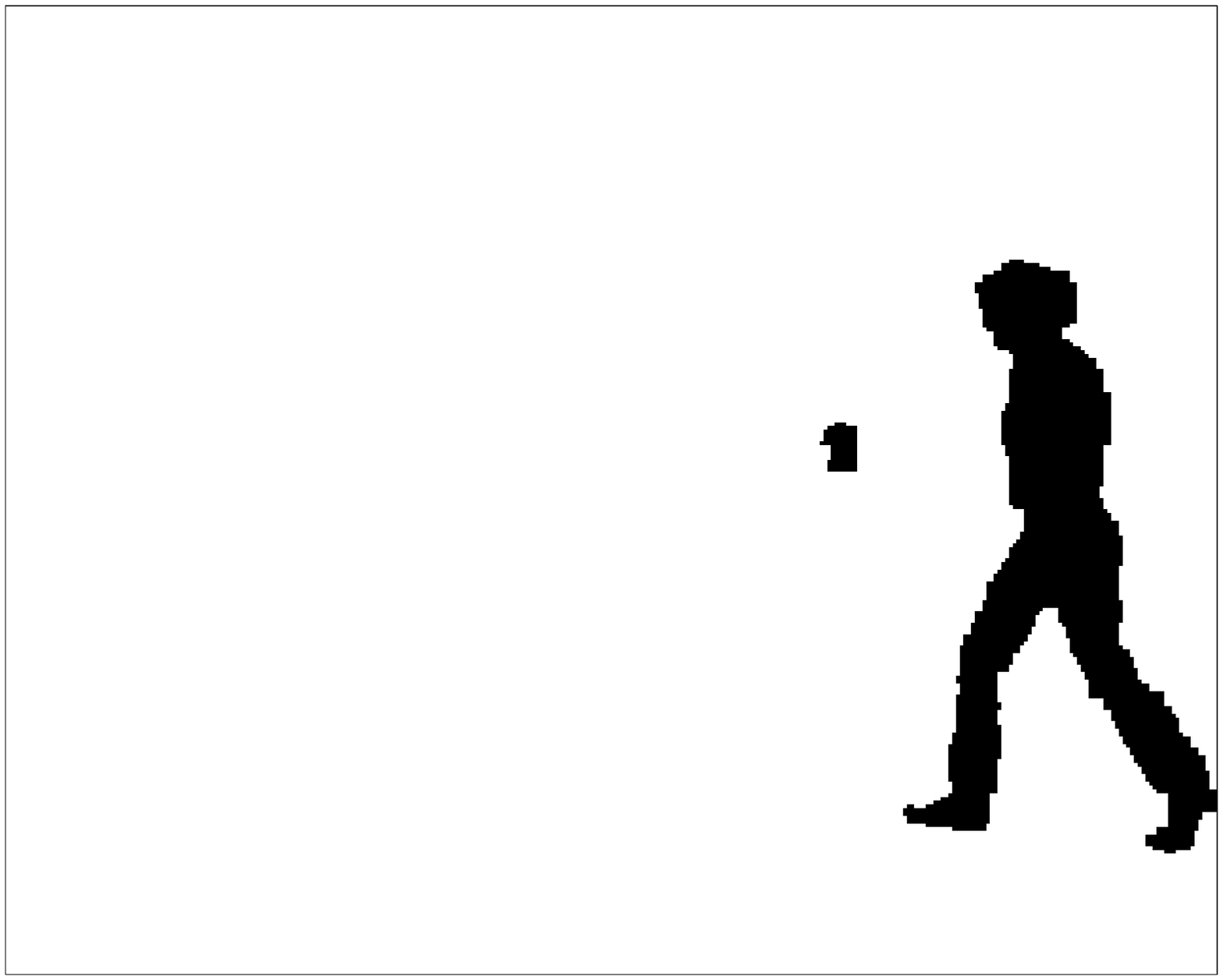} \\ \hline
\epsfxsize=1in\epsfysize=0.67in\epsfbox{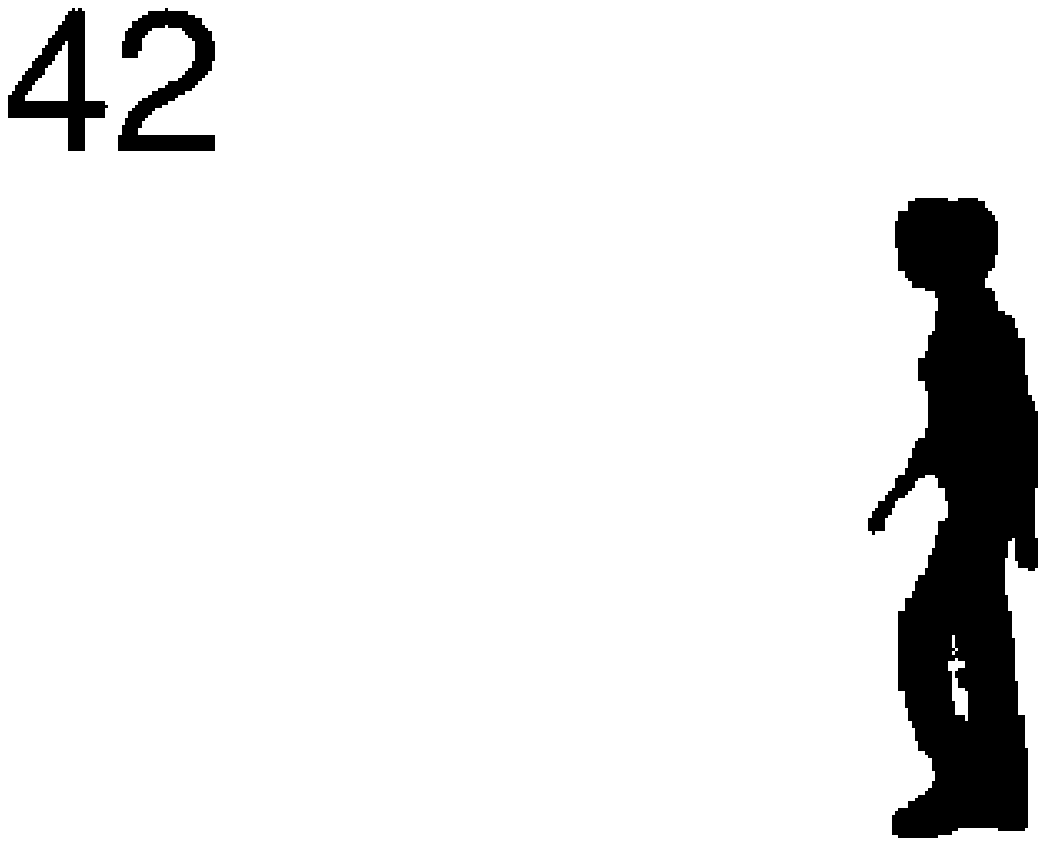} &
\epsfxsize=1in\epsfysize=0.67in\epsfbox{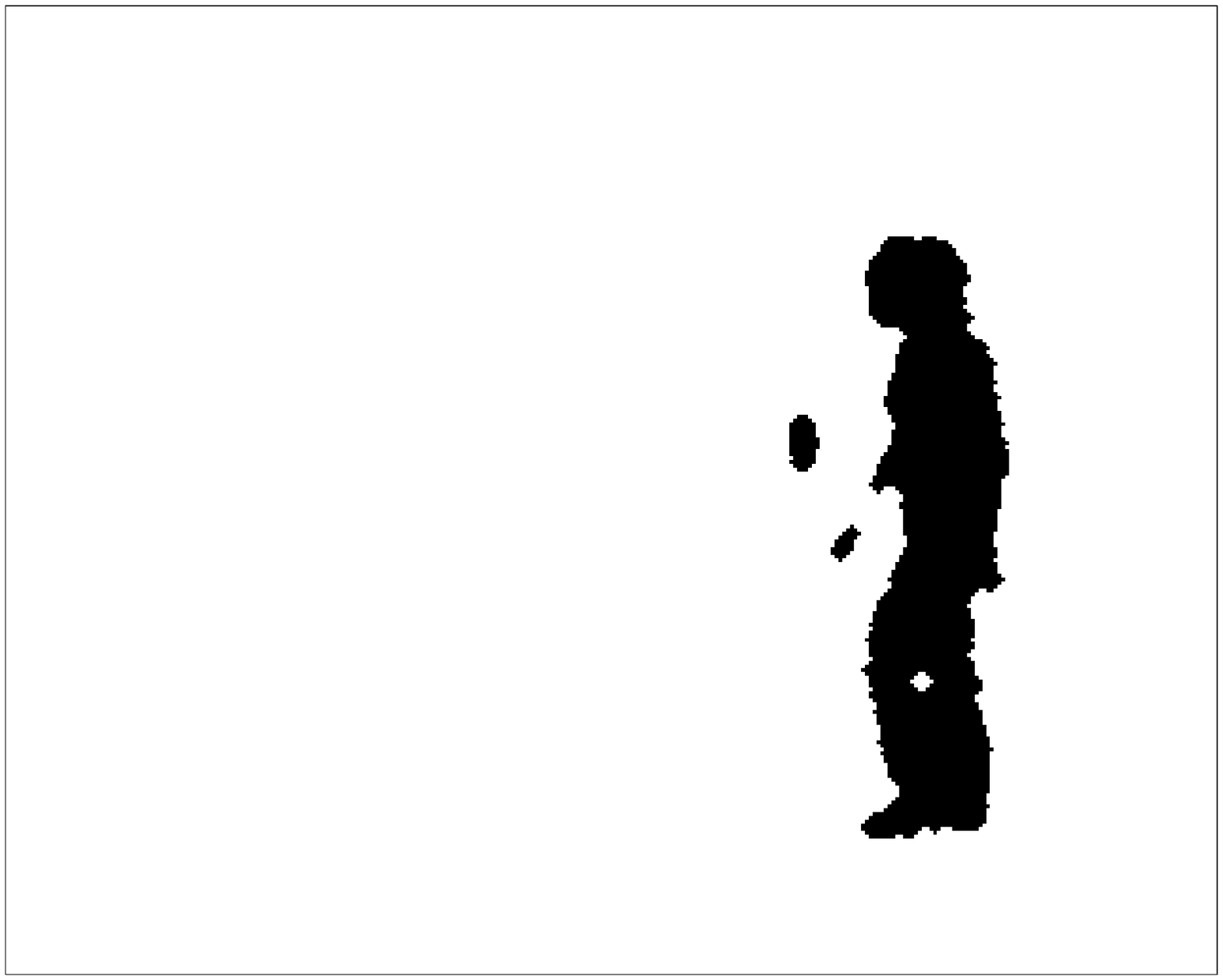} &
\epsfxsize=1in\epsfysize=0.67in\epsfbox{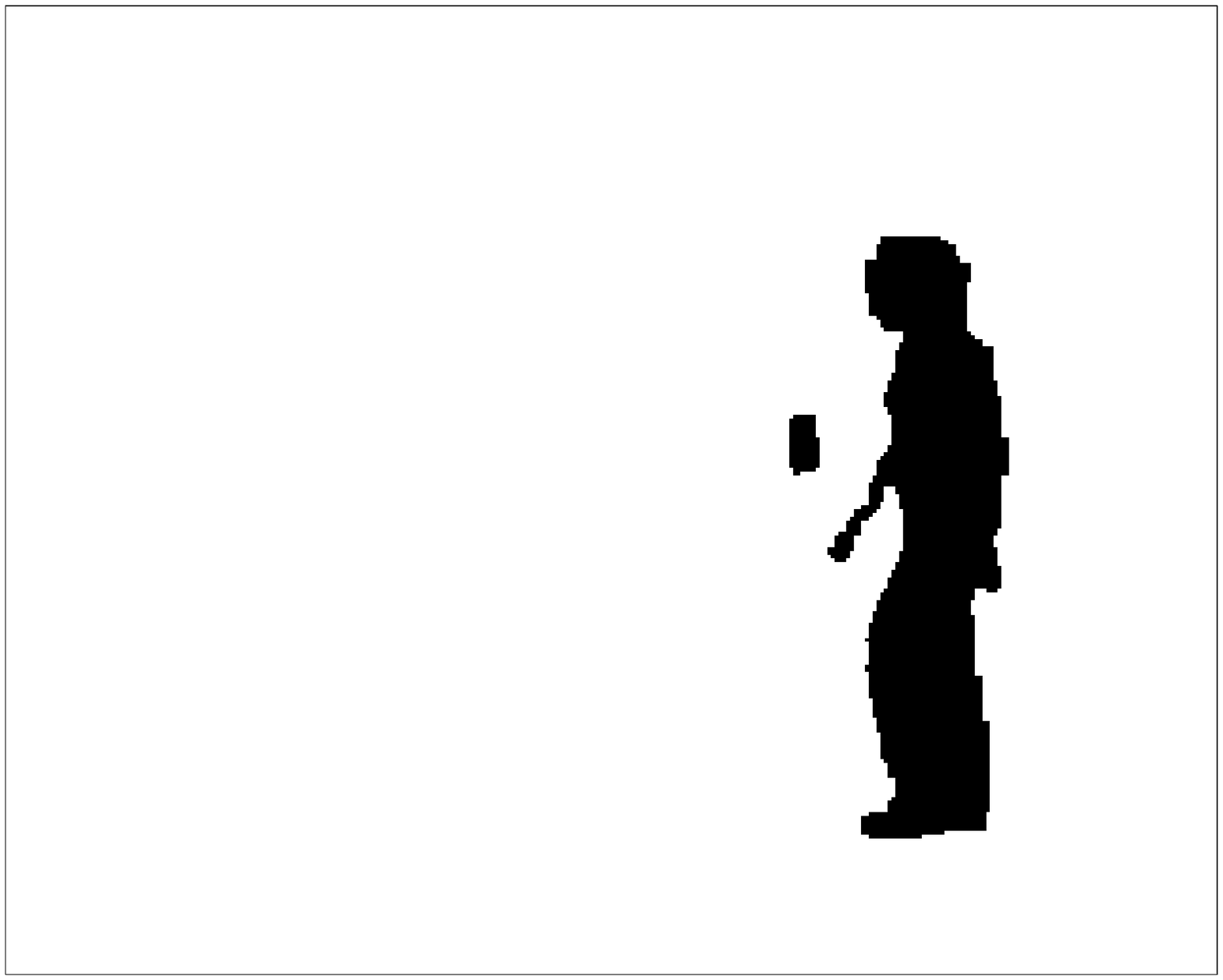} \\ \hline
\epsfxsize=1in\epsfysize=0.67in\epsfbox{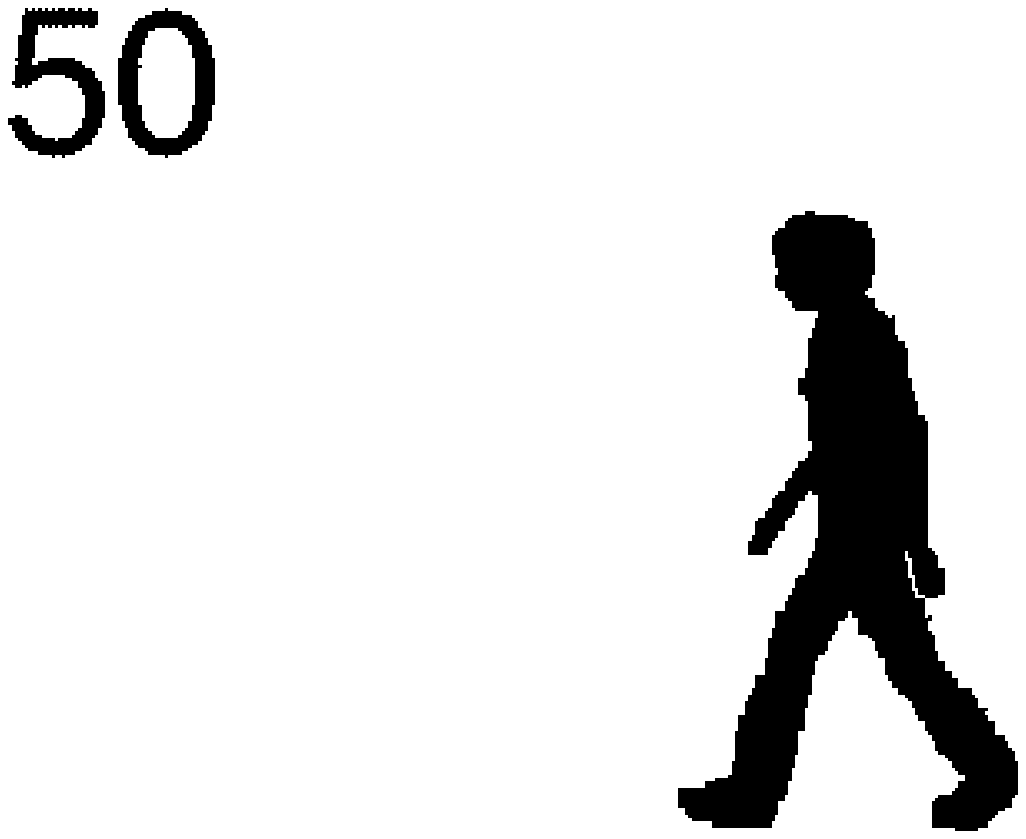} &
\epsfxsize=1in\epsfysize=0.67in\epsfbox{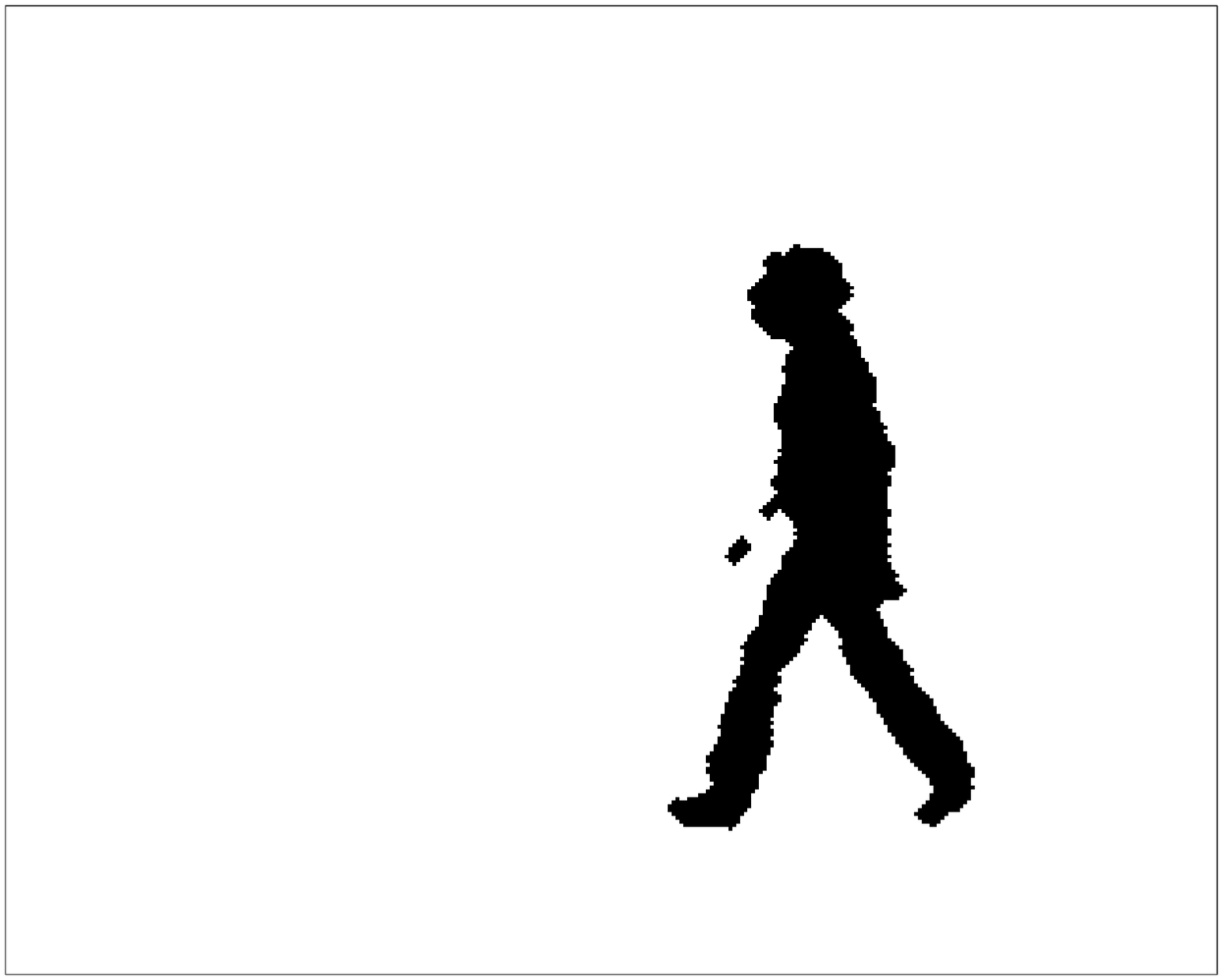} &
\epsfxsize=1in\epsfysize=0.67in\epsfbox{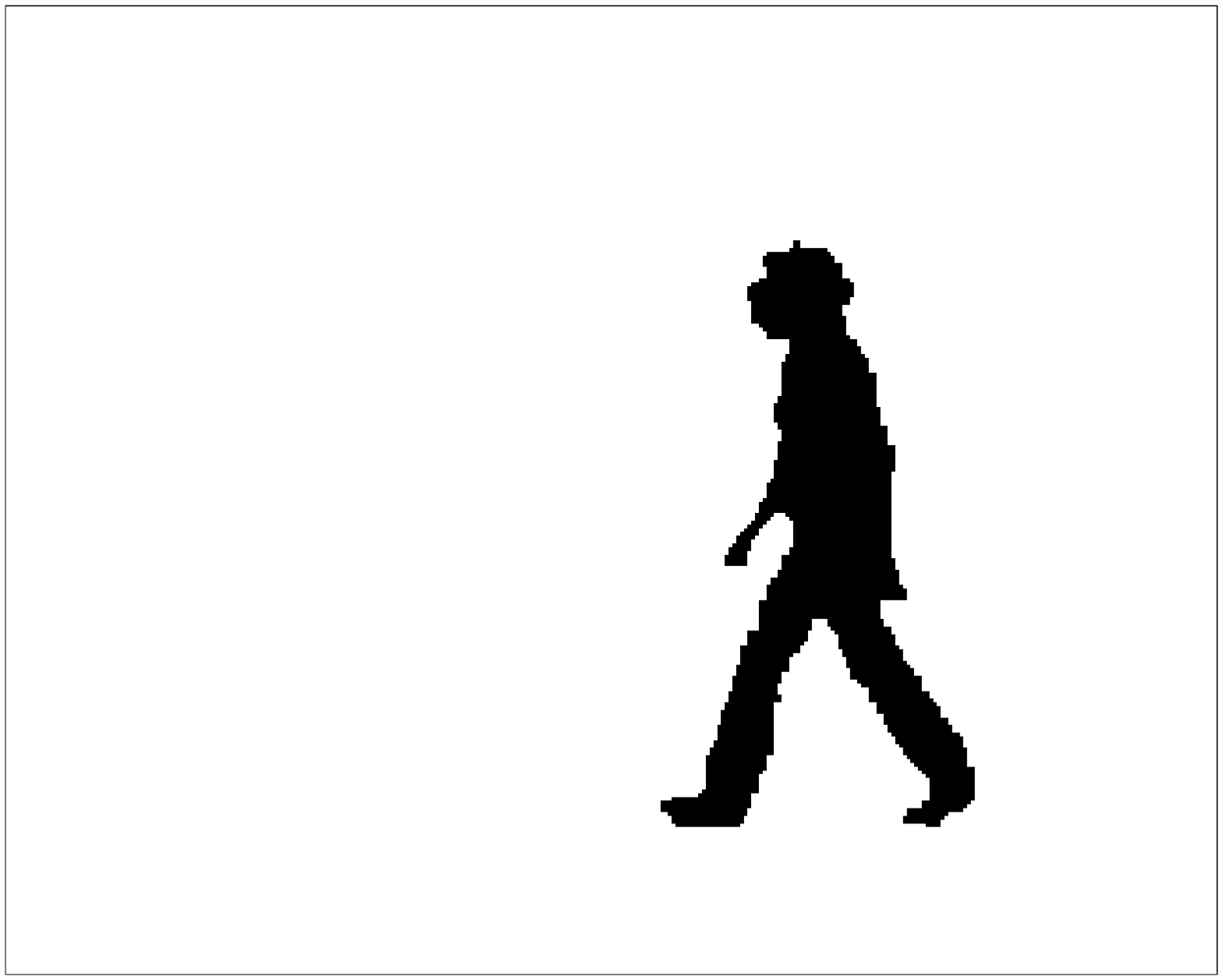} \\ \hline
\epsfxsize=1in\epsfysize=0.67in\epsfbox{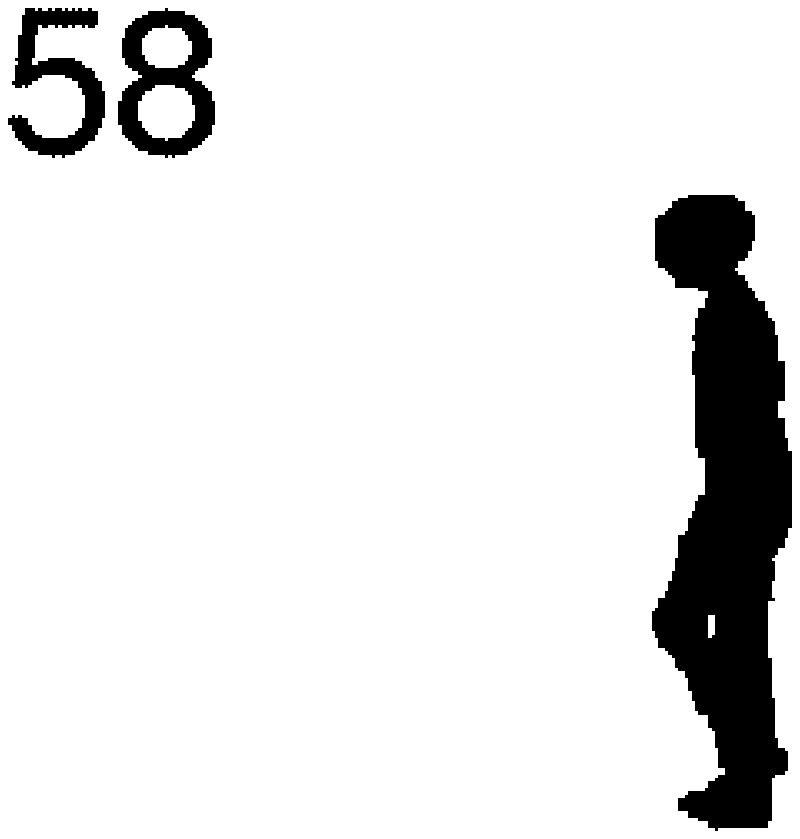} &
\epsfxsize=1in\epsfysize=0.67in\epsfbox{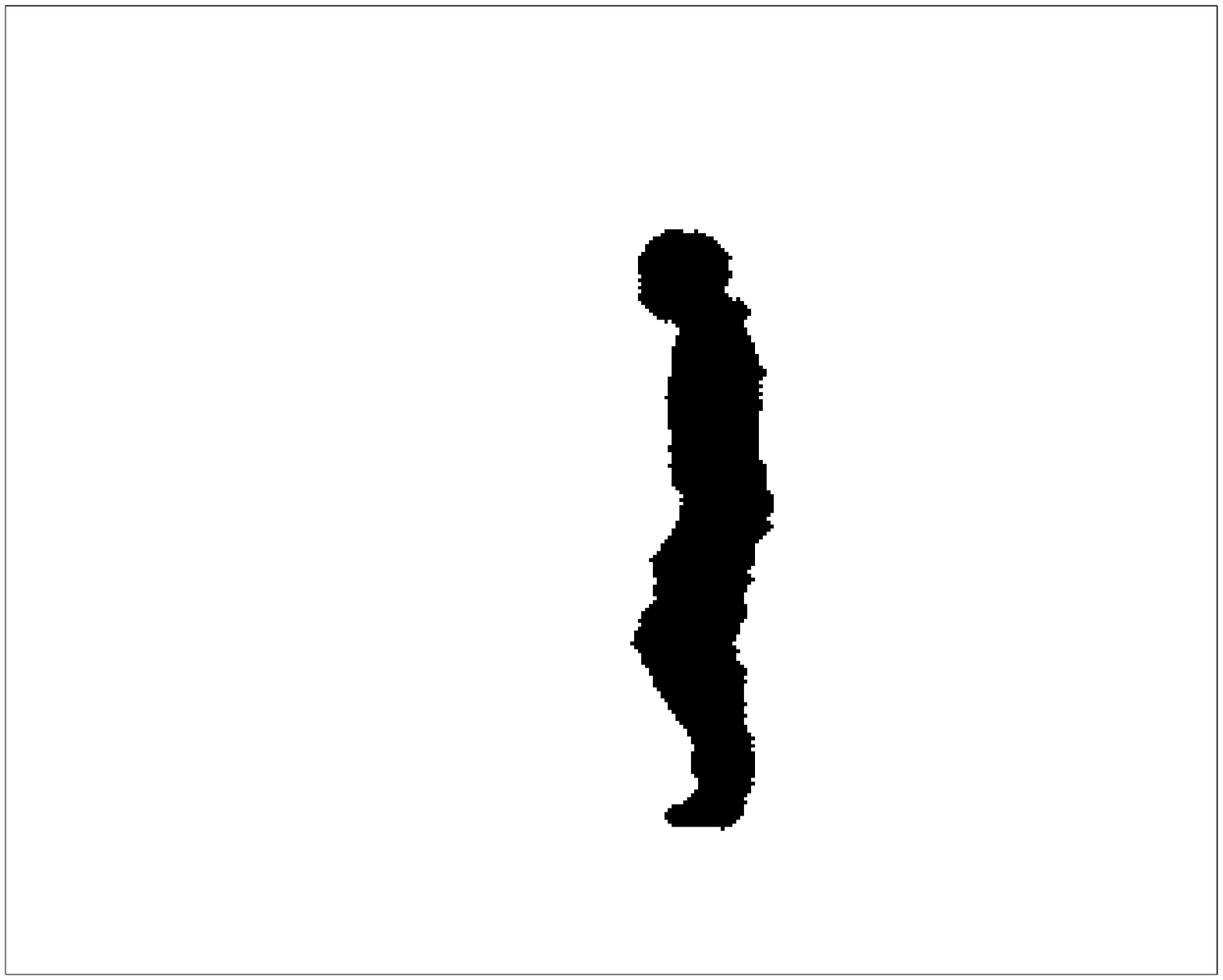} &
\epsfxsize=1in\epsfysize=0.67in\epsfbox{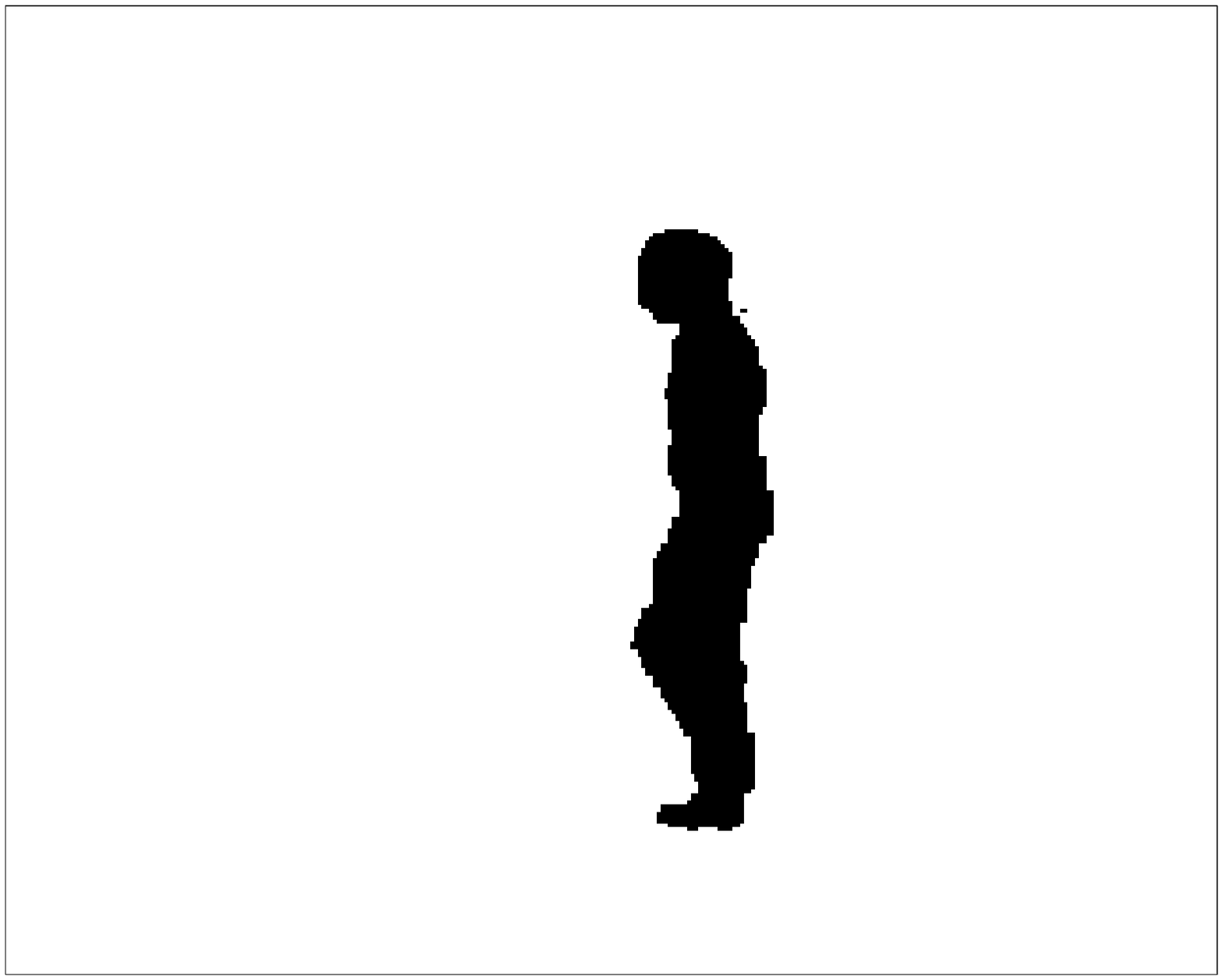} \\ \hline
\epsfxsize=1in\epsfysize=0.67in\epsfbox{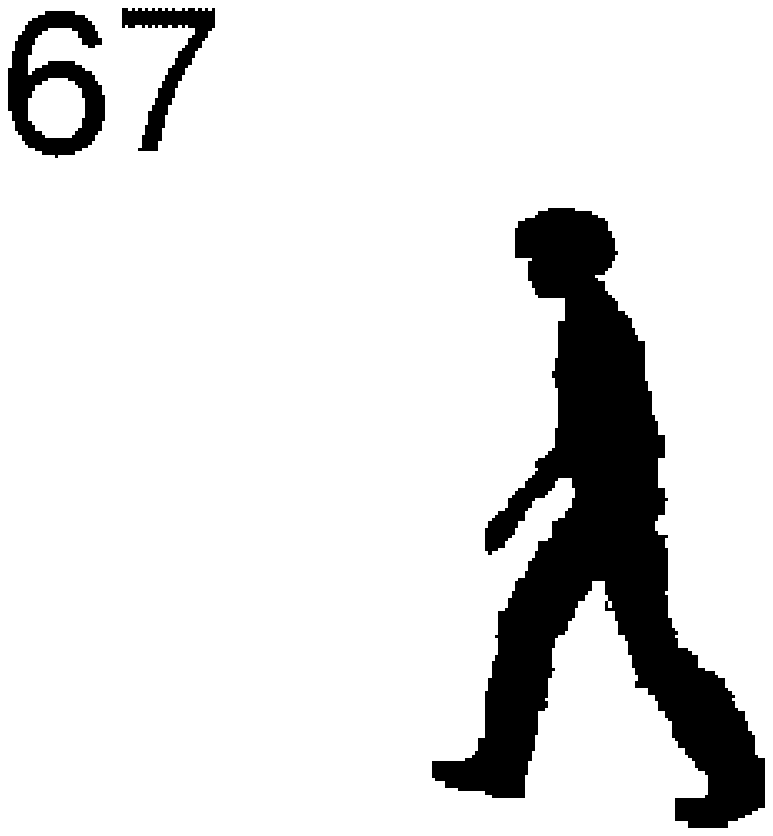} &
\epsfxsize=1in\epsfysize=0.67in\epsfbox{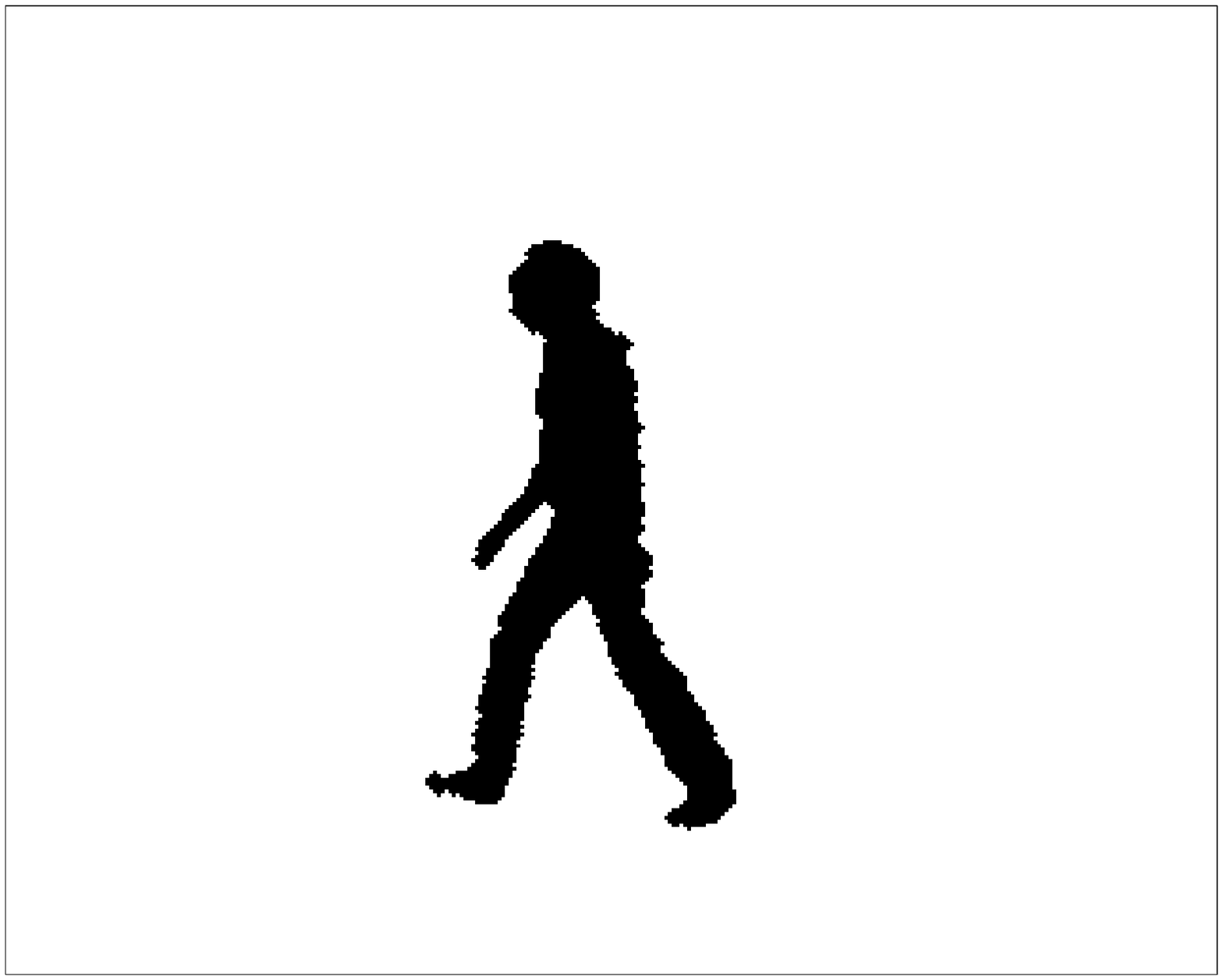} &
\epsfxsize=1in\epsfysize=0.67in\epsfbox{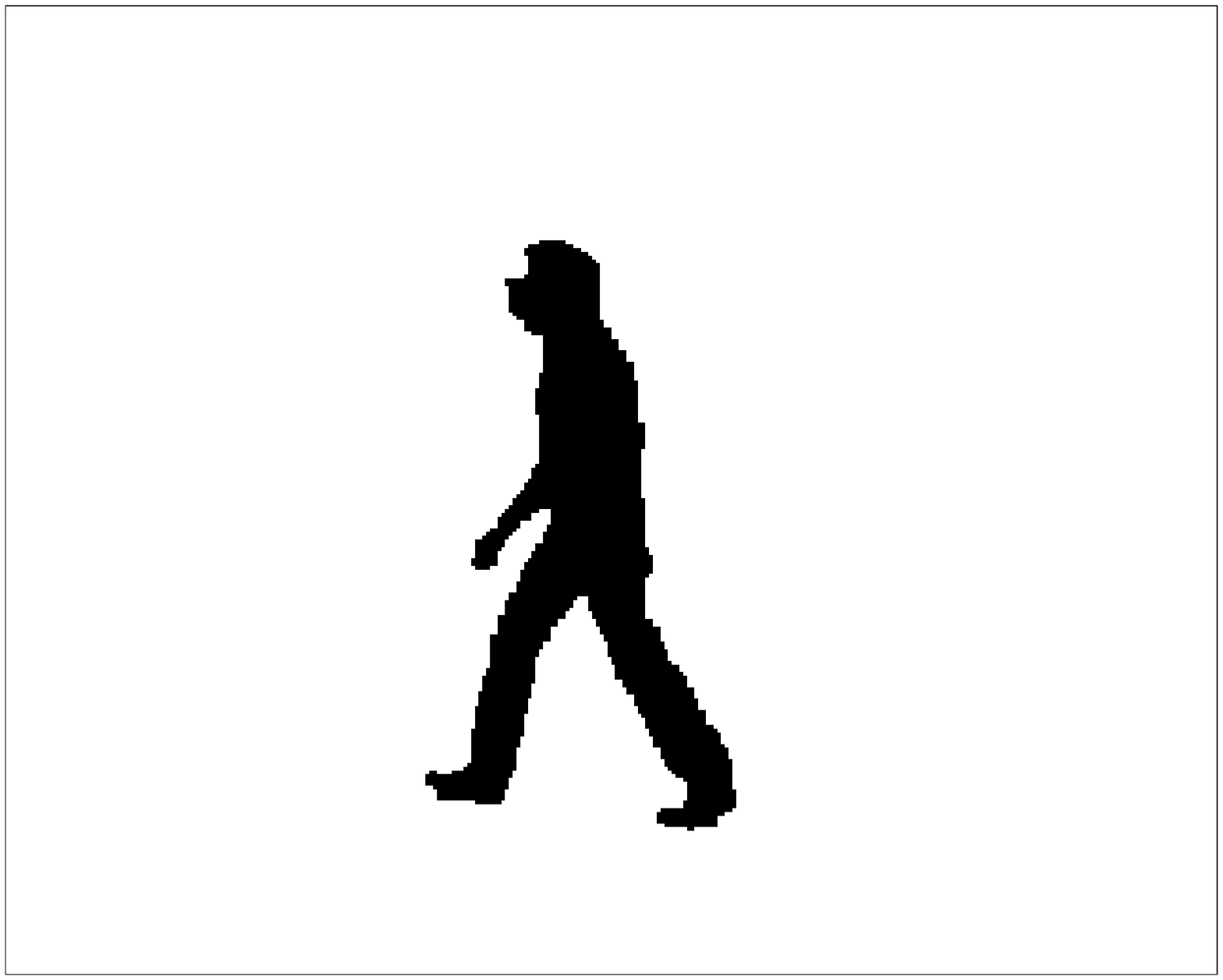} \\ \hline
\epsfxsize=1in\epsfysize=0.67in\epsfbox{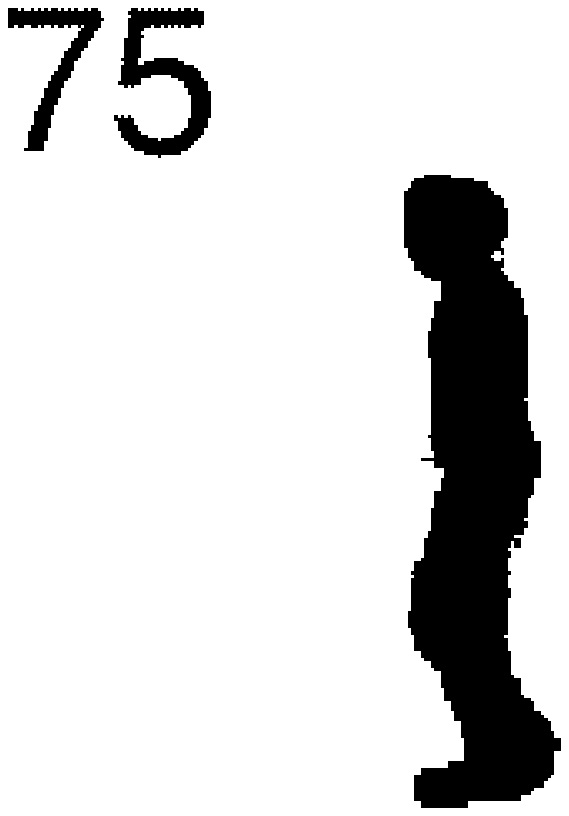} &
\epsfxsize=1in\epsfysize=0.67in\epsfbox{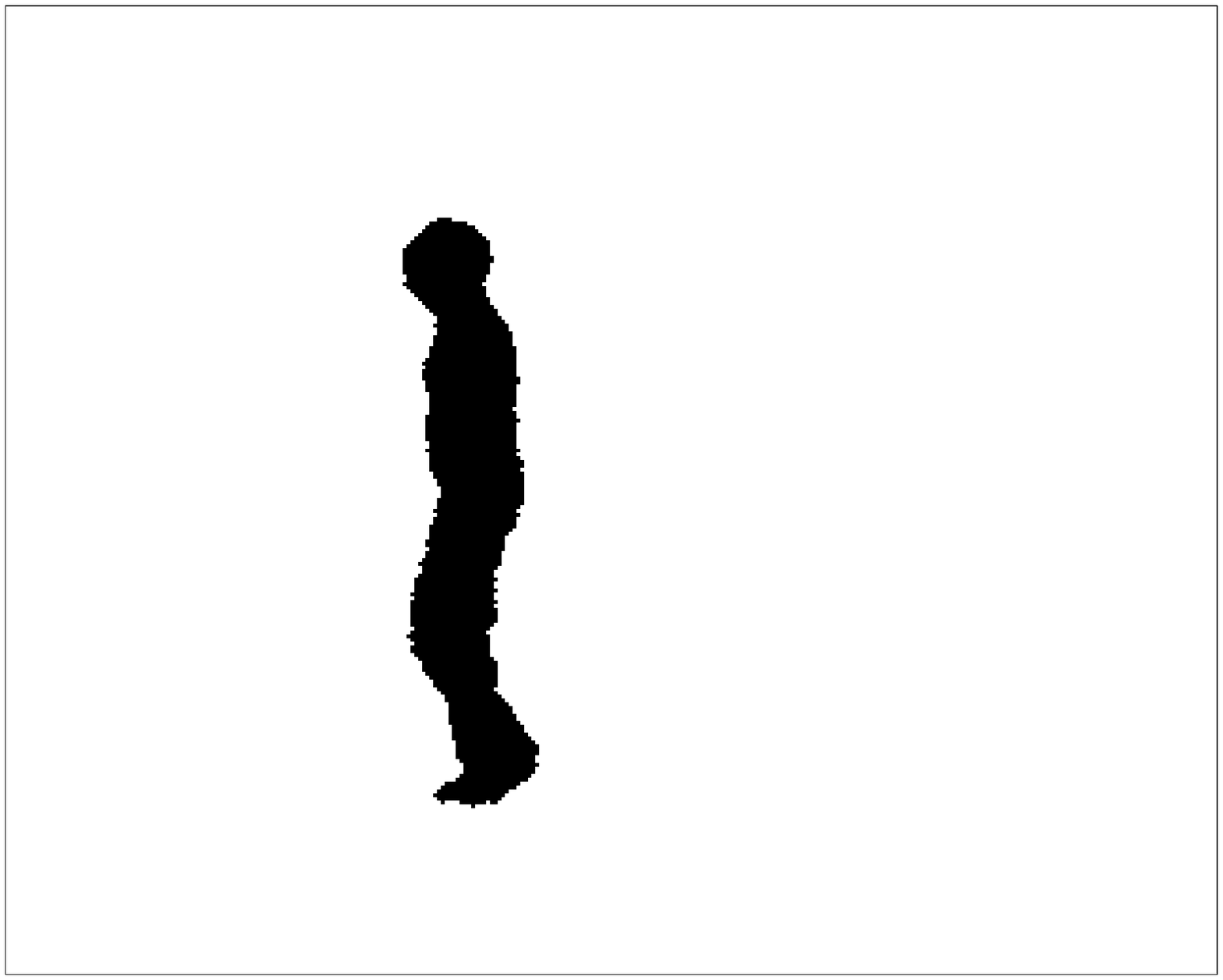} &
\epsfxsize=1in\epsfysize=0.67in\epsfbox{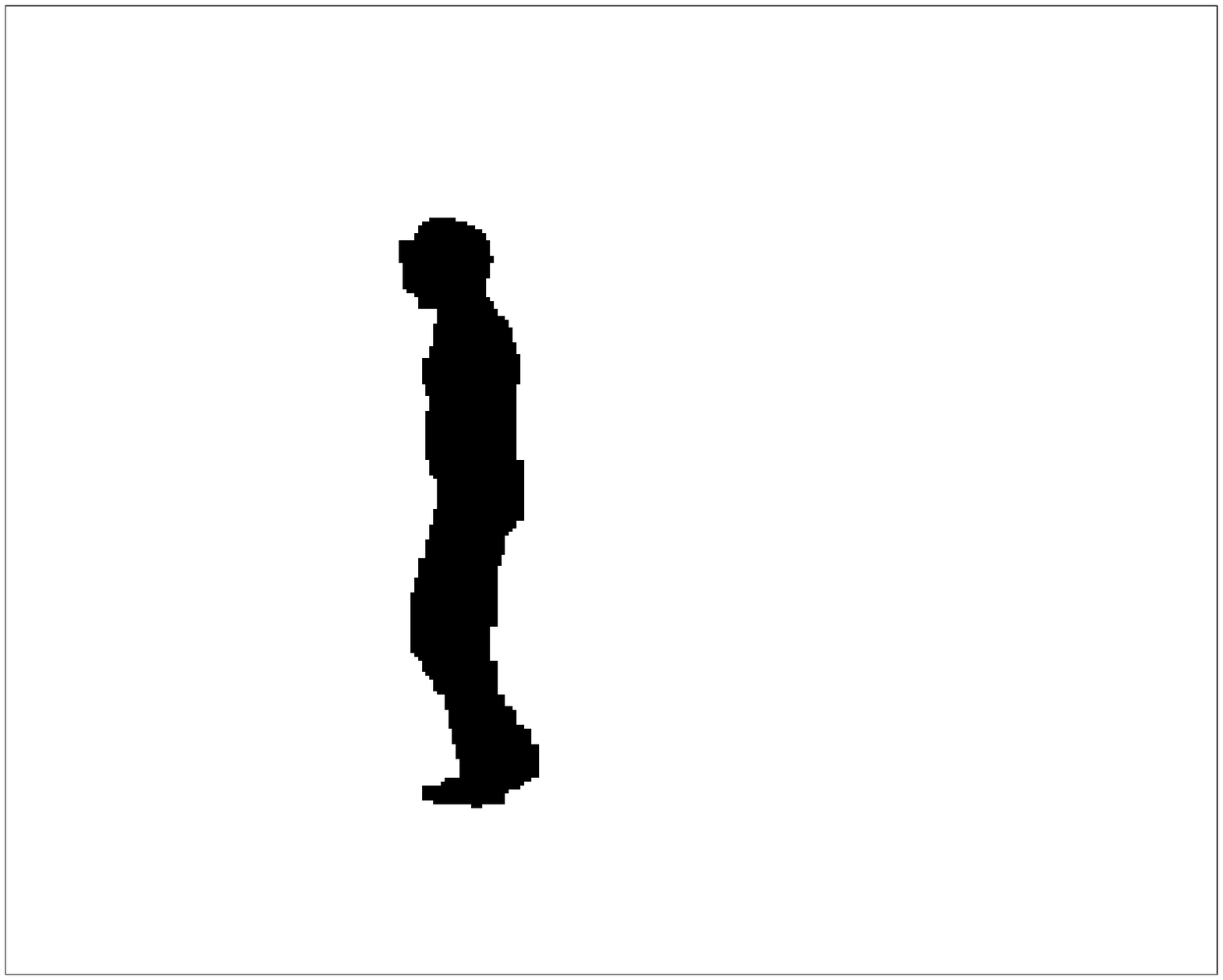} \\ \hline
\epsfxsize=1in\epsfysize=0.67in\epsfbox{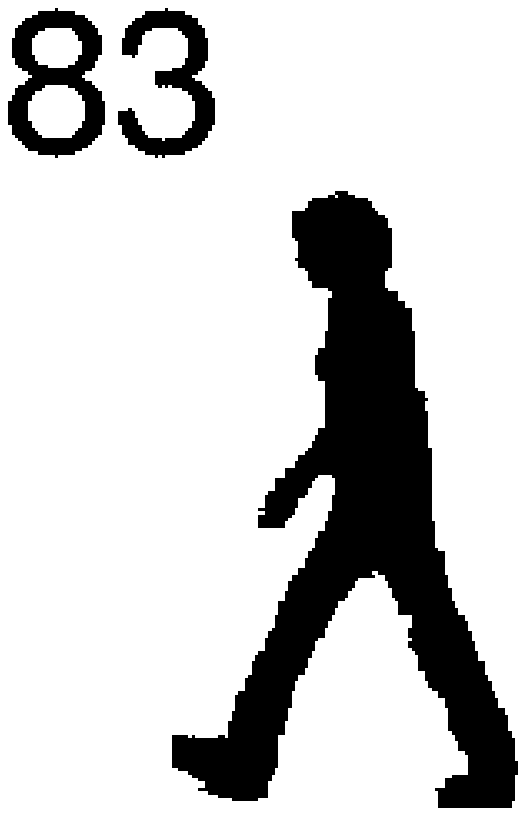} &
\epsfxsize=1in\epsfysize=0.67in\epsfbox{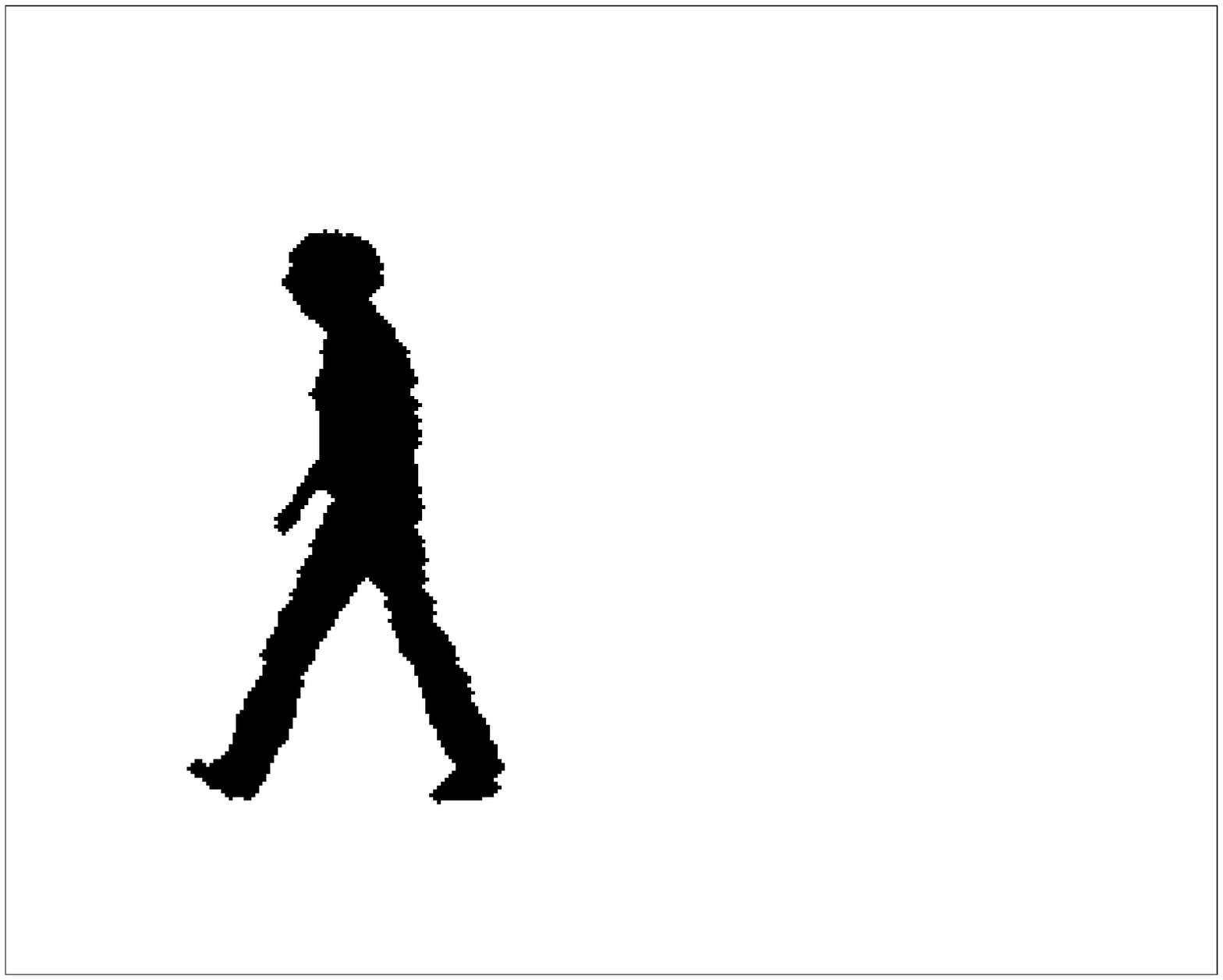} &
\epsfxsize=1in\epsfysize=0.67in\epsfbox{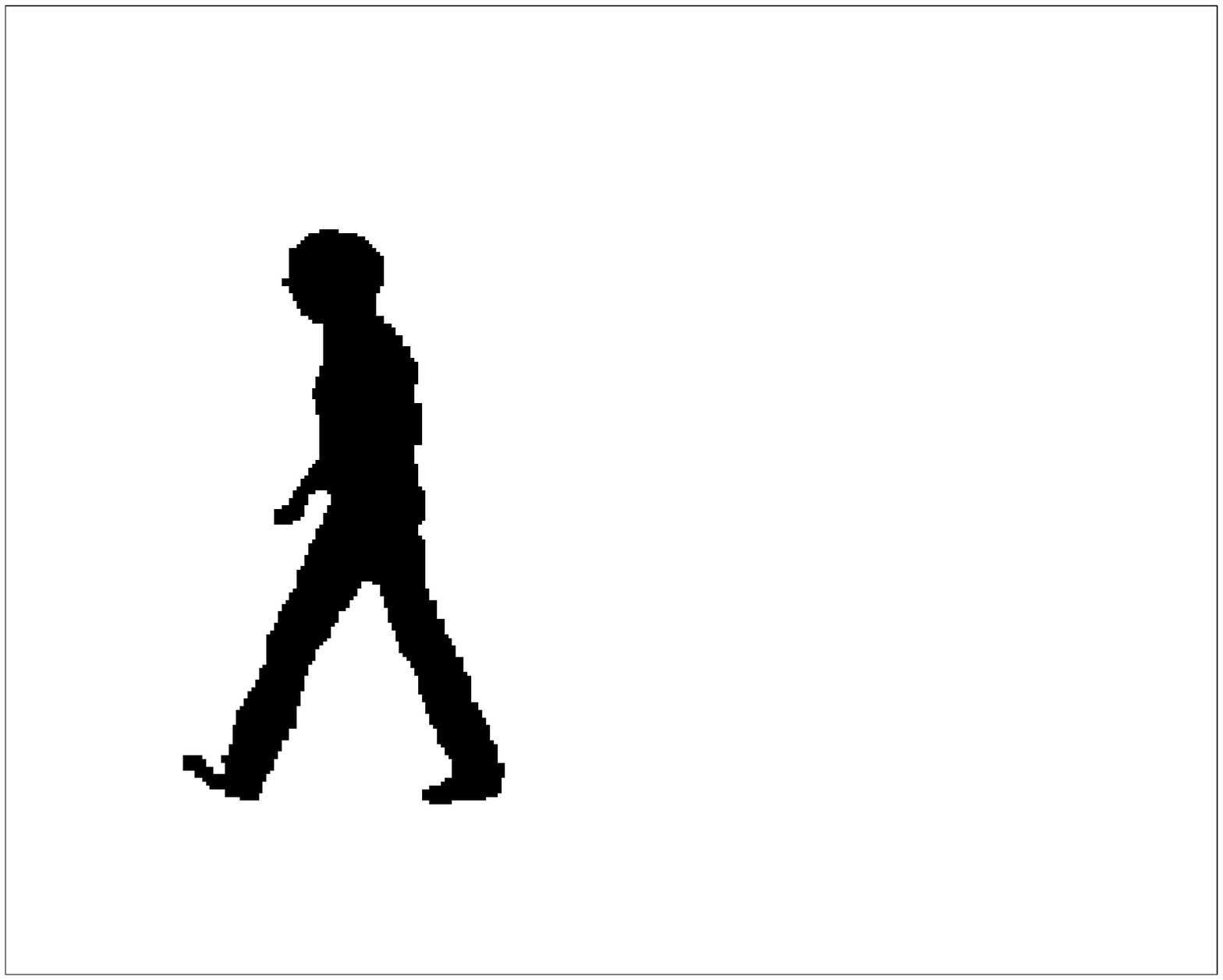} \\ \hline
\epsfxsize=1in\epsfysize=0.67in\epsfbox{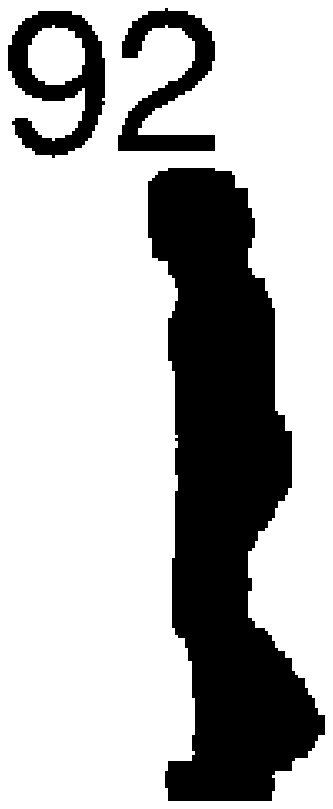} &
\epsfxsize=1in\epsfysize=0.67in\epsfbox{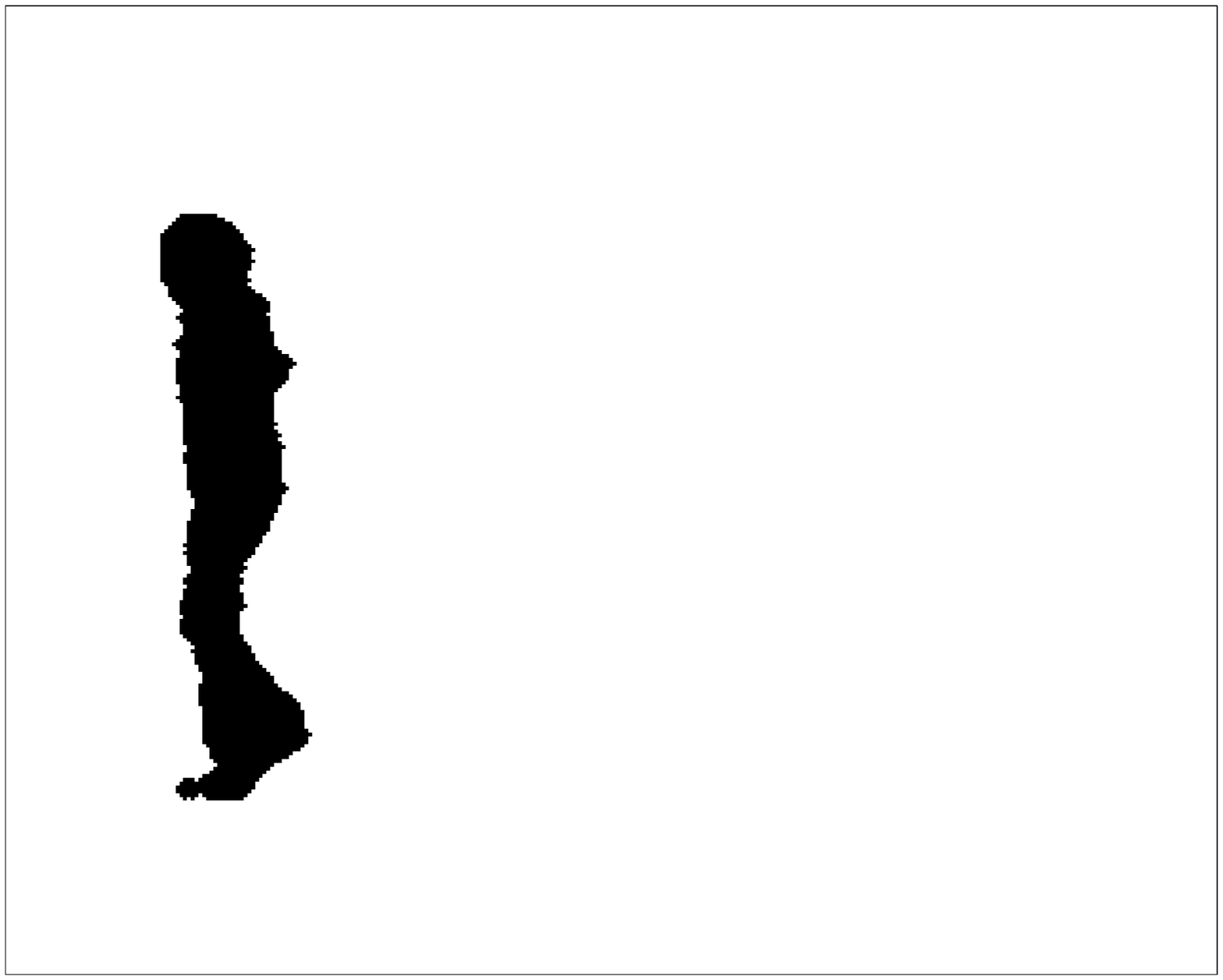} &
\epsfxsize=1in\epsfysize=0.67in\epsfbox{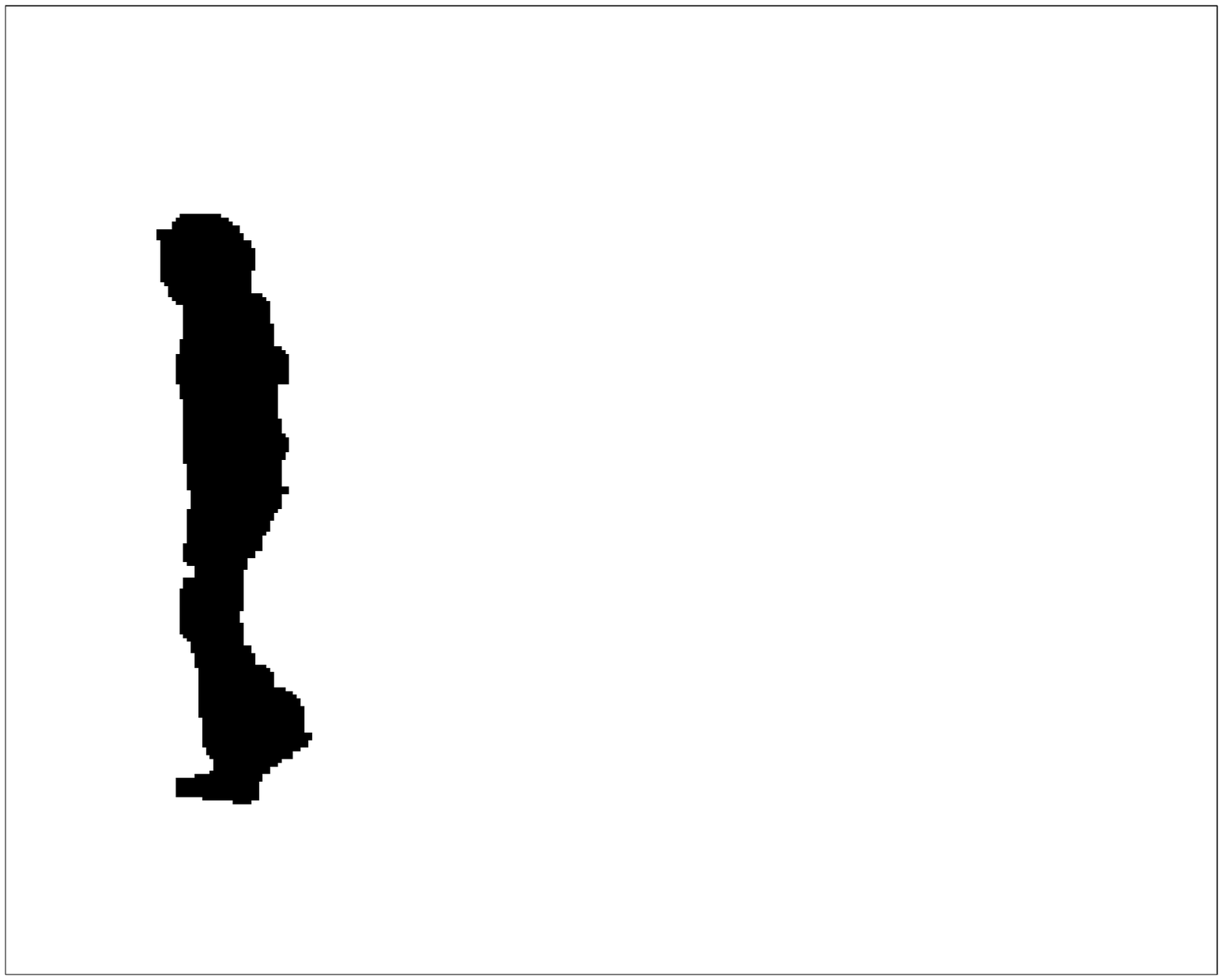} \\ \hline
\epsfxsize=1in\epsfysize=0.67in\epsfbox{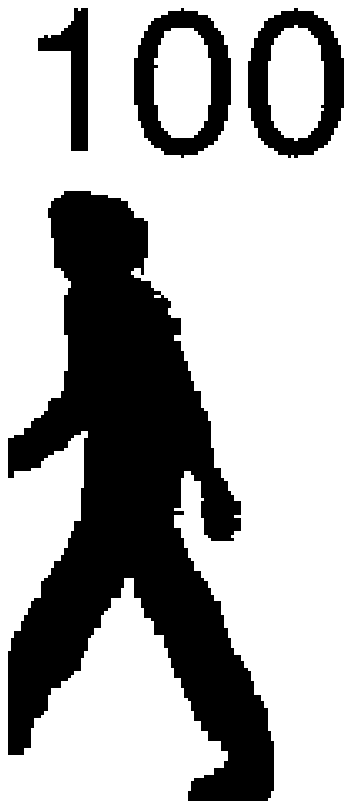} &
\epsfxsize=1in\epsfysize=0.67in\epsfbox{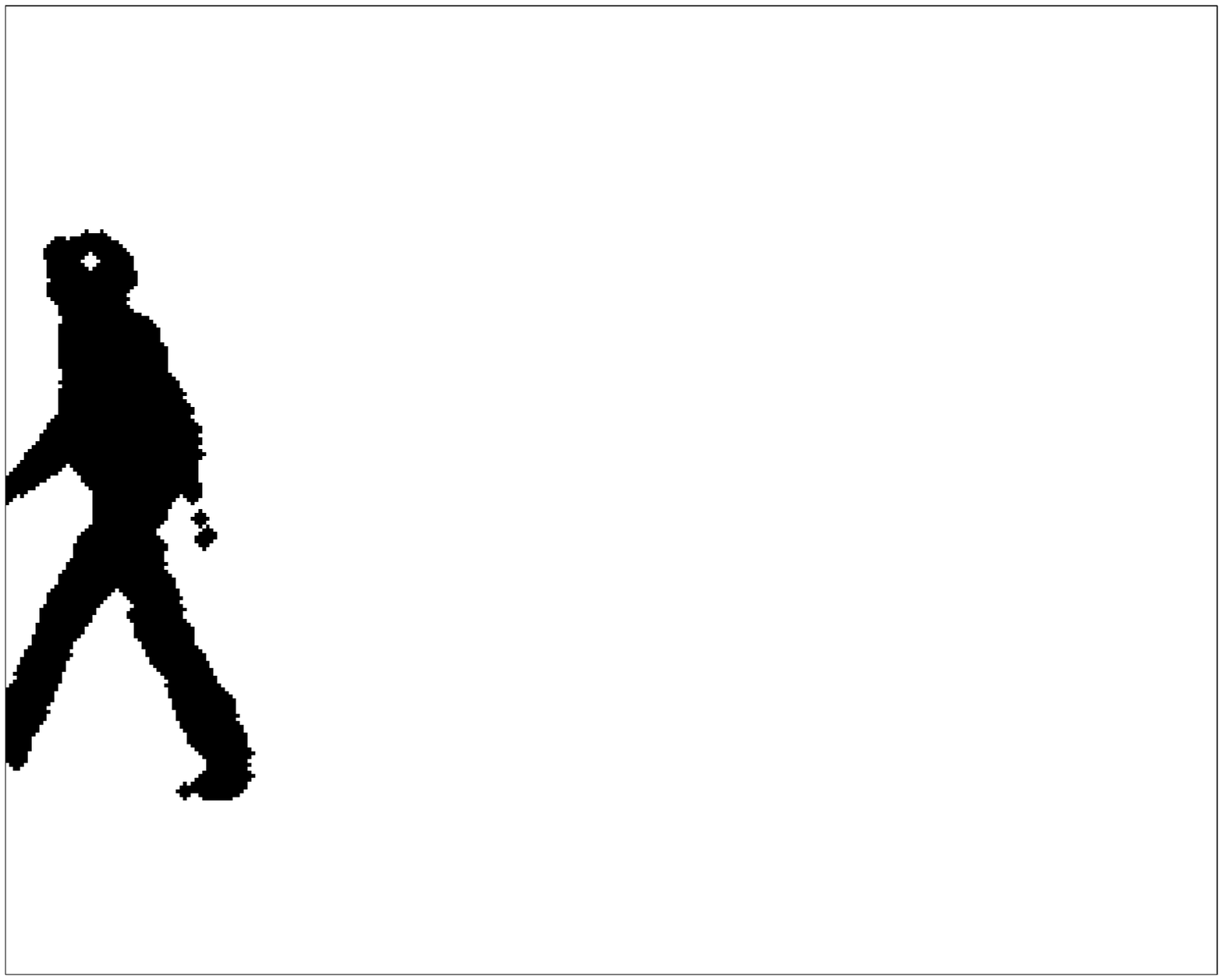} &
\epsfxsize=1in\epsfysize=0.67in\epsfbox{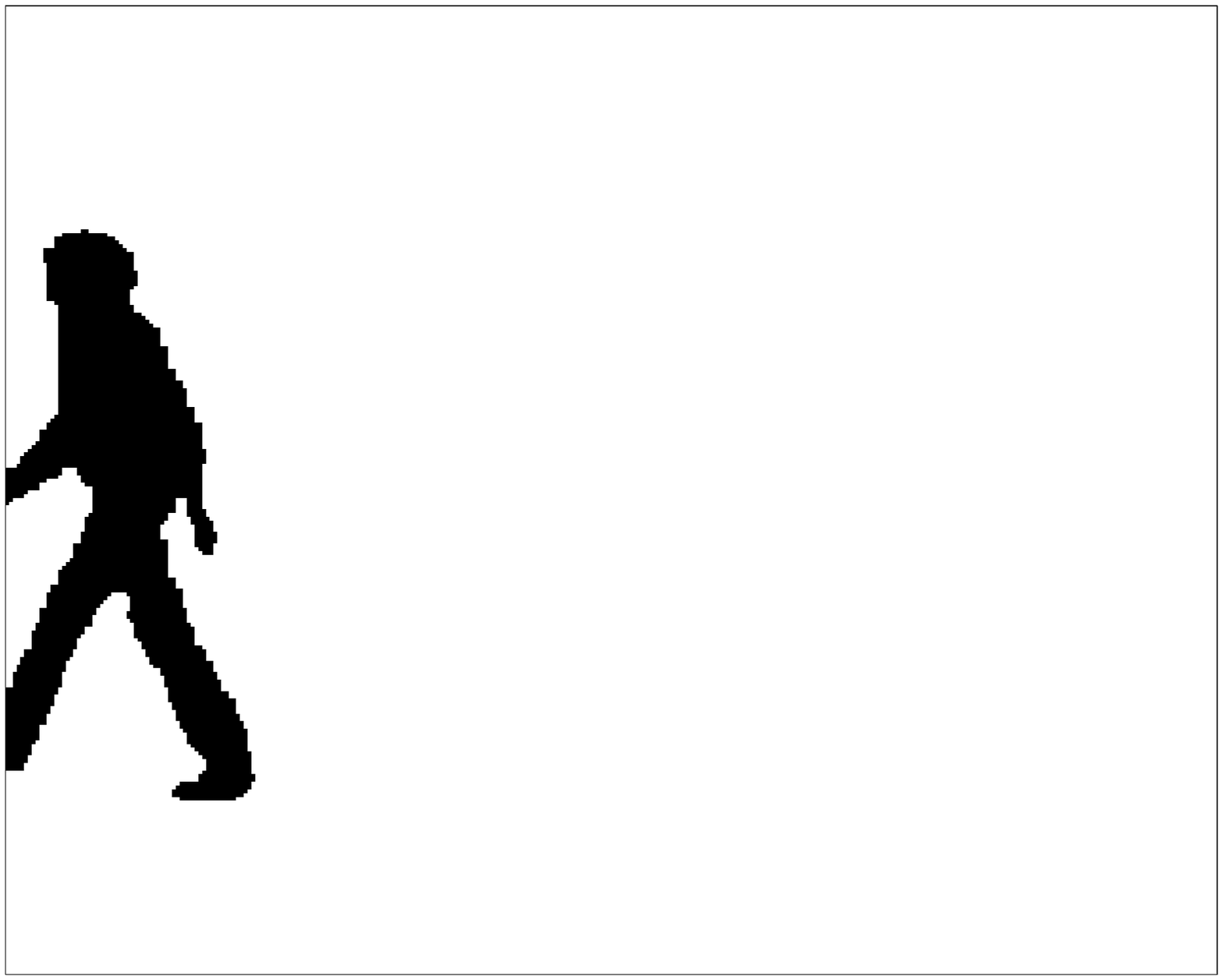} \\ \hline
(Truth) & (Morph.) & (Graph) \\ \hline
\end{tabular}
\end{center}
\caption{A sequence of frames from the {\em Outdoor} clip.  The extra
spot in earlier time frames is a reflection.}
\label{fig-hedvigseq}
\end{figure}

\begin{figure}
\begin{center}
\begin{tabular}{|@{}c@{}|@{}c@{}|@{}c@{}|} \hline
\epsfxsize=1in\epsfysize=0.67in\epsfbox{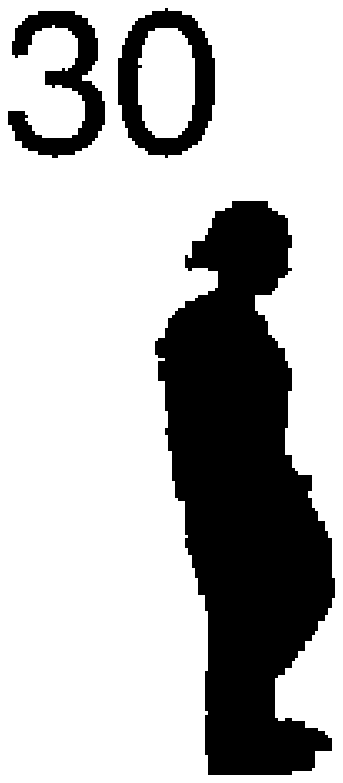} &
\epsfxsize=1in\epsfysize=0.67in\epsfbox{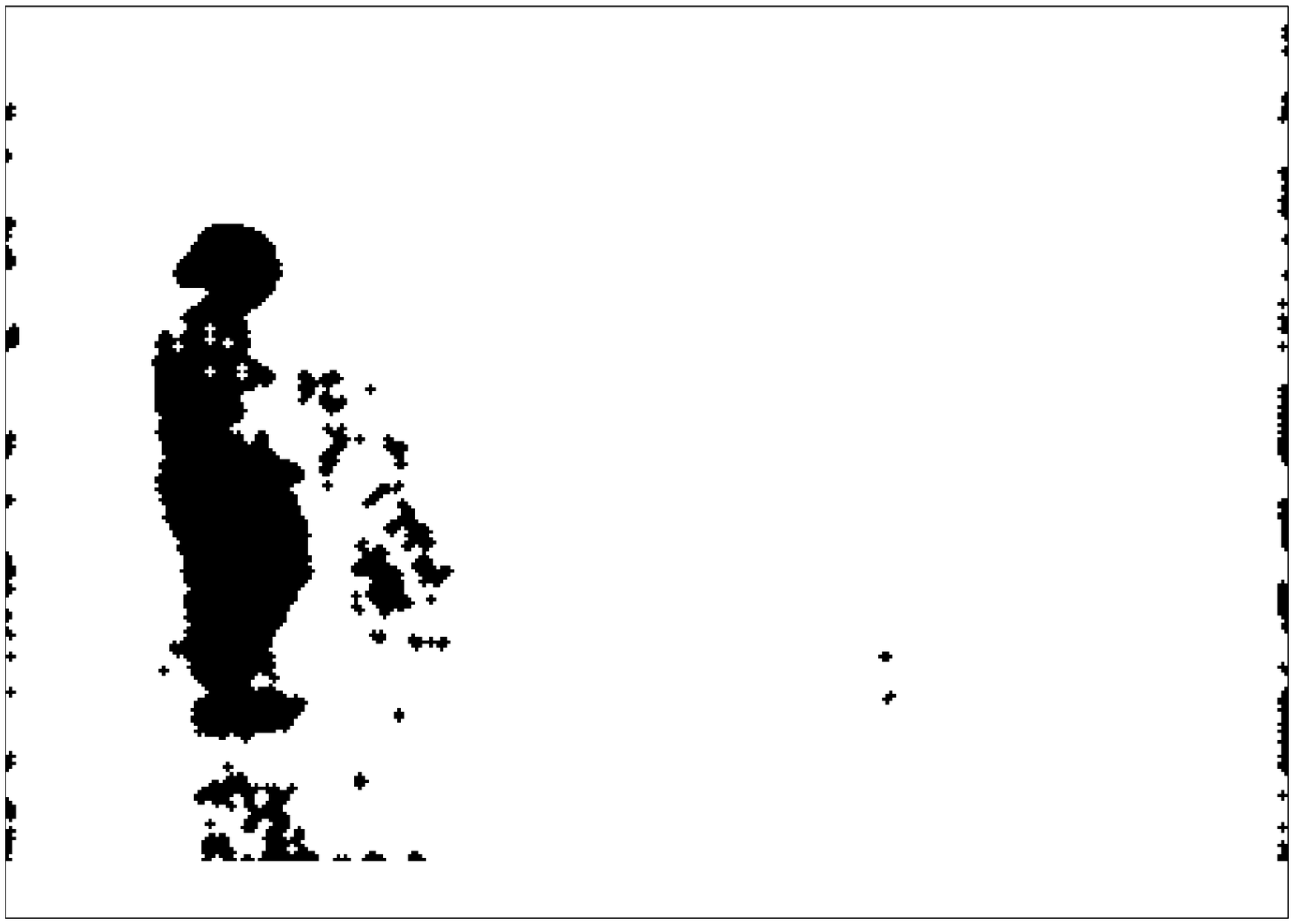} &
\epsfxsize=1in\epsfysize=0.67in\epsfbox{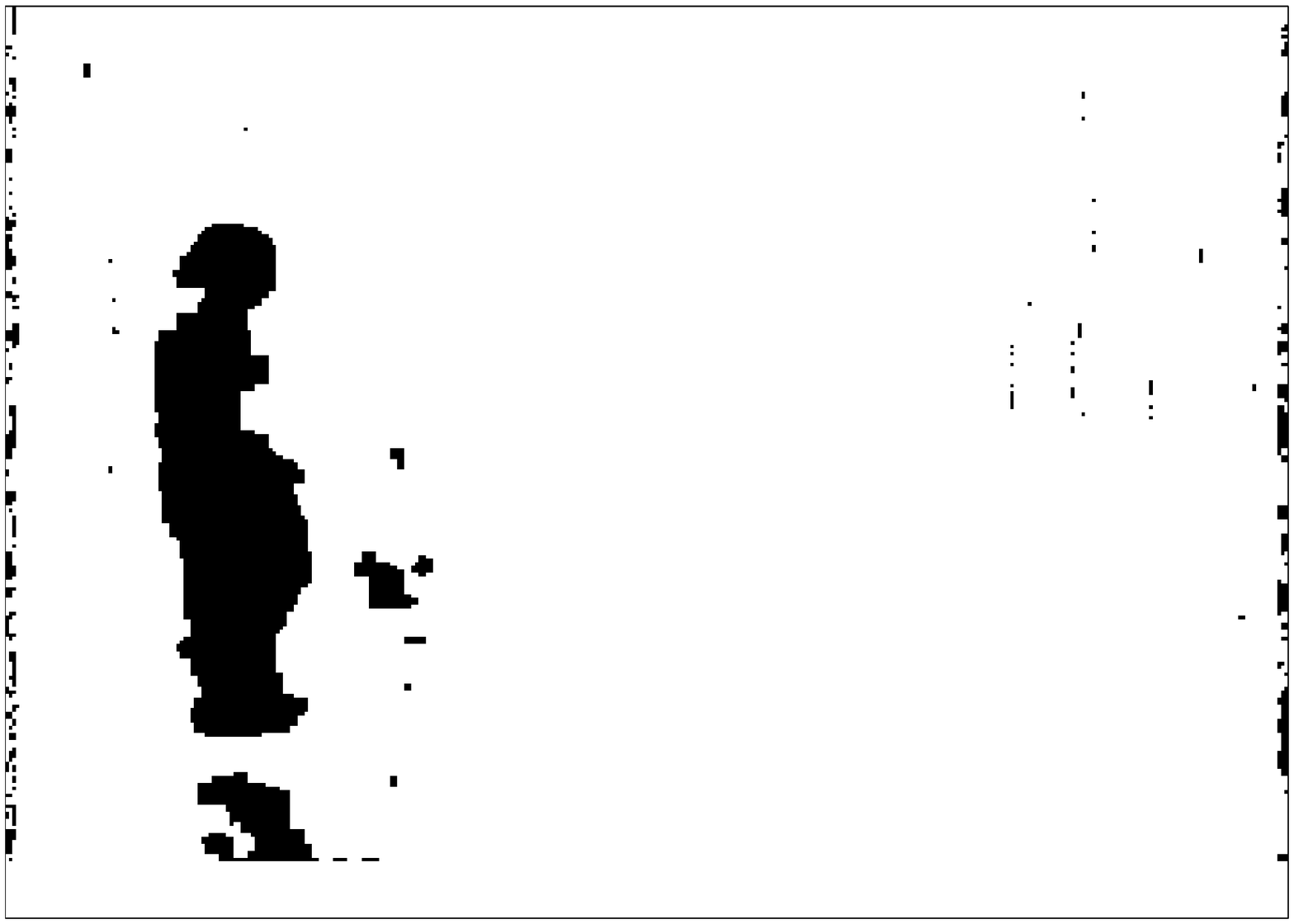} \\ \hline
\epsfxsize=1in\epsfysize=0.67in\epsfbox{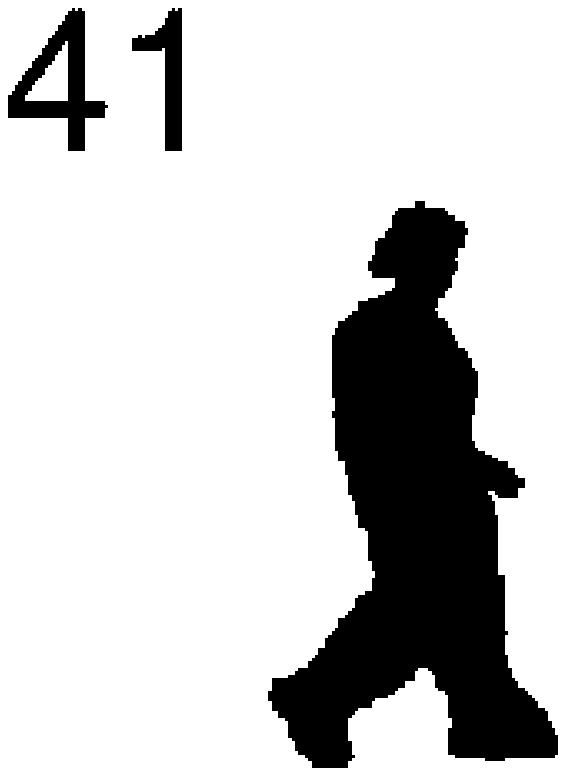} &
\epsfxsize=1in\epsfysize=0.67in\epsfbox{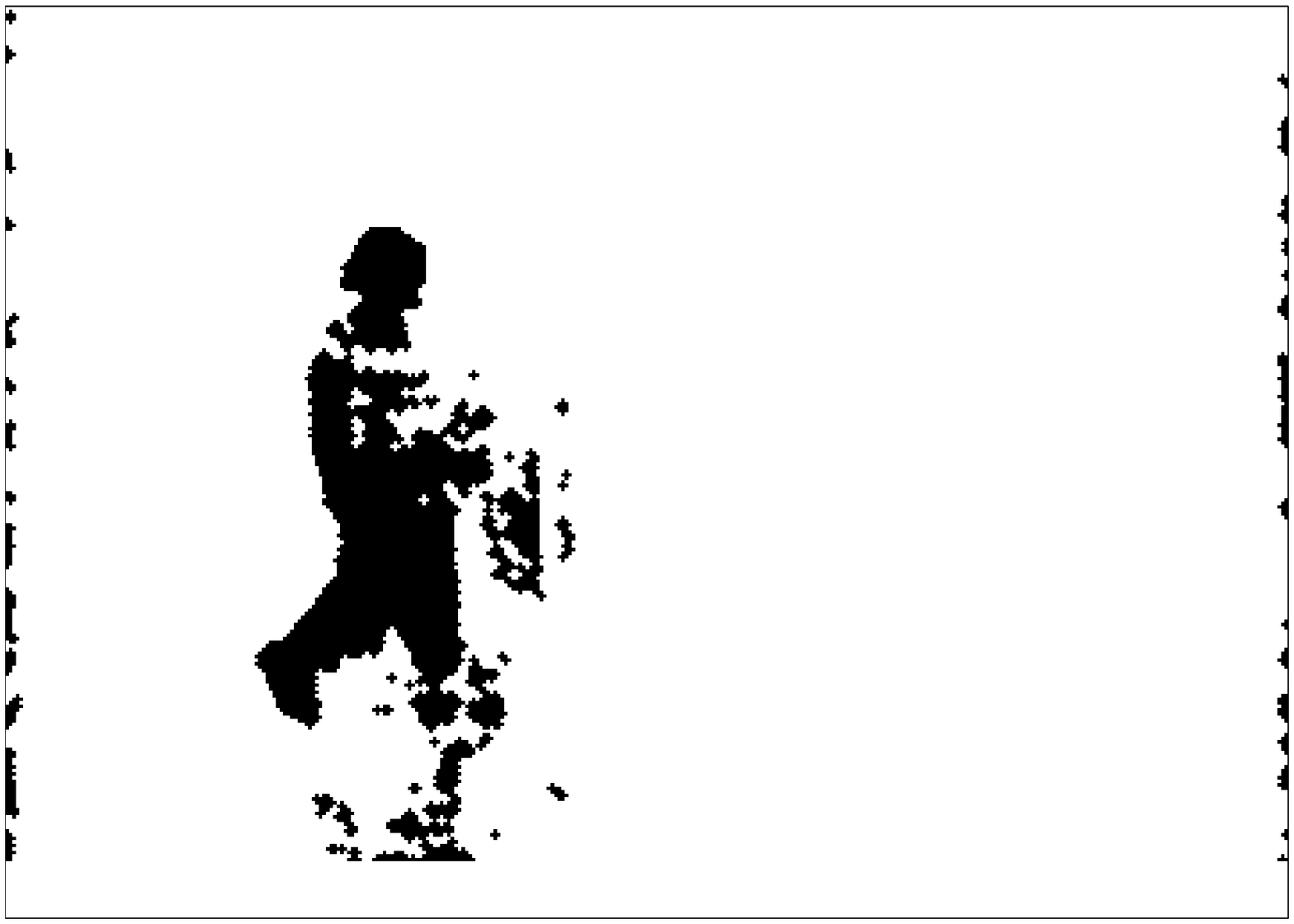} &
\epsfxsize=1in\epsfysize=0.67in\epsfbox{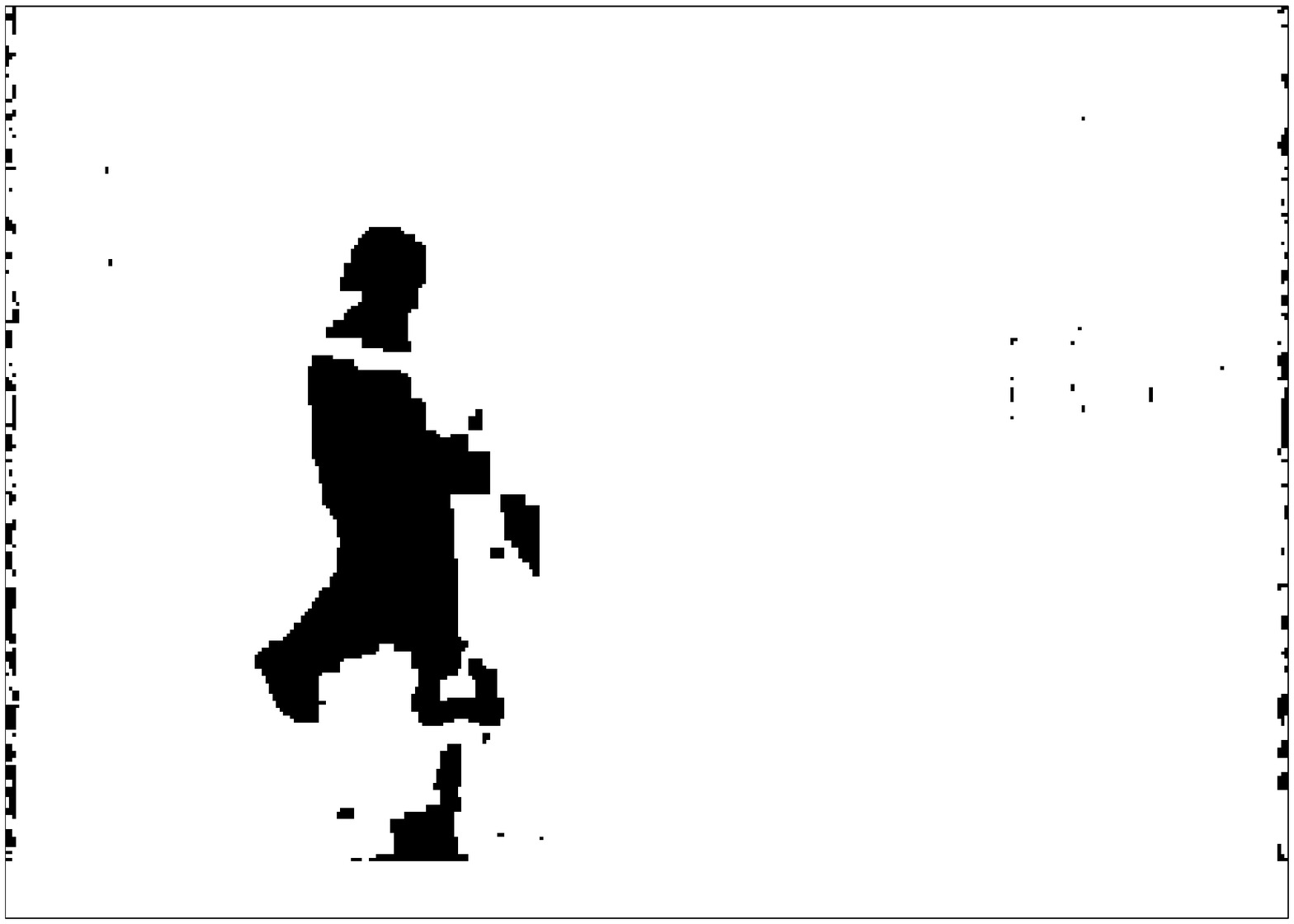} \\ \hline
\epsfxsize=1in\epsfysize=0.67in\epsfbox{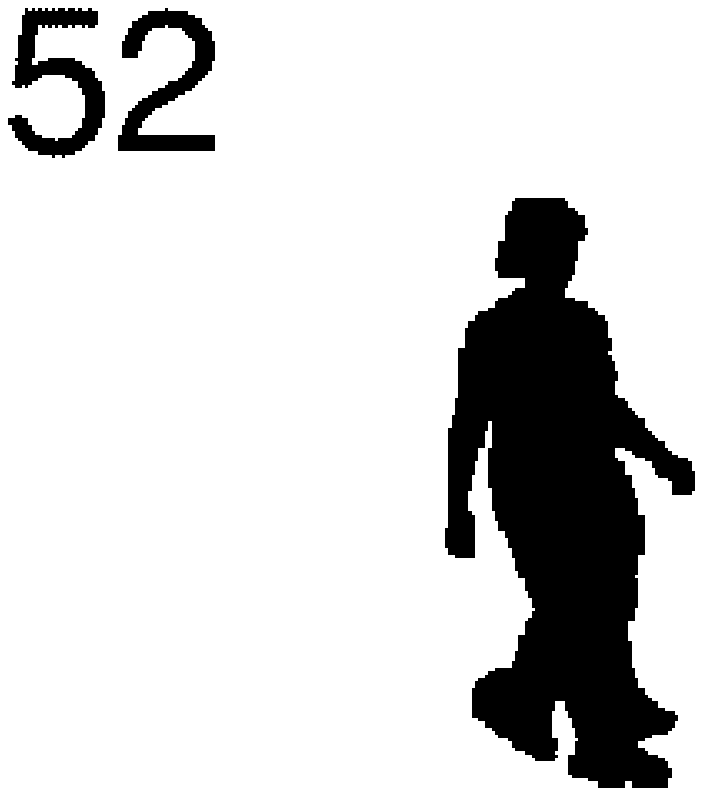} &
\epsfxsize=1in\epsfysize=0.67in\epsfbox{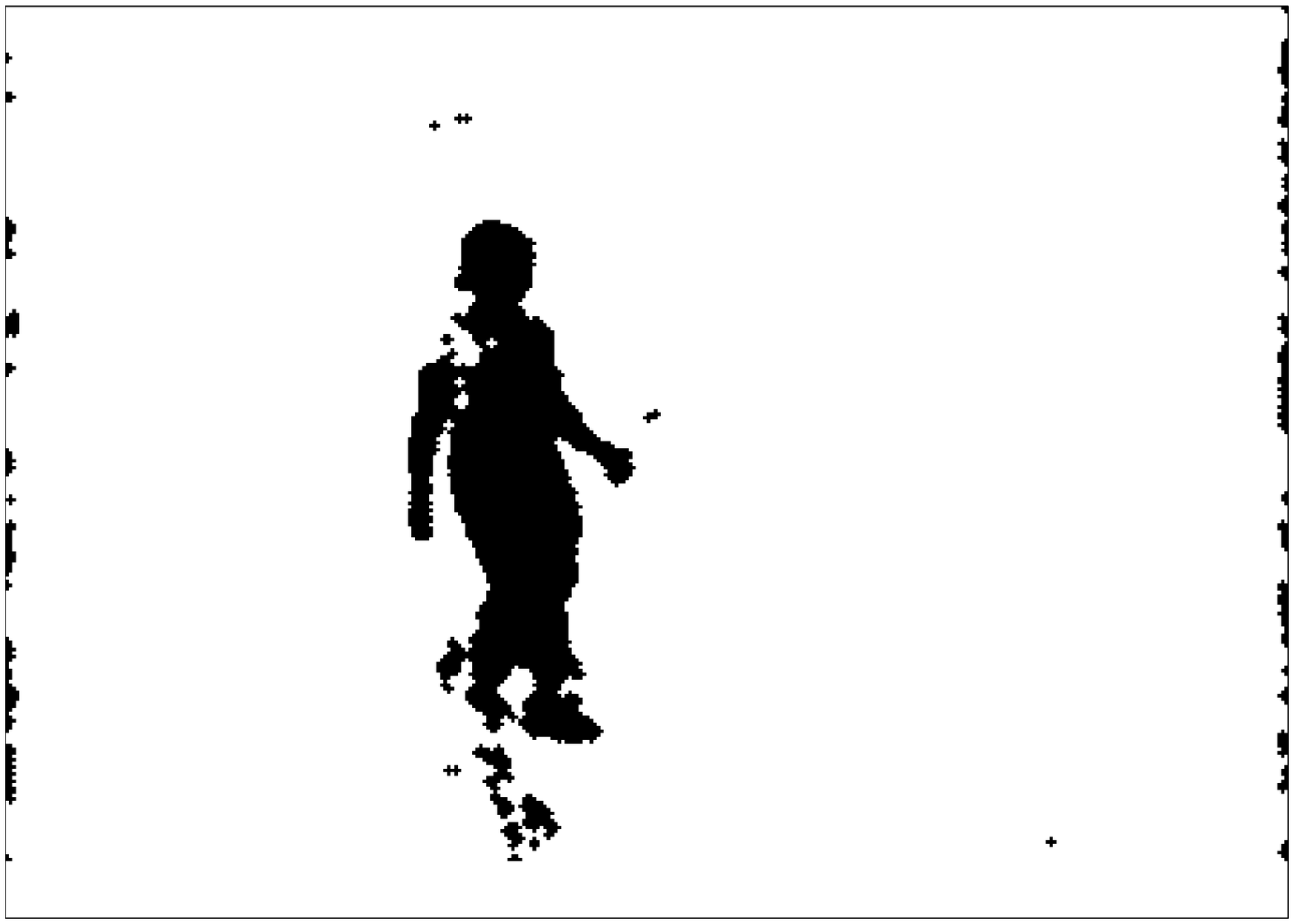} &
\epsfxsize=1in\epsfysize=0.67in\epsfbox{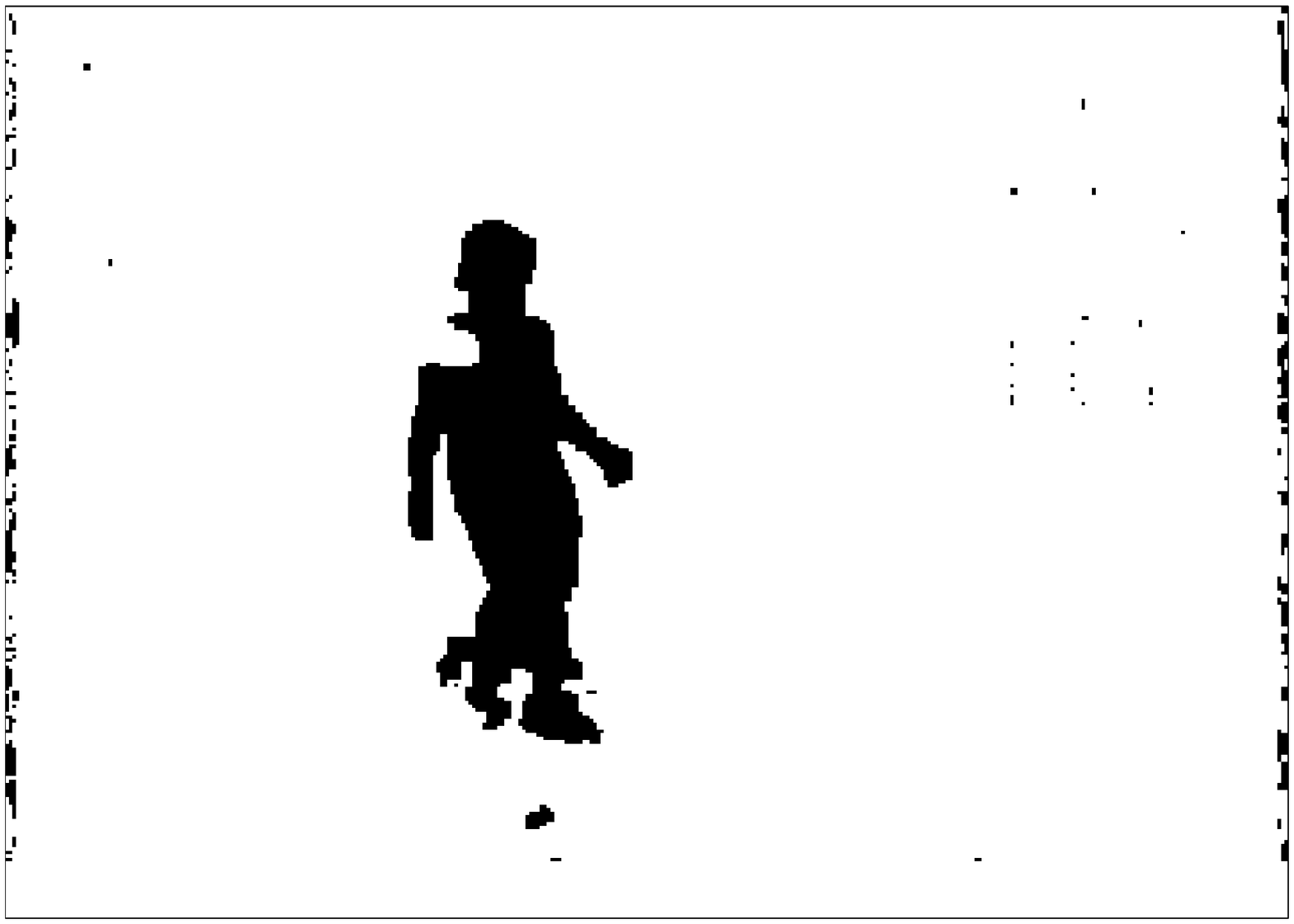} \\ \hline
\epsfxsize=1in\epsfysize=0.67in\epsfbox{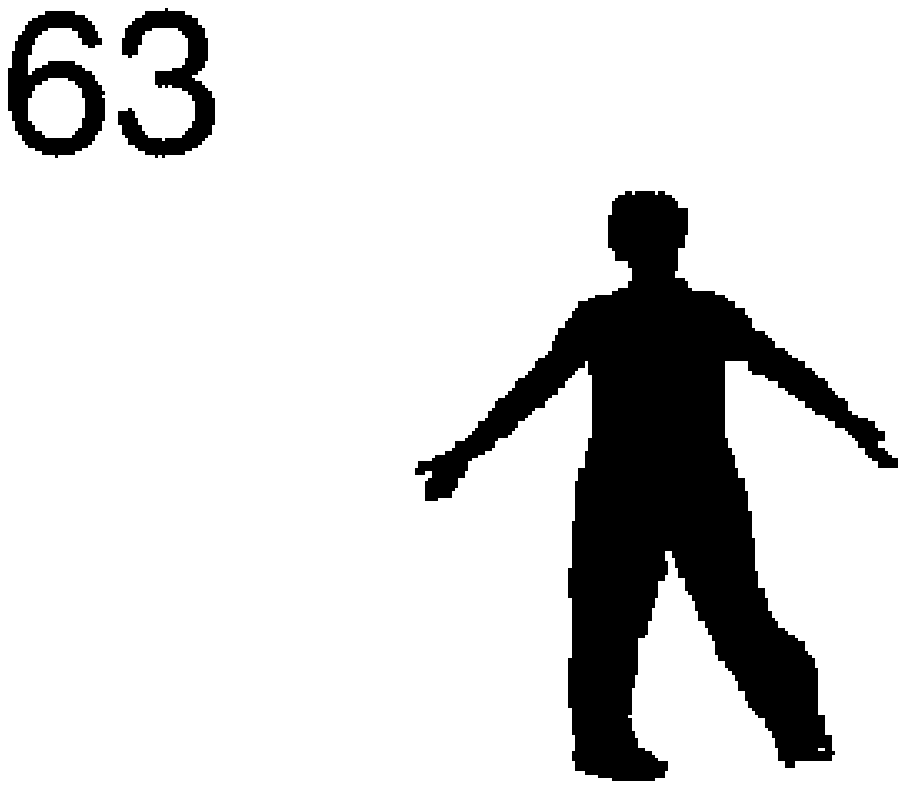} &
\epsfxsize=1in\epsfysize=0.67in\epsfbox{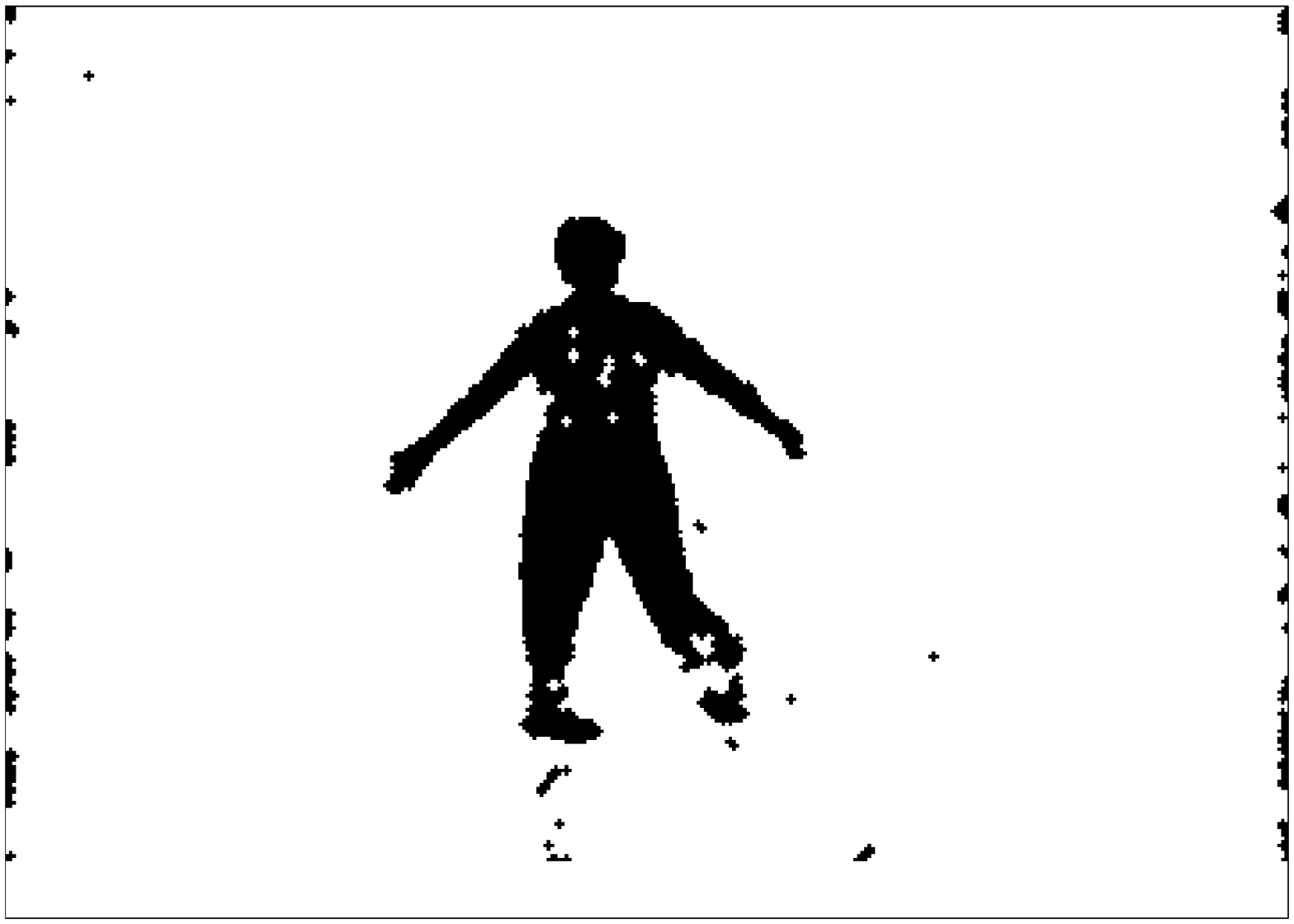} &
\epsfxsize=1in\epsfysize=0.67in\epsfbox{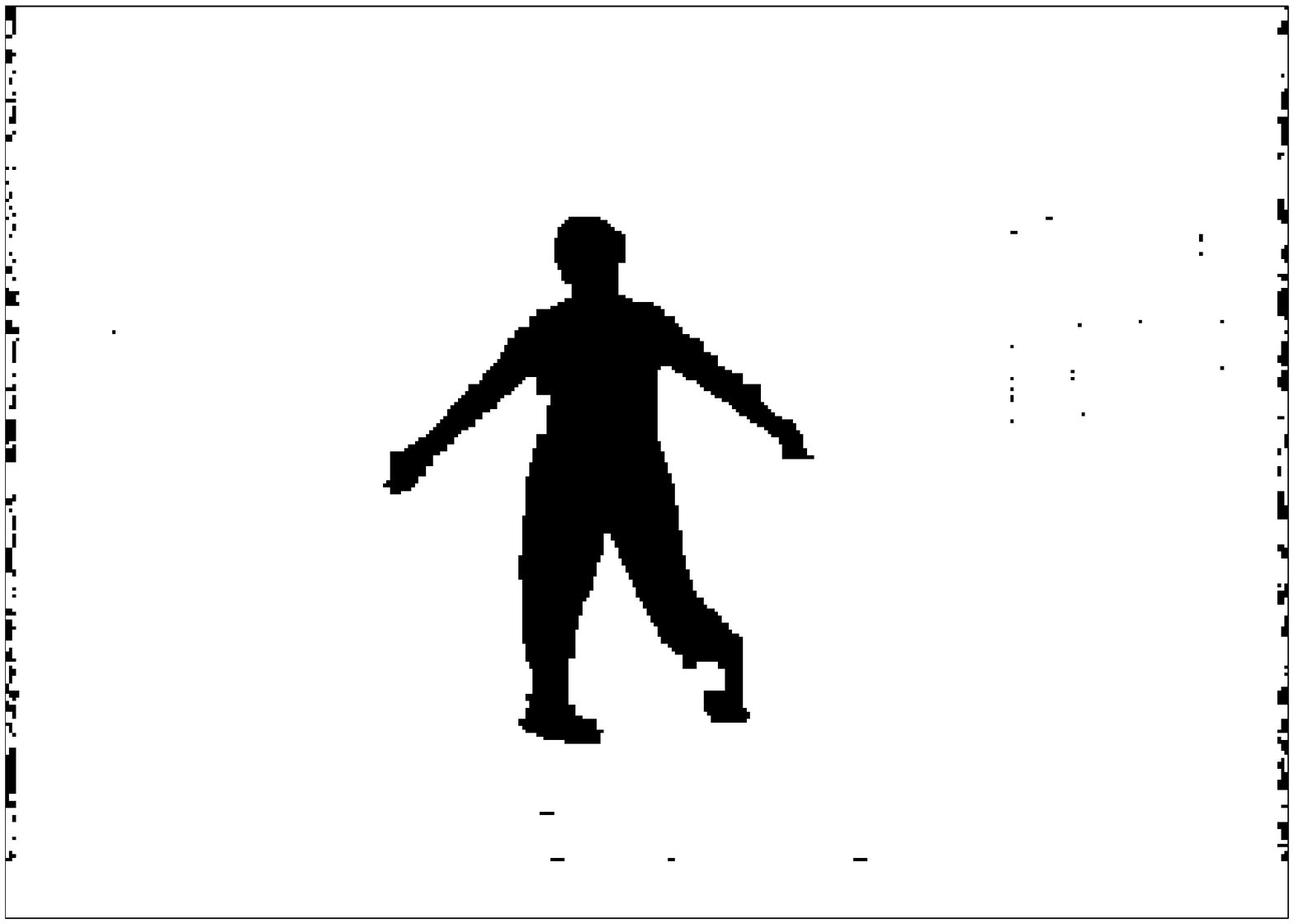} \\ \hline
\epsfxsize=1in\epsfysize=0.67in\epsfbox{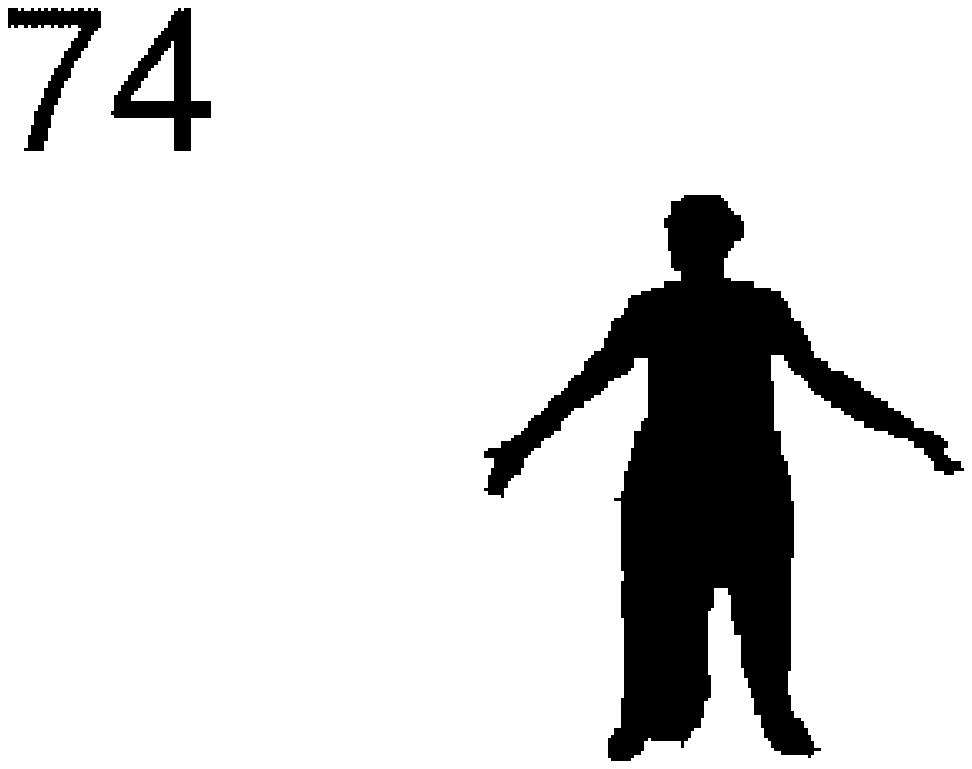} &
\epsfxsize=1in\epsfysize=0.67in\epsfbox{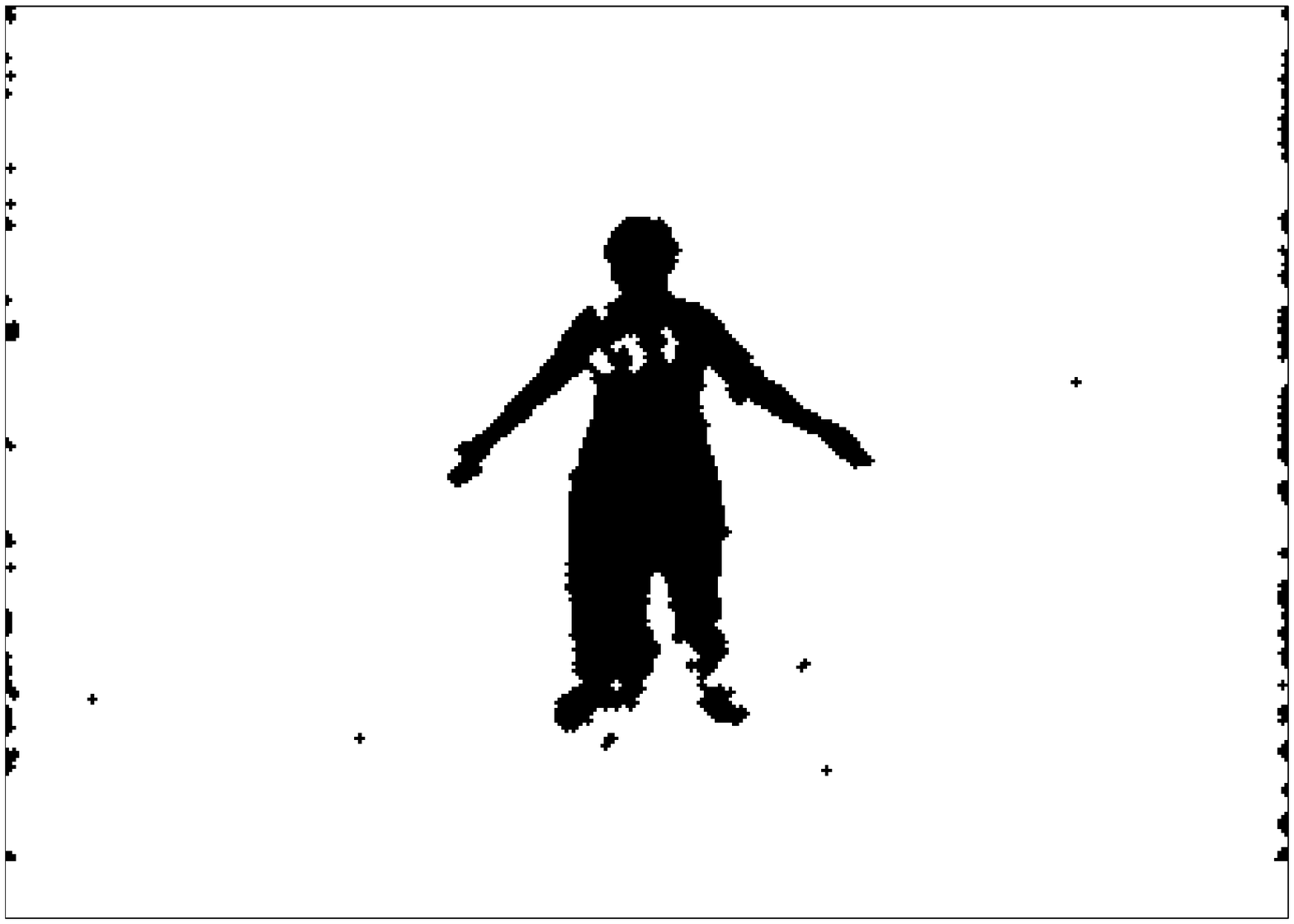} &
\epsfxsize=1in\epsfysize=0.67in\epsfbox{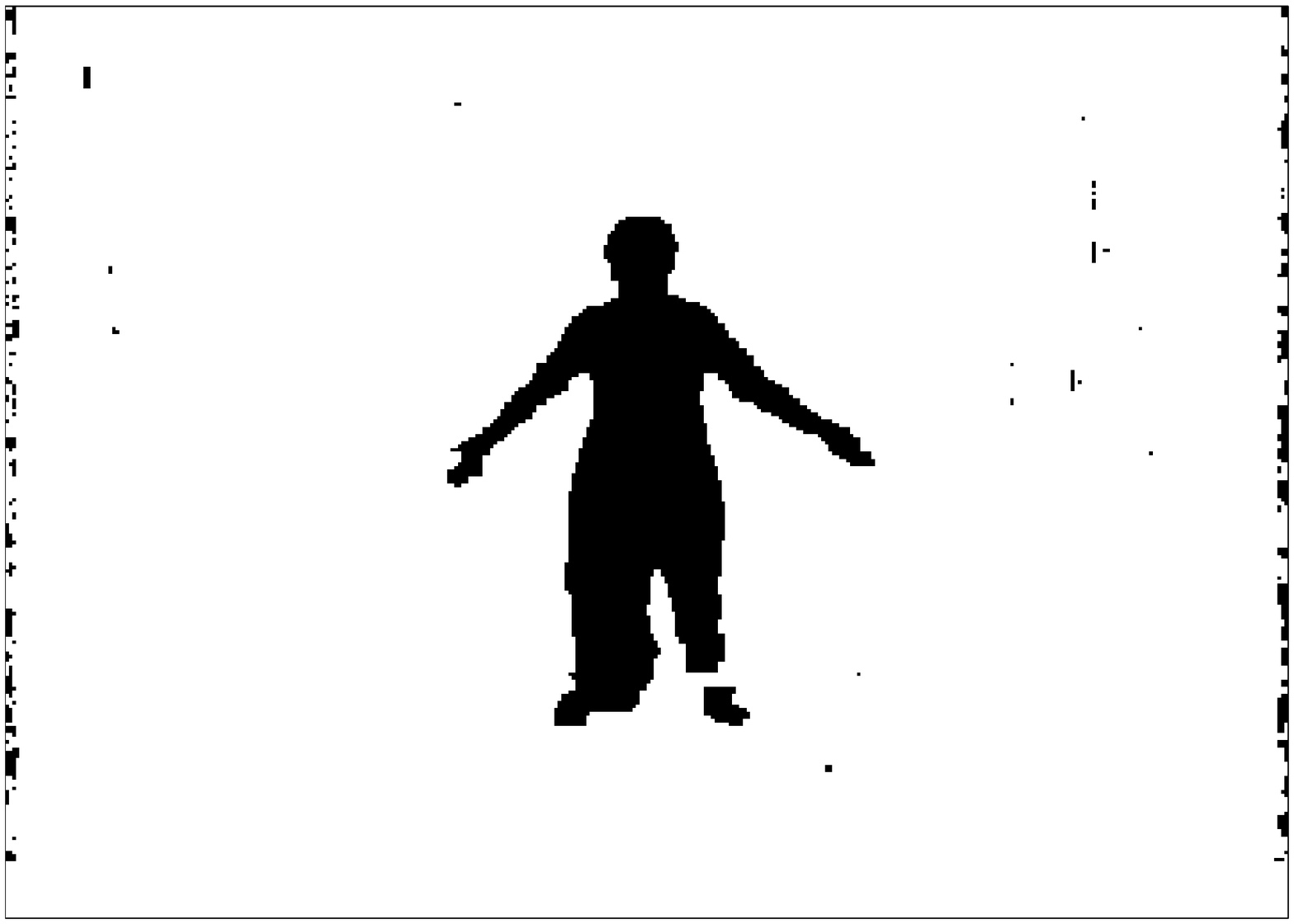} \\ \hline
\epsfxsize=1in\epsfysize=0.67in\epsfbox{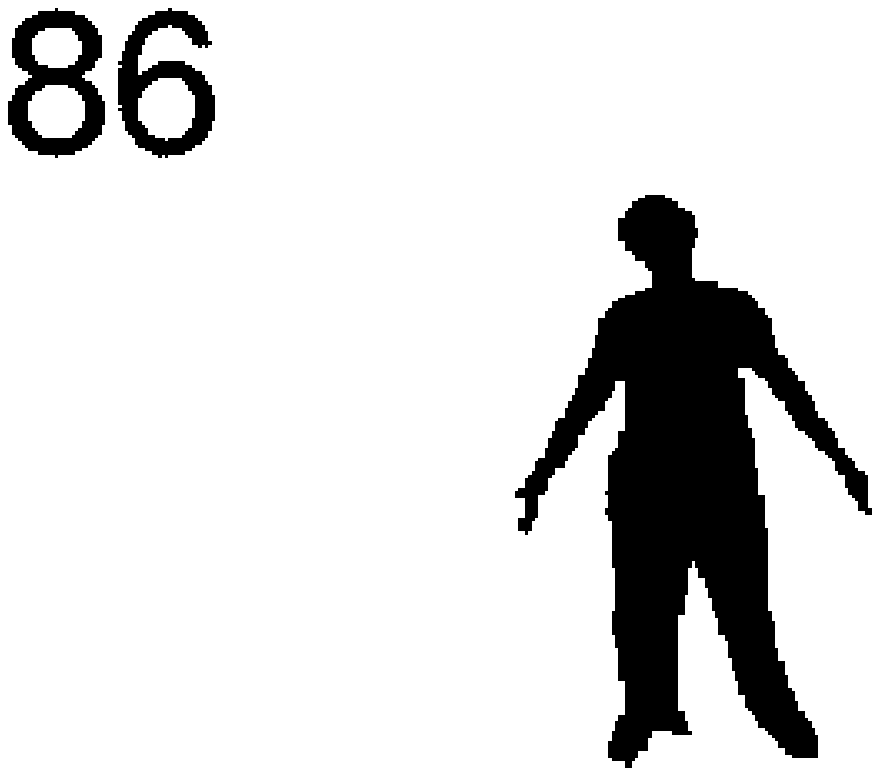} &
\epsfxsize=1in\epsfysize=0.67in\epsfbox{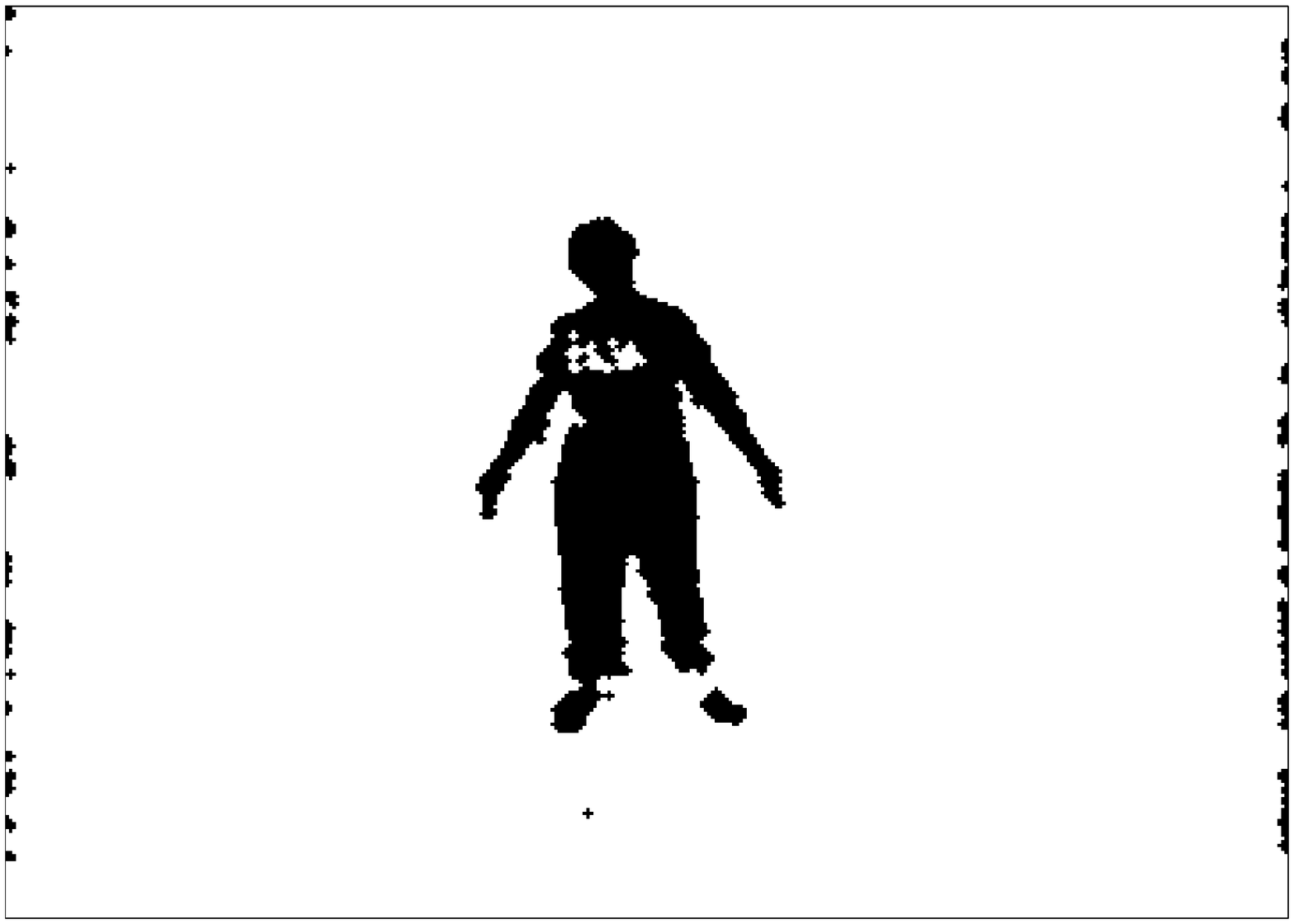} &
\epsfxsize=1in\epsfysize=0.67in\epsfbox{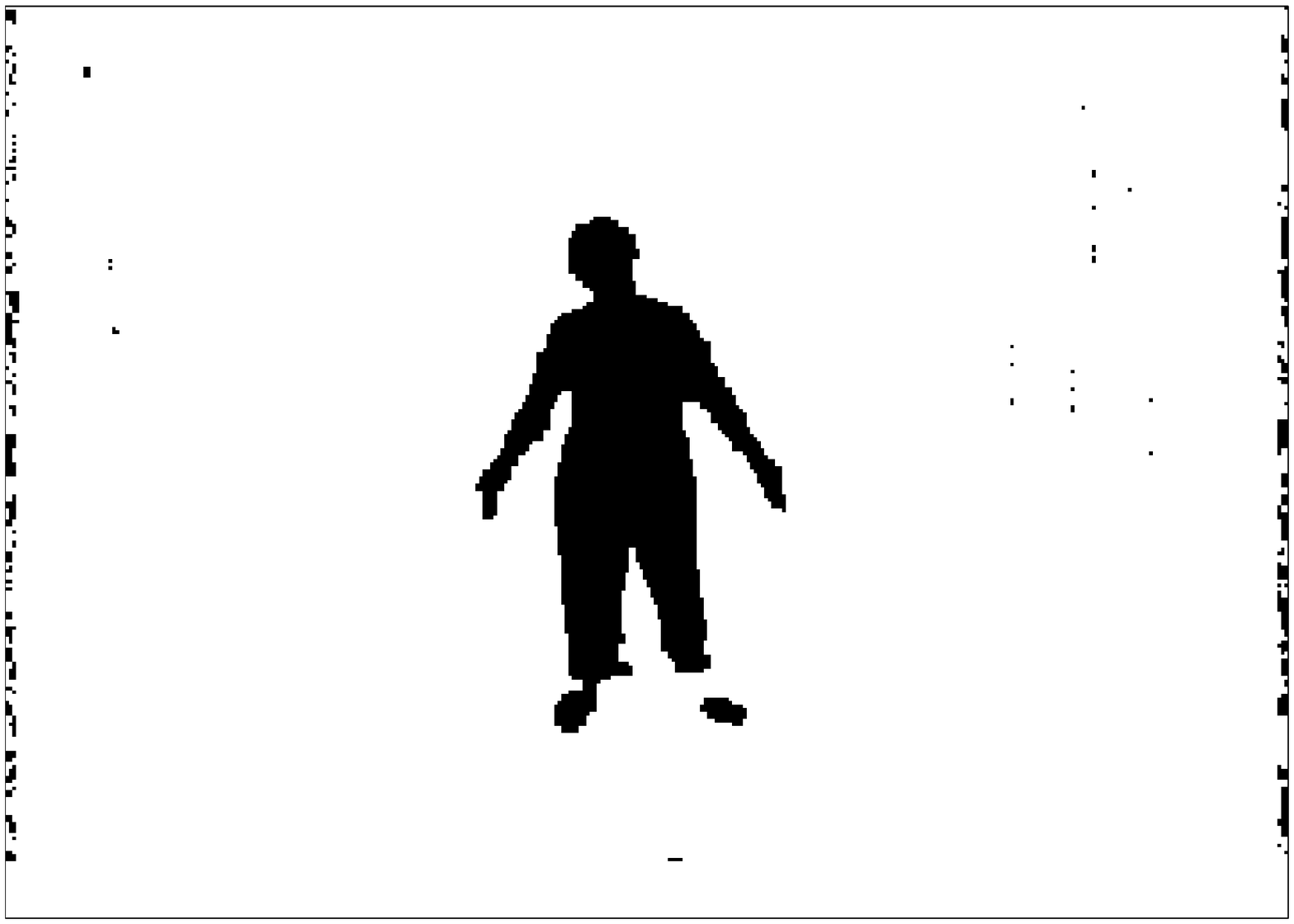} \\ \hline
\epsfxsize=1in\epsfysize=0.67in\epsfbox{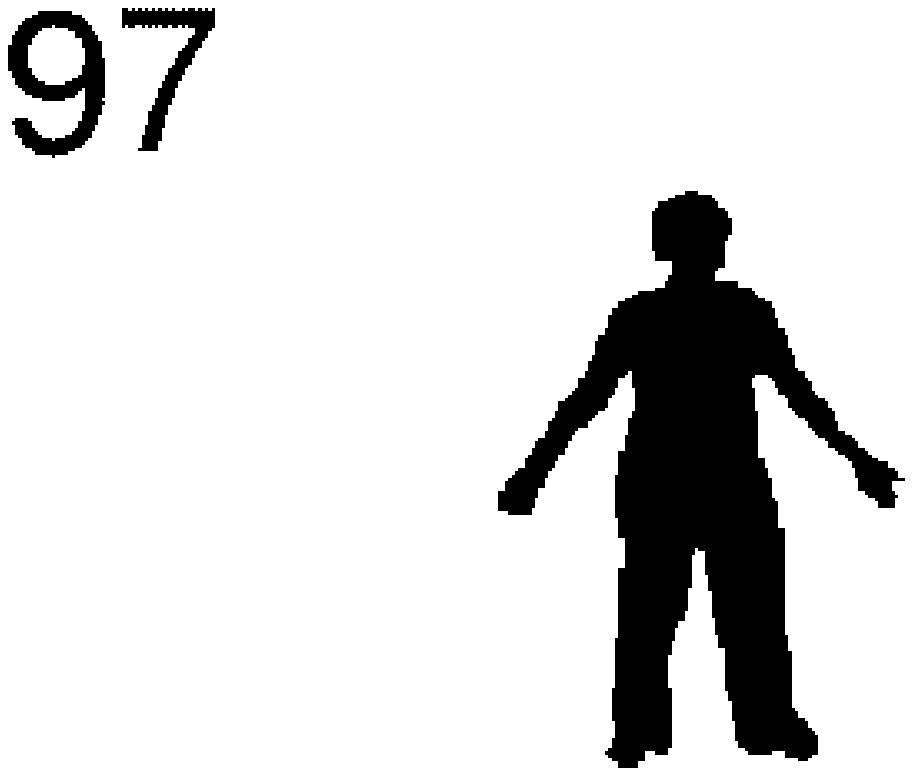} &
\epsfxsize=1in\epsfysize=0.67in\epsfbox{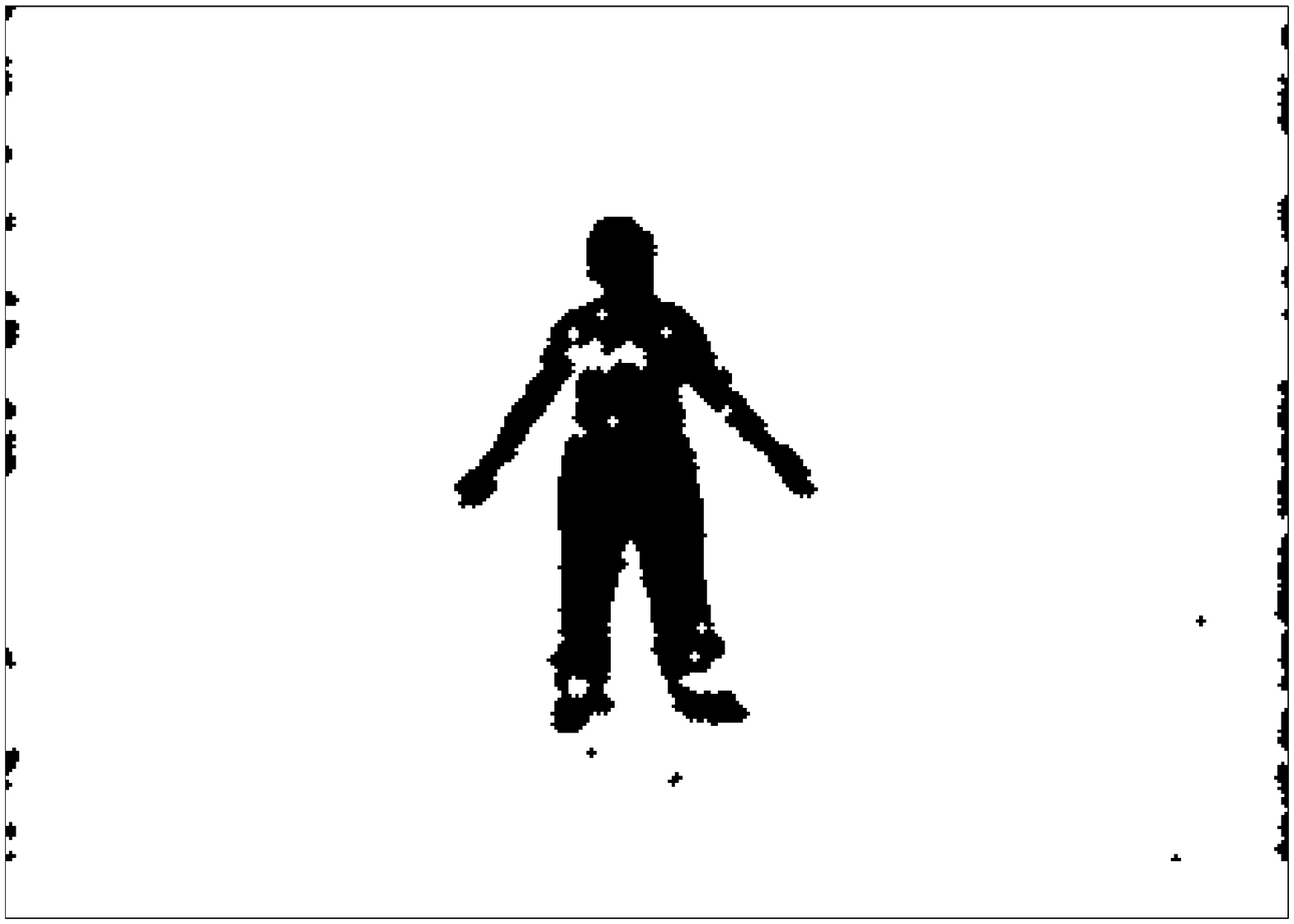} &
\epsfxsize=1in\epsfysize=0.67in\epsfbox{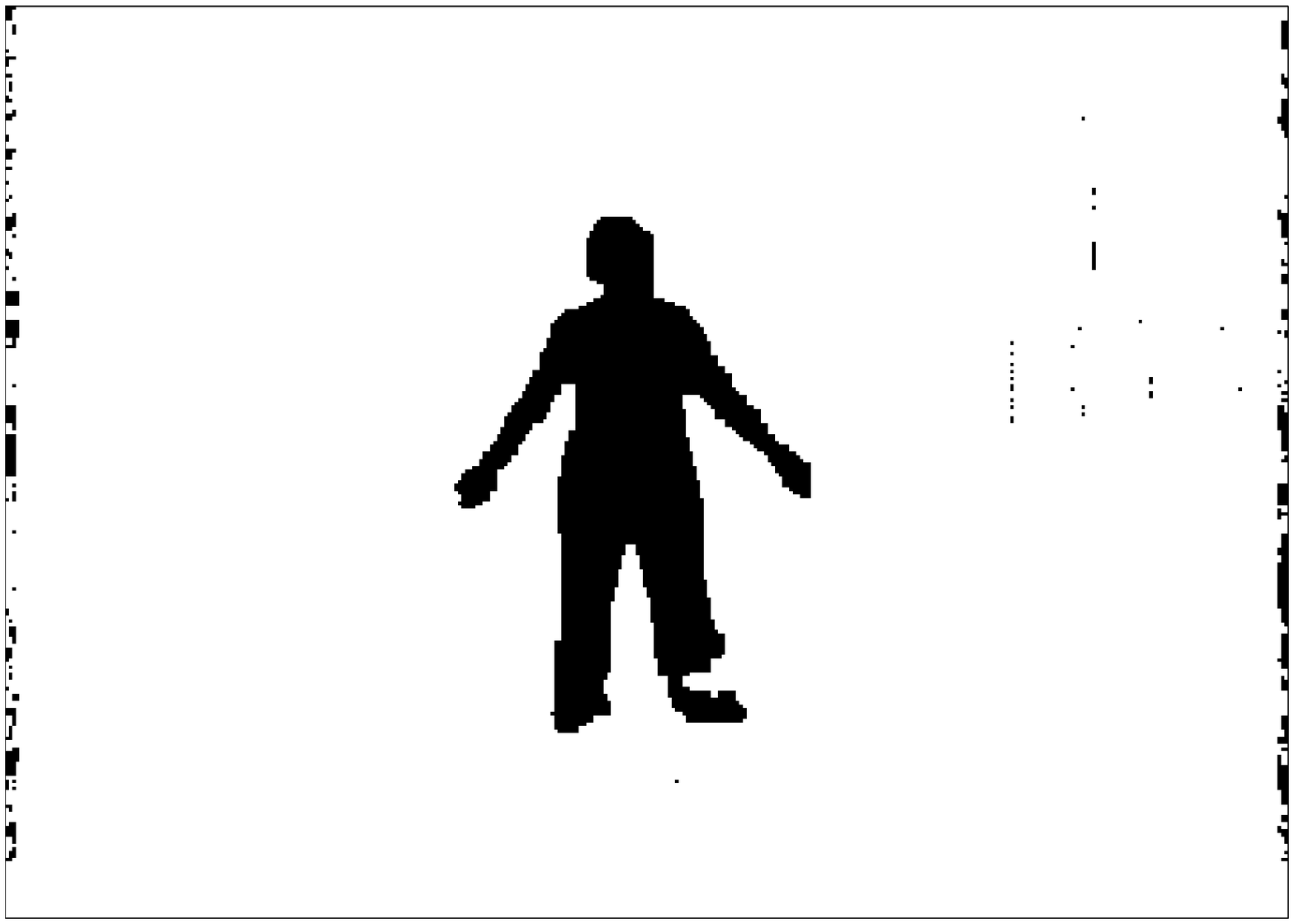} \\ \hline
\epsfxsize=1in\epsfysize=0.67in\epsfbox{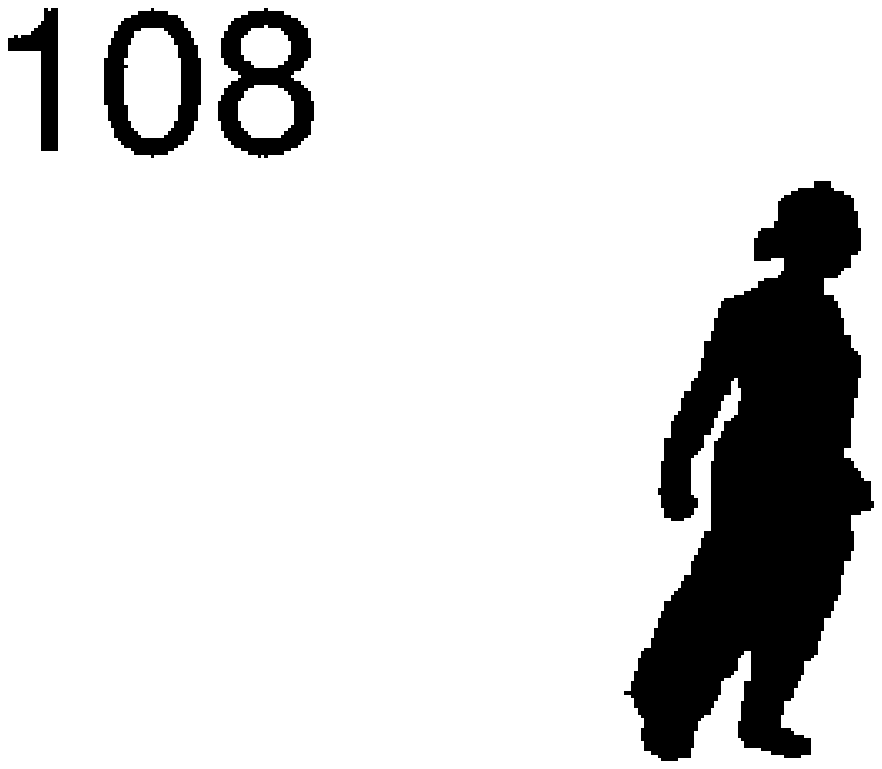} &
\epsfxsize=1in\epsfysize=0.67in\epsfbox{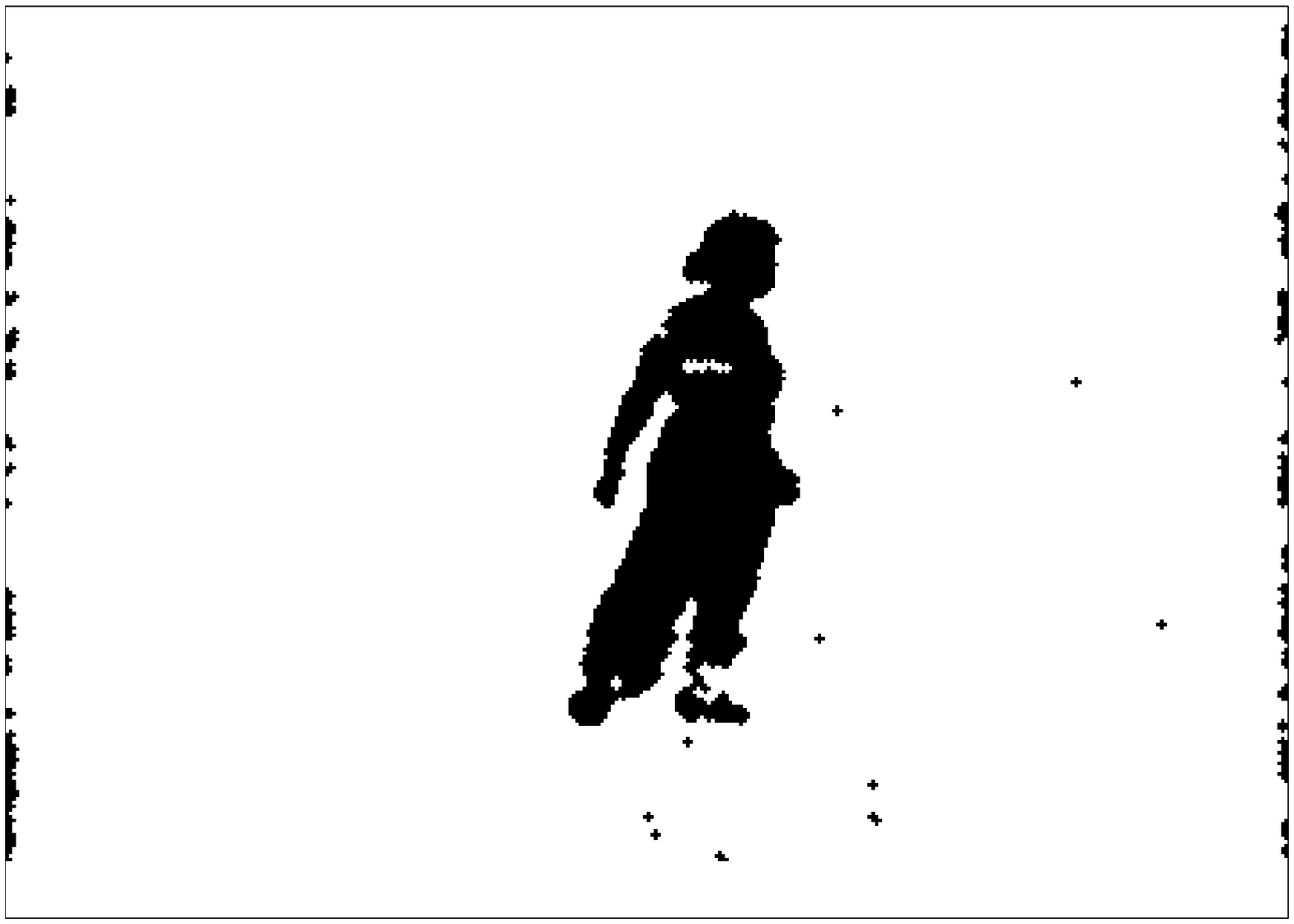} &
\epsfxsize=1in\epsfysize=0.67in\epsfbox{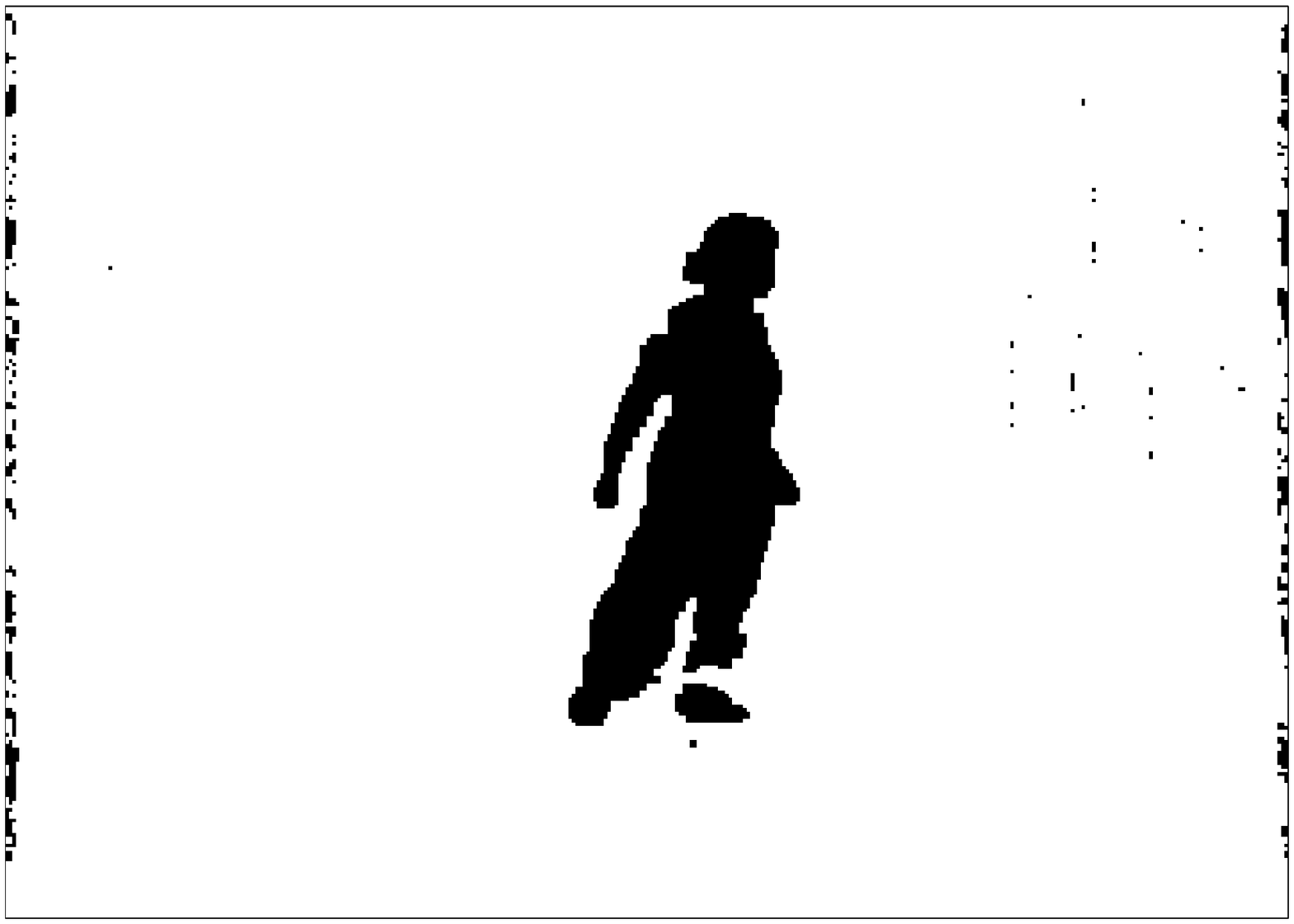} \\ \hline
\epsfxsize=1in\epsfysize=0.67in\epsfbox{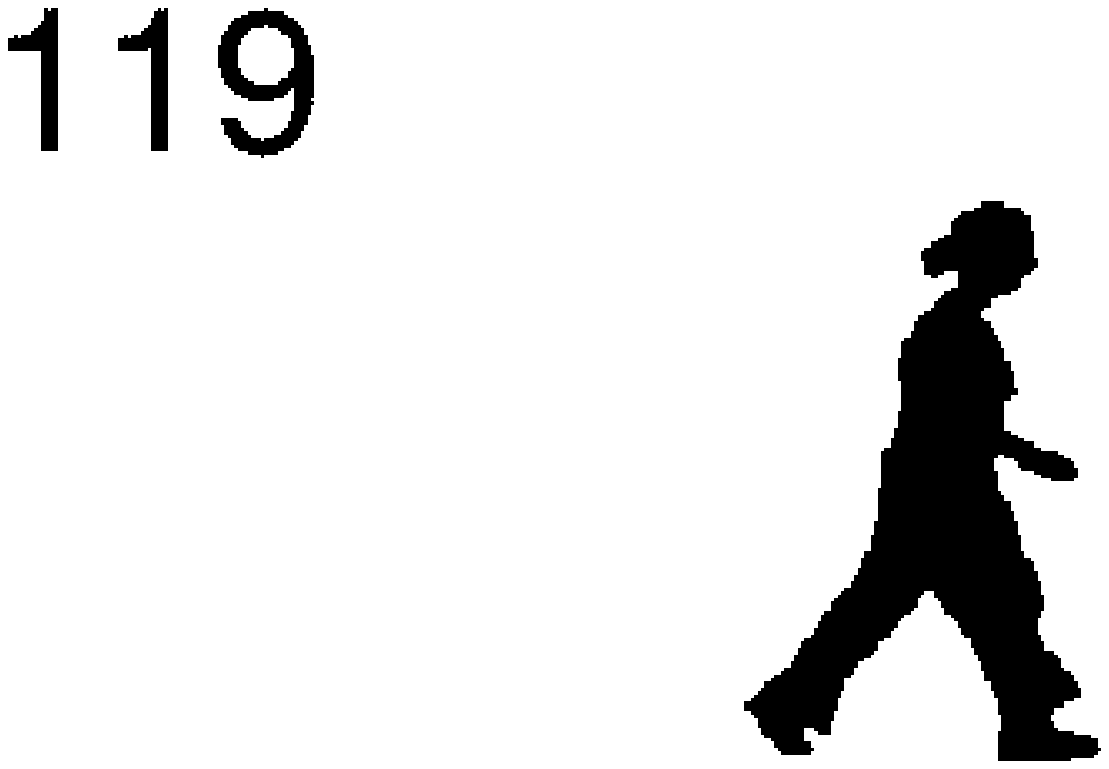} &
\epsfxsize=1in\epsfysize=0.67in\epsfbox{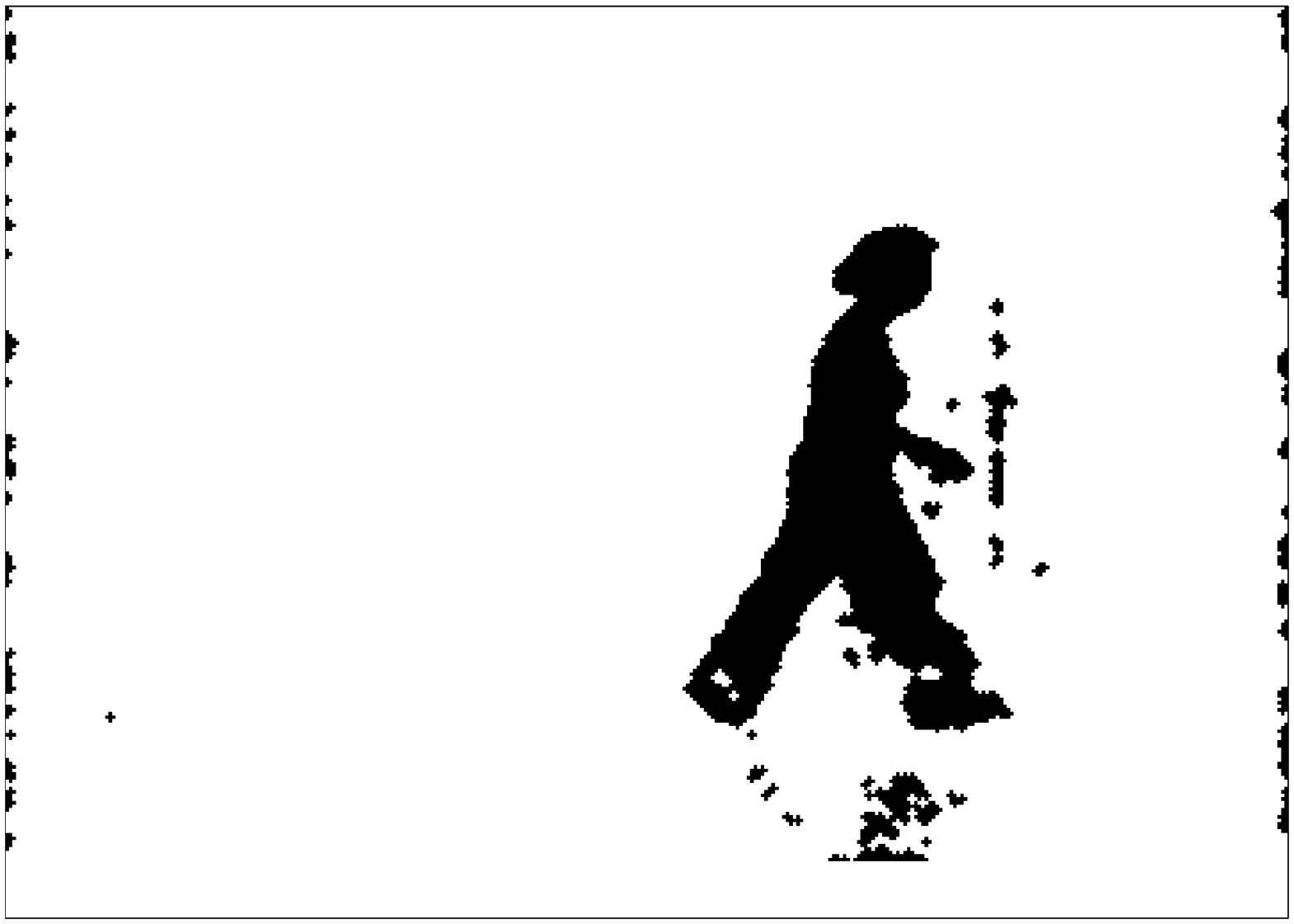} &
\epsfxsize=1in\epsfysize=0.67in\epsfbox{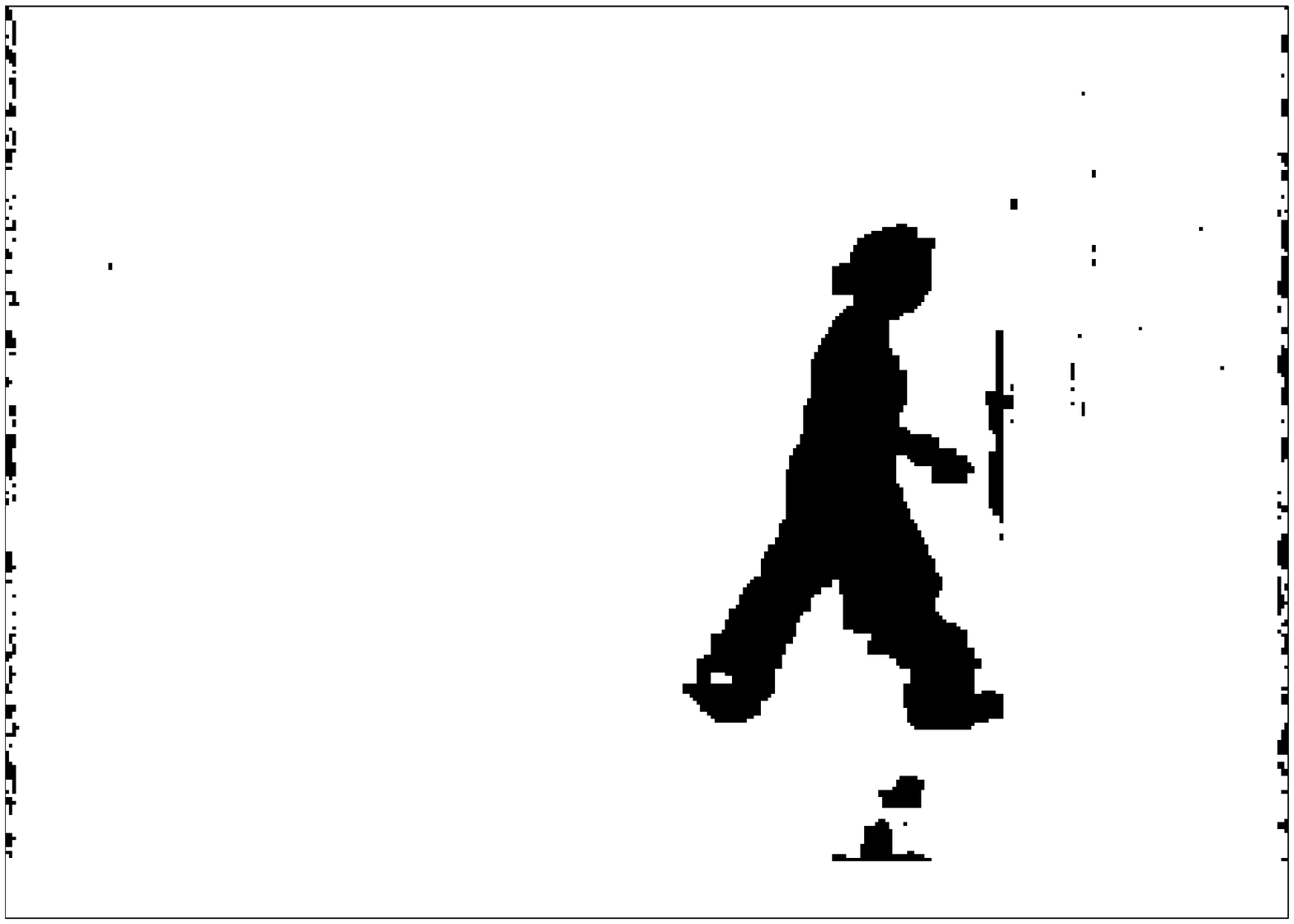} \\ \hline
\epsfxsize=1in\epsfysize=0.67in\epsfbox{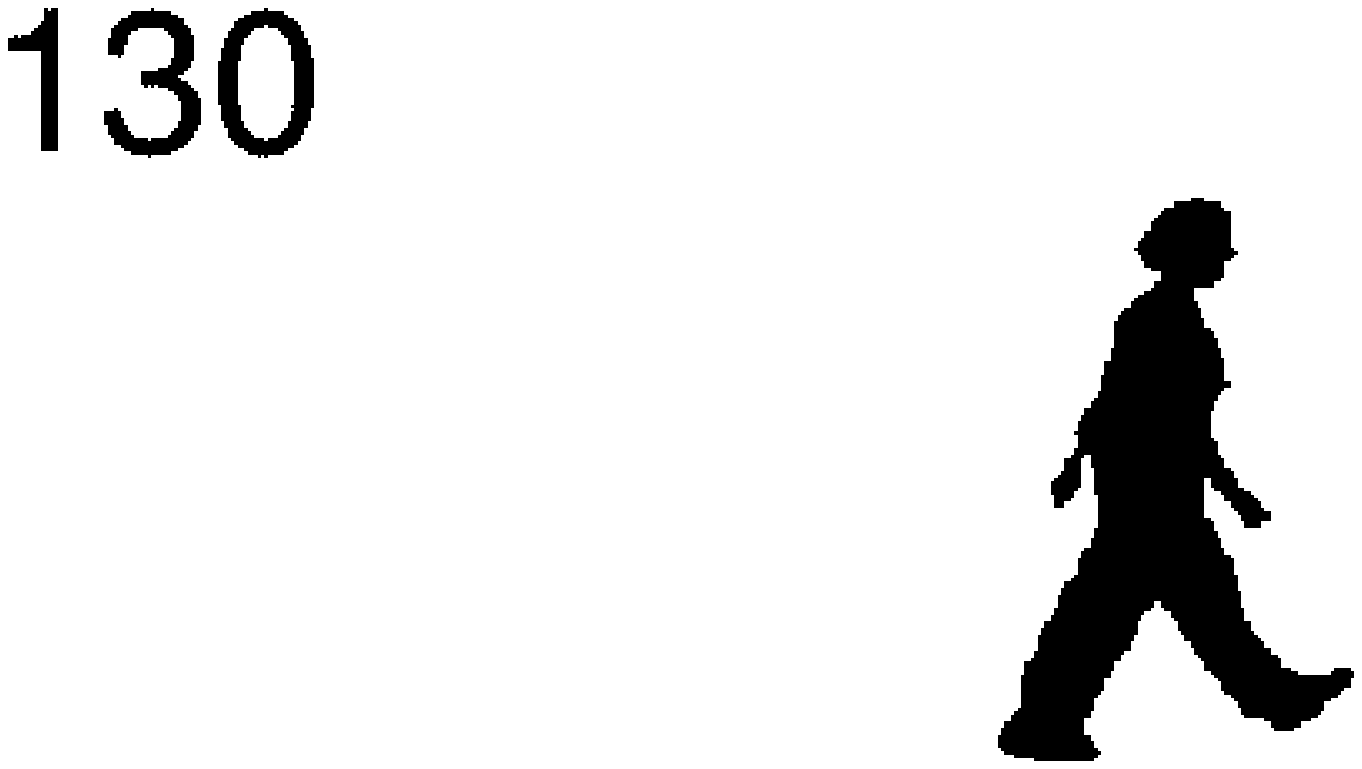} &
\epsfxsize=1in\epsfysize=0.67in\epsfbox{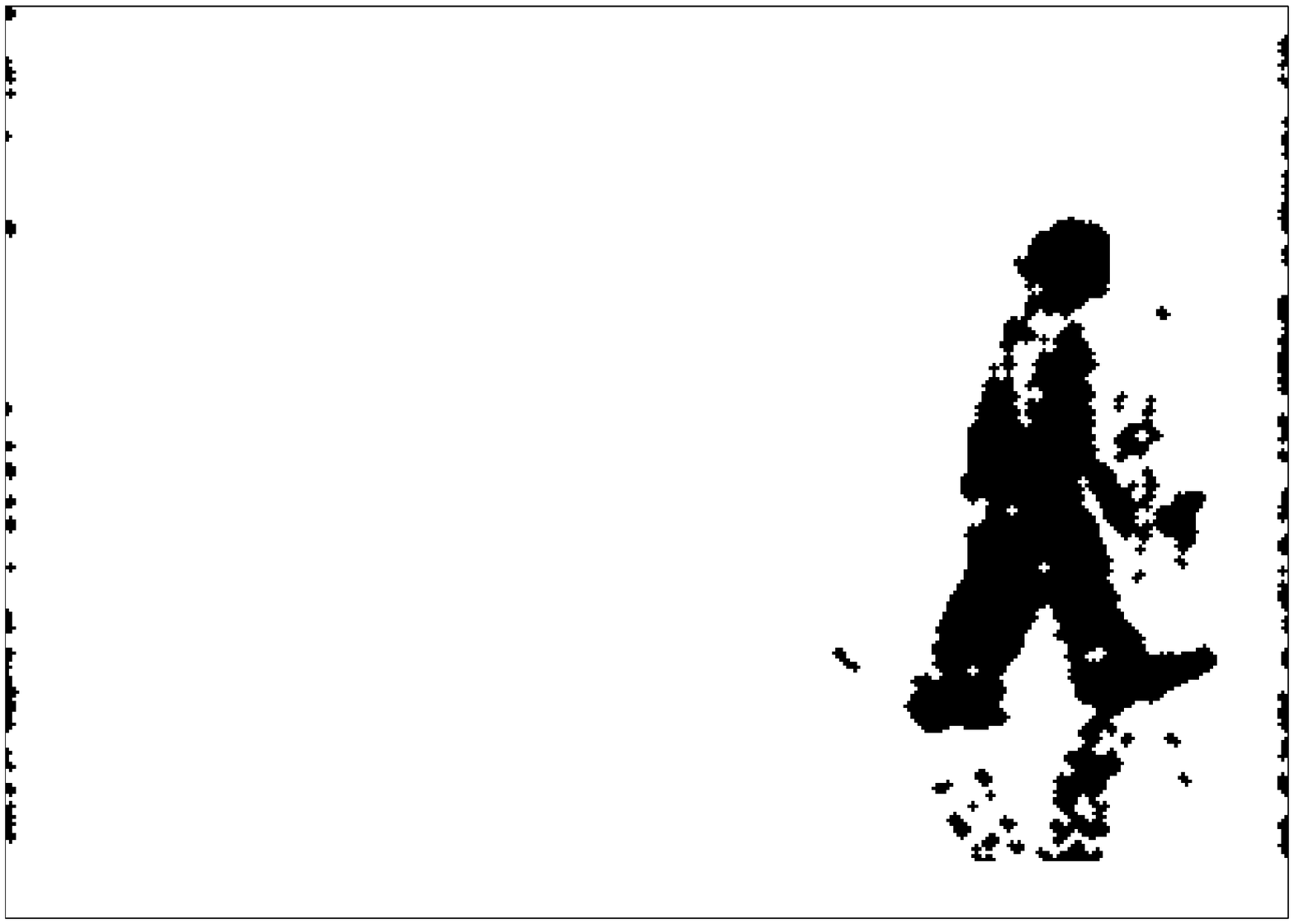} &
\epsfxsize=1in\epsfysize=0.67in\epsfbox{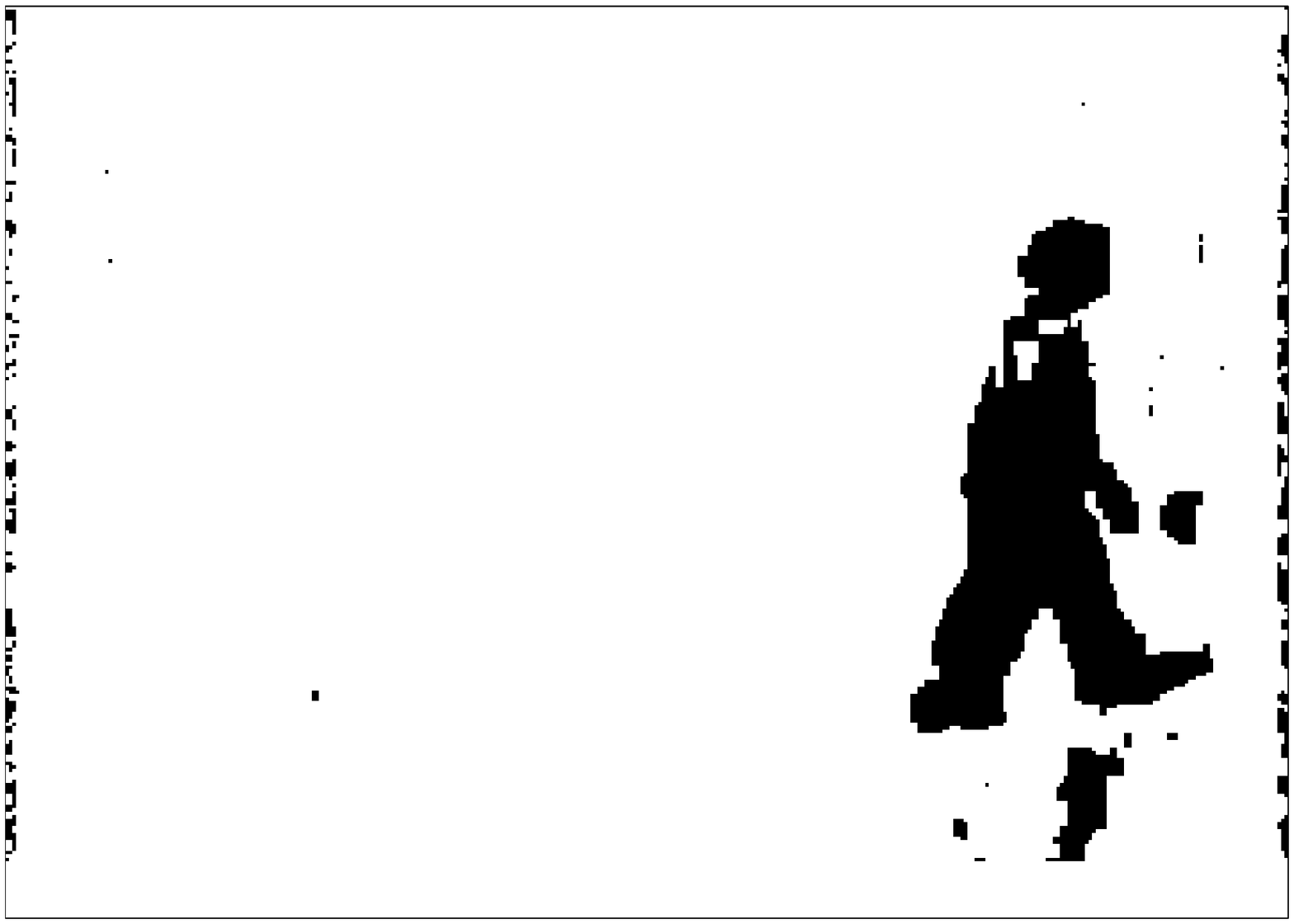} \\ \hline
(Truth) & (Morph.) & (Graph) \\ \hline
\end{tabular}
\end{center}
\caption{A sequence of frames from the {\em Indoor} clip.  The
subject's shirt provides low contrast with the background.}
\label{fig-mccseq}
\end{figure}

\begin{figure}
\begin{center}
\begin{tabular}{|@{}c@{}|@{}c@{}|@{}c@{}|} \hline
\epsfxsize=1in\epsfysize=0.67in\epsfbox{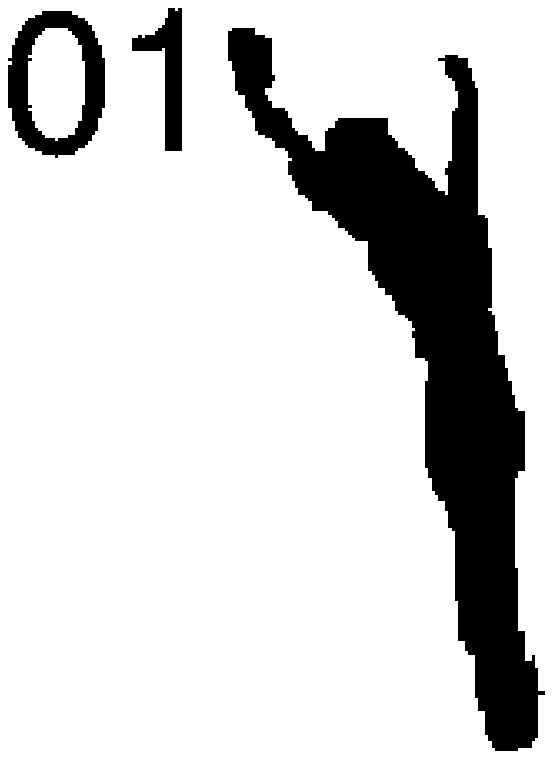} &
\epsfxsize=1in\epsfysize=0.67in\epsfbox{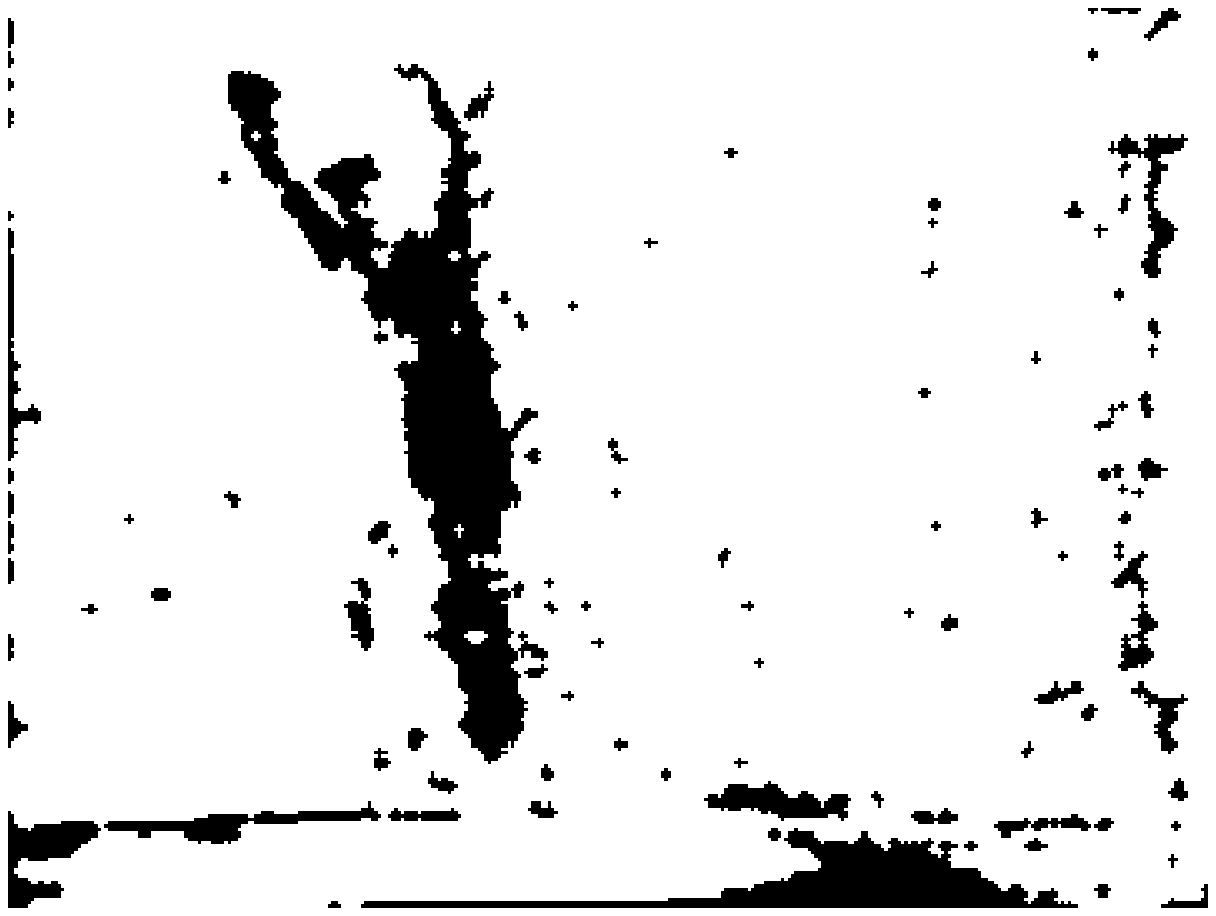} &
\epsfxsize=1in\epsfysize=0.67in\epsfbox{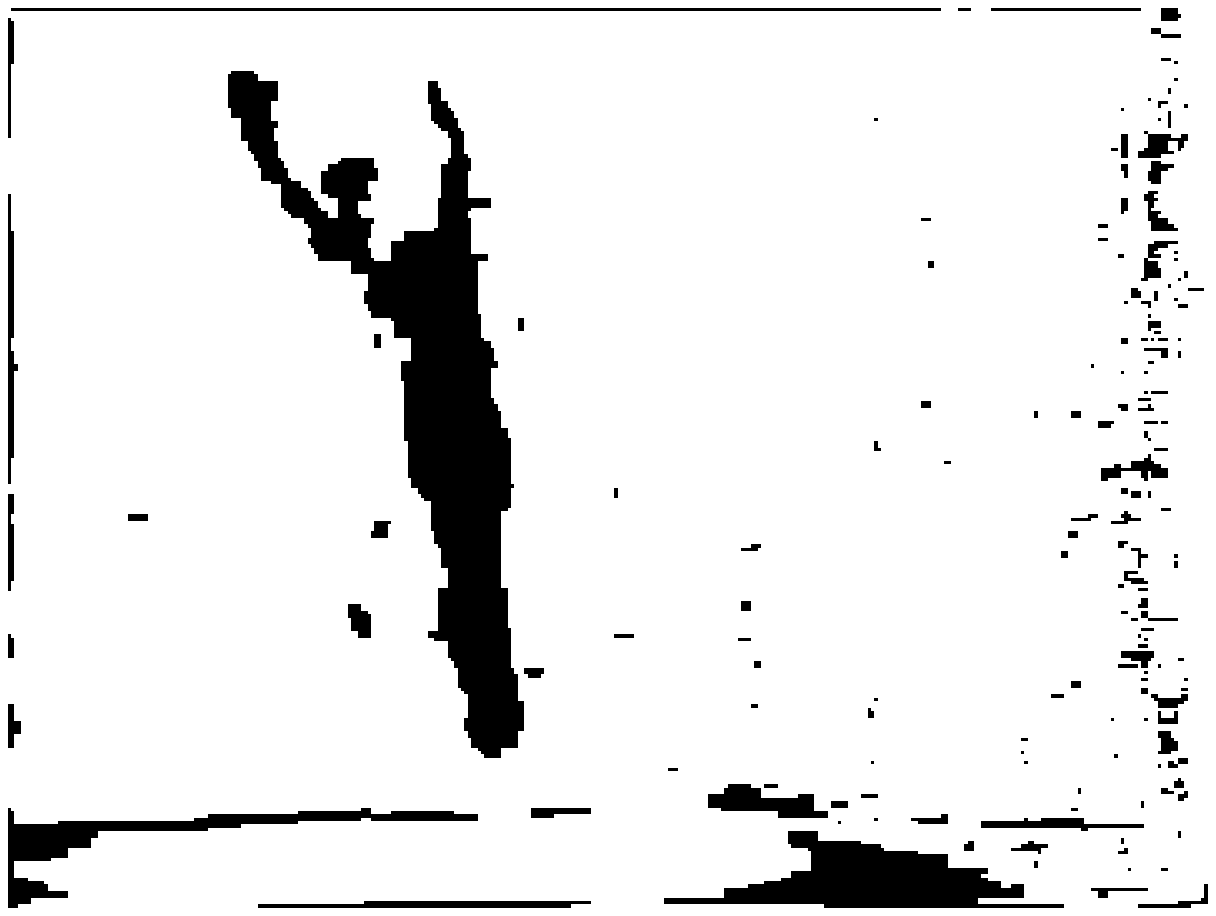} \\ \hline
\epsfxsize=1in\epsfysize=0.67in\epsfbox{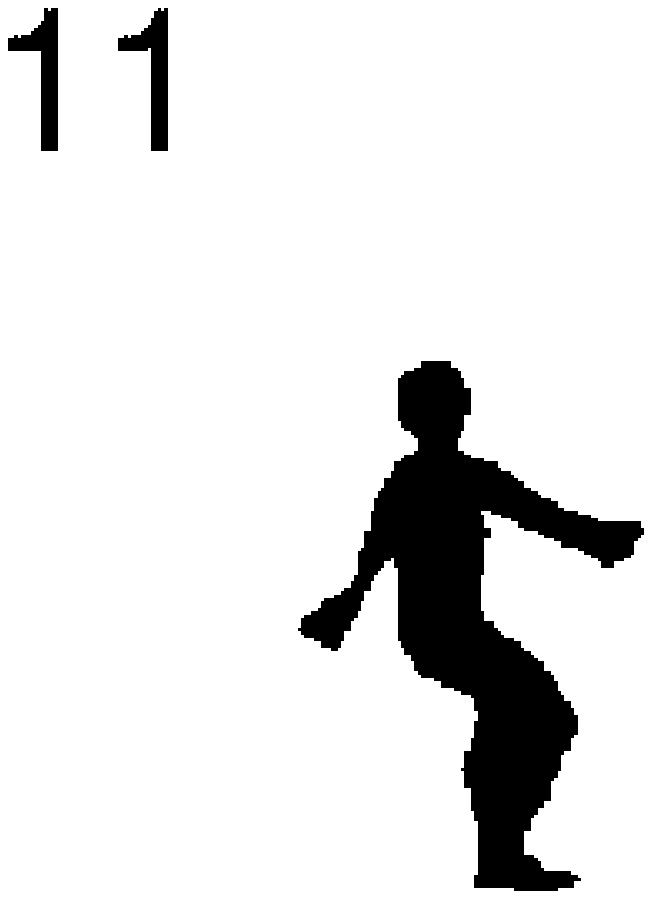} &
\epsfxsize=1in\epsfysize=0.67in\epsfbox{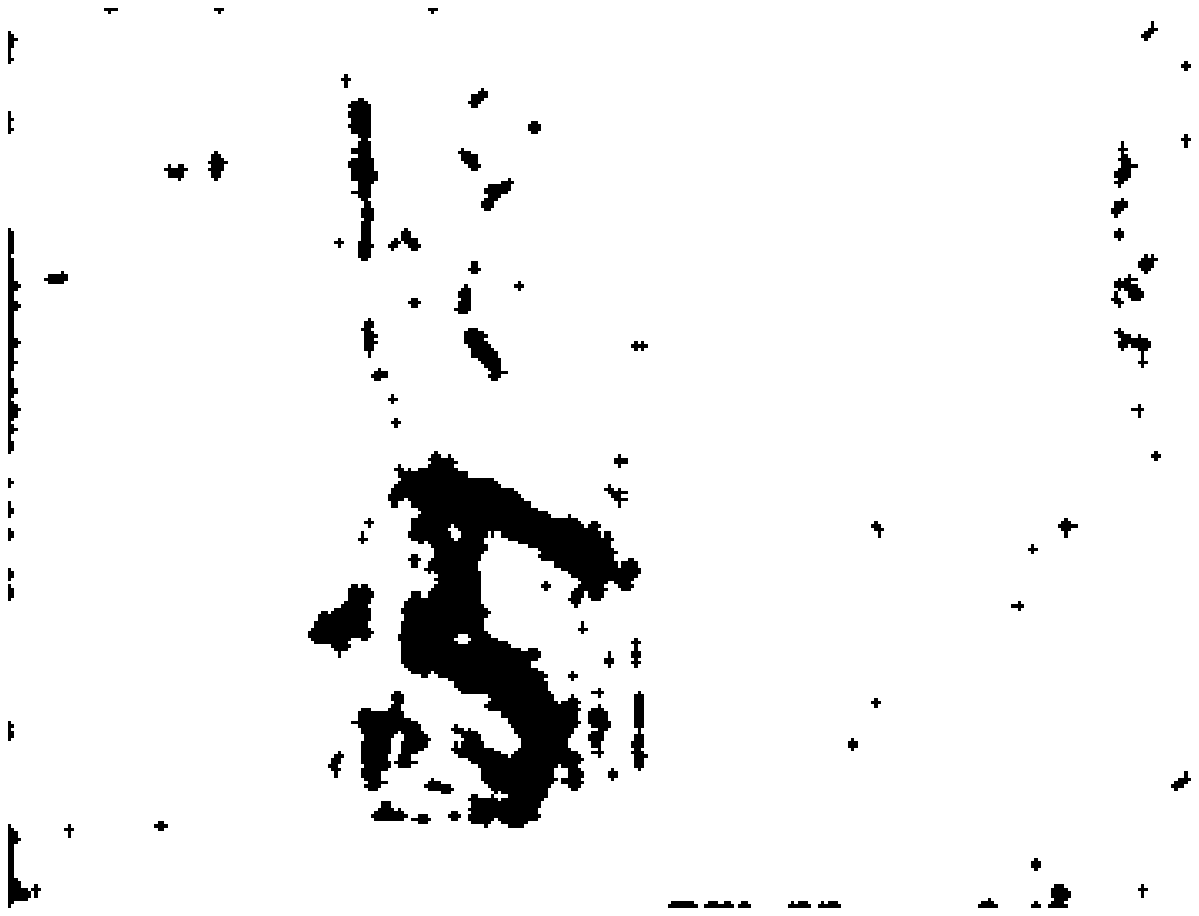} &
\epsfxsize=1in\epsfysize=0.67in\epsfbox{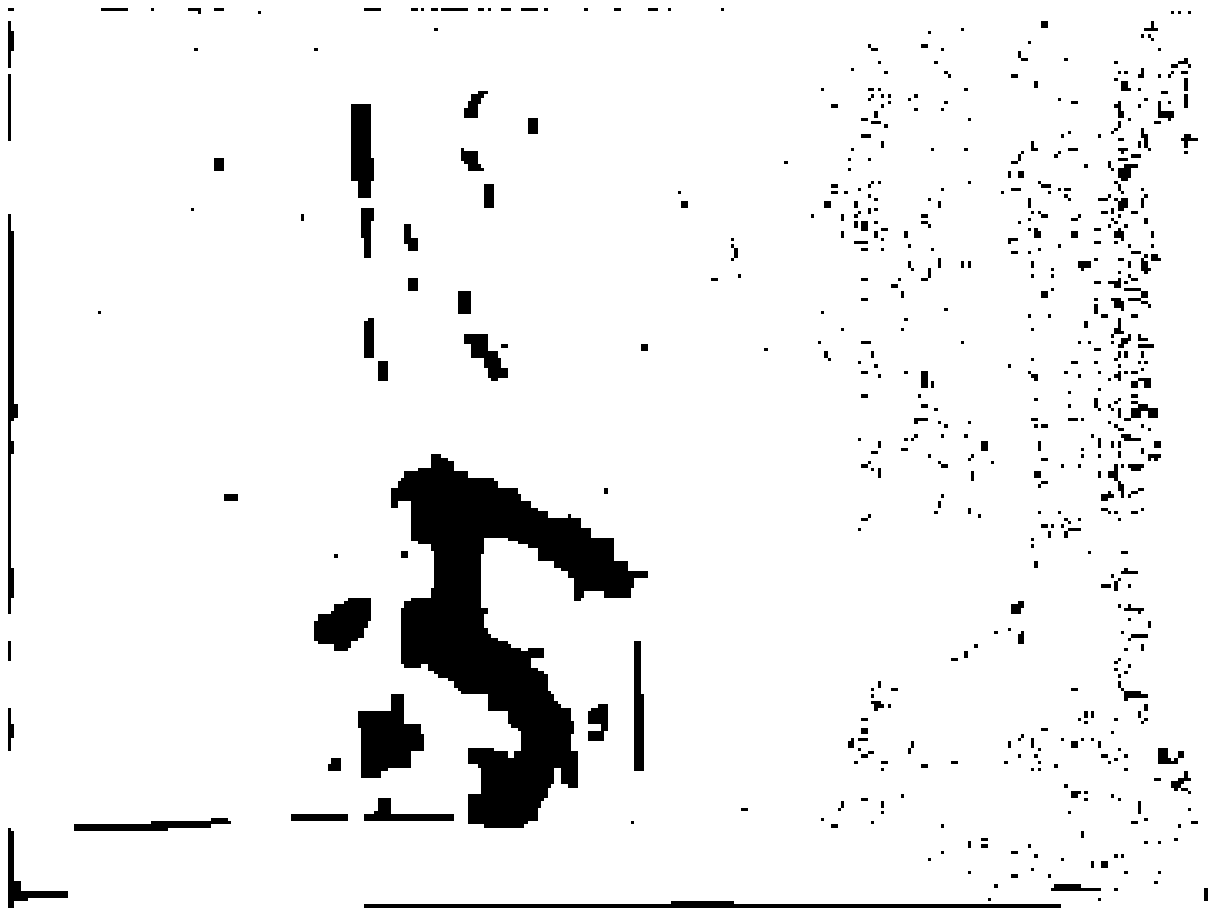} \\ \hline
\epsfxsize=1in\epsfysize=0.67in\epsfbox{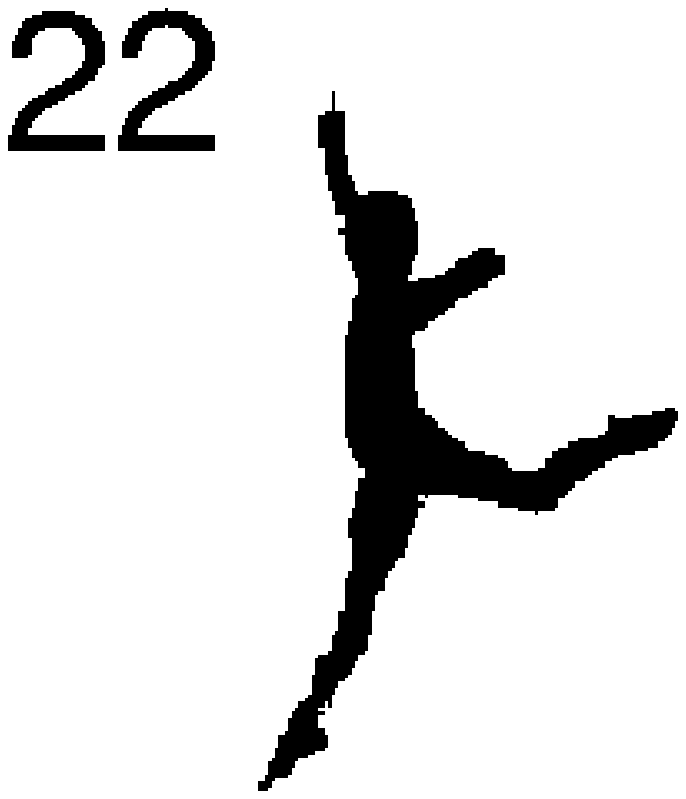} &
\epsfxsize=1in\epsfysize=0.67in\epsfbox{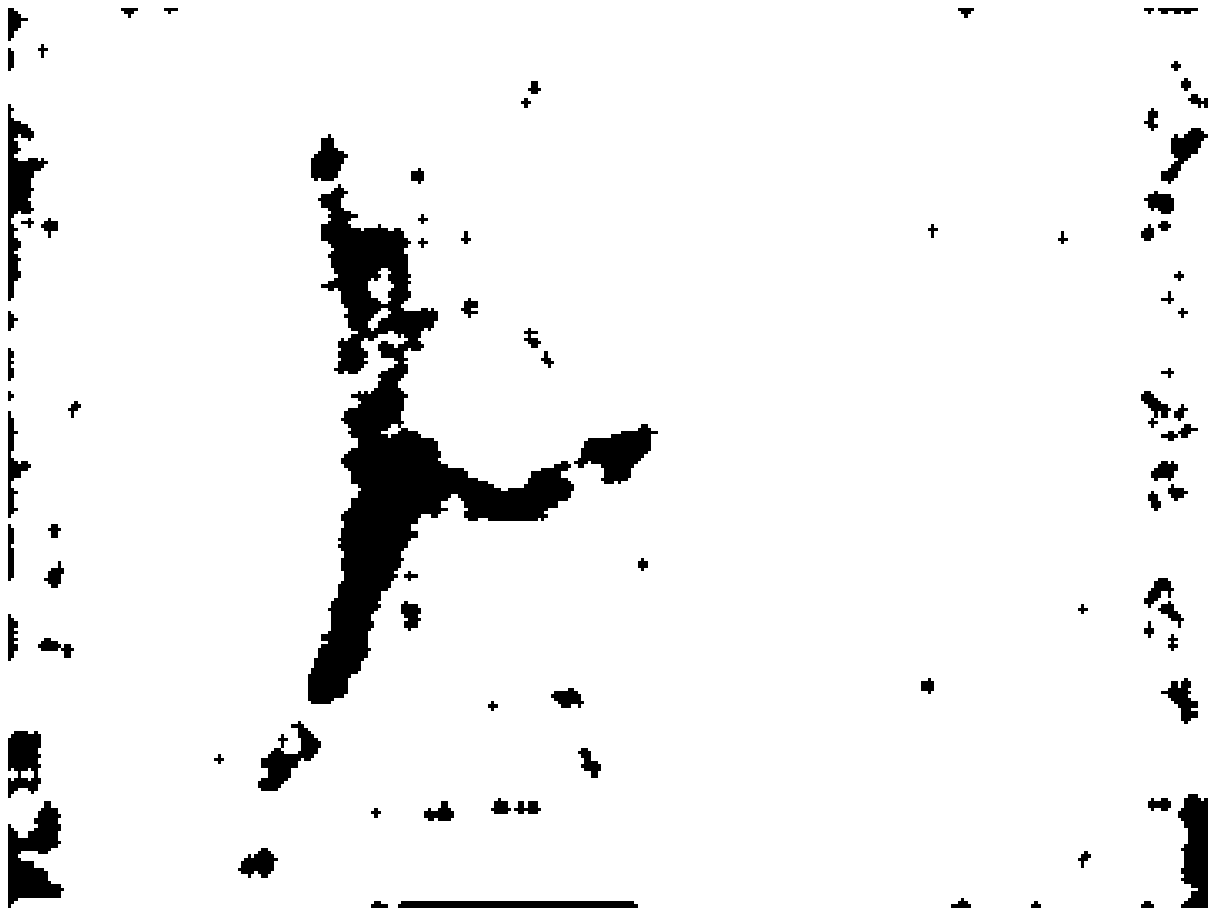} &
\epsfxsize=1in\epsfysize=0.67in\epsfbox{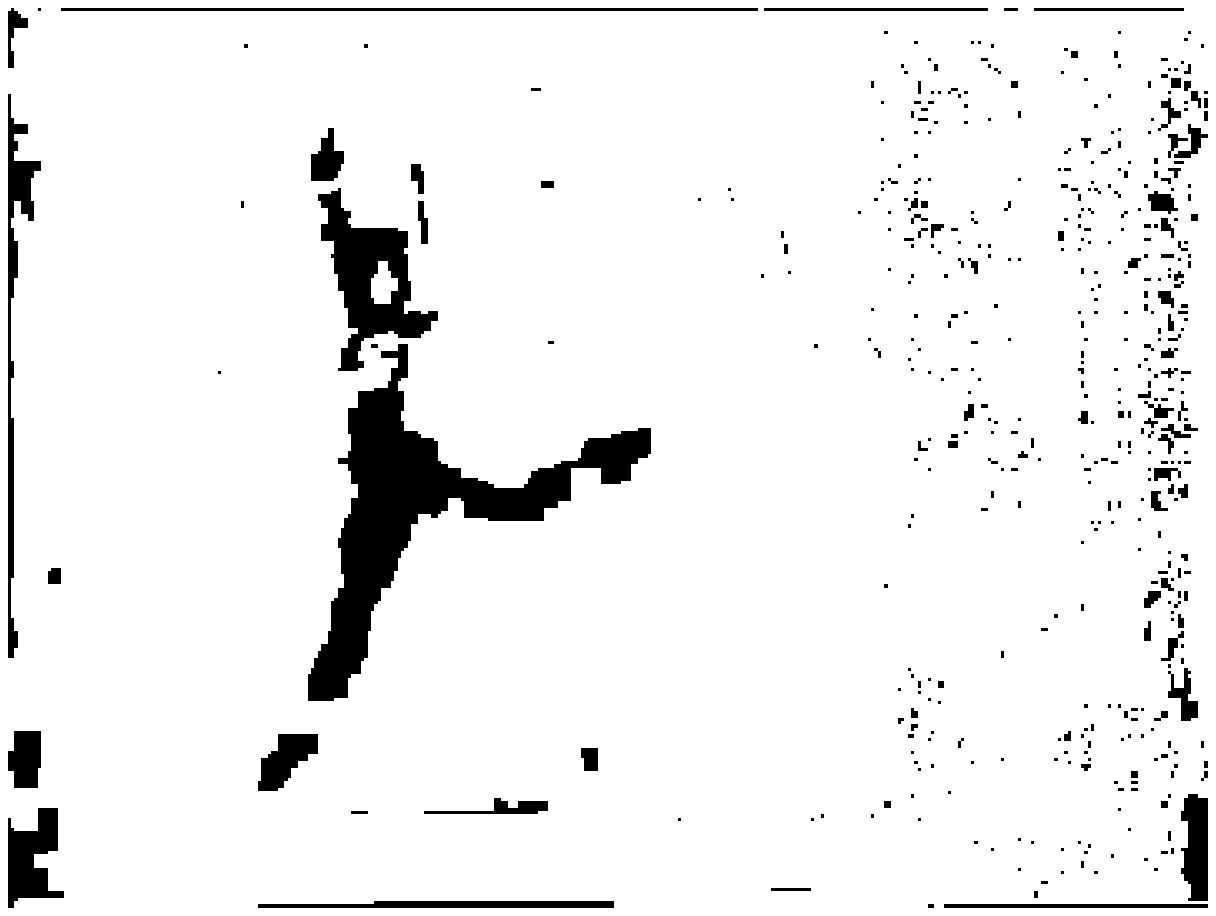} \\ \hline
\epsfxsize=1in\epsfysize=0.67in\epsfbox{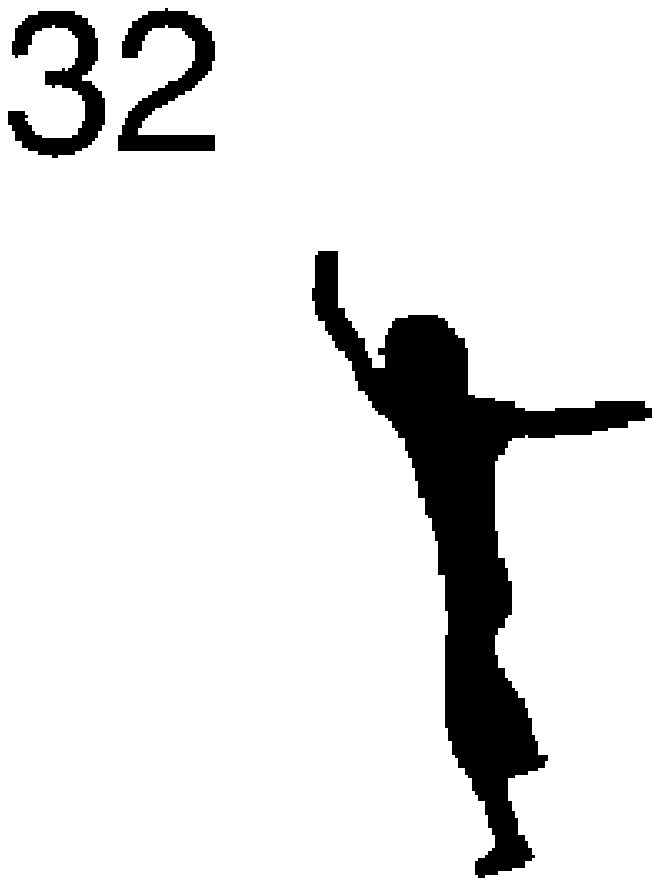} &
\epsfxsize=1in\epsfysize=0.67in\epsfbox{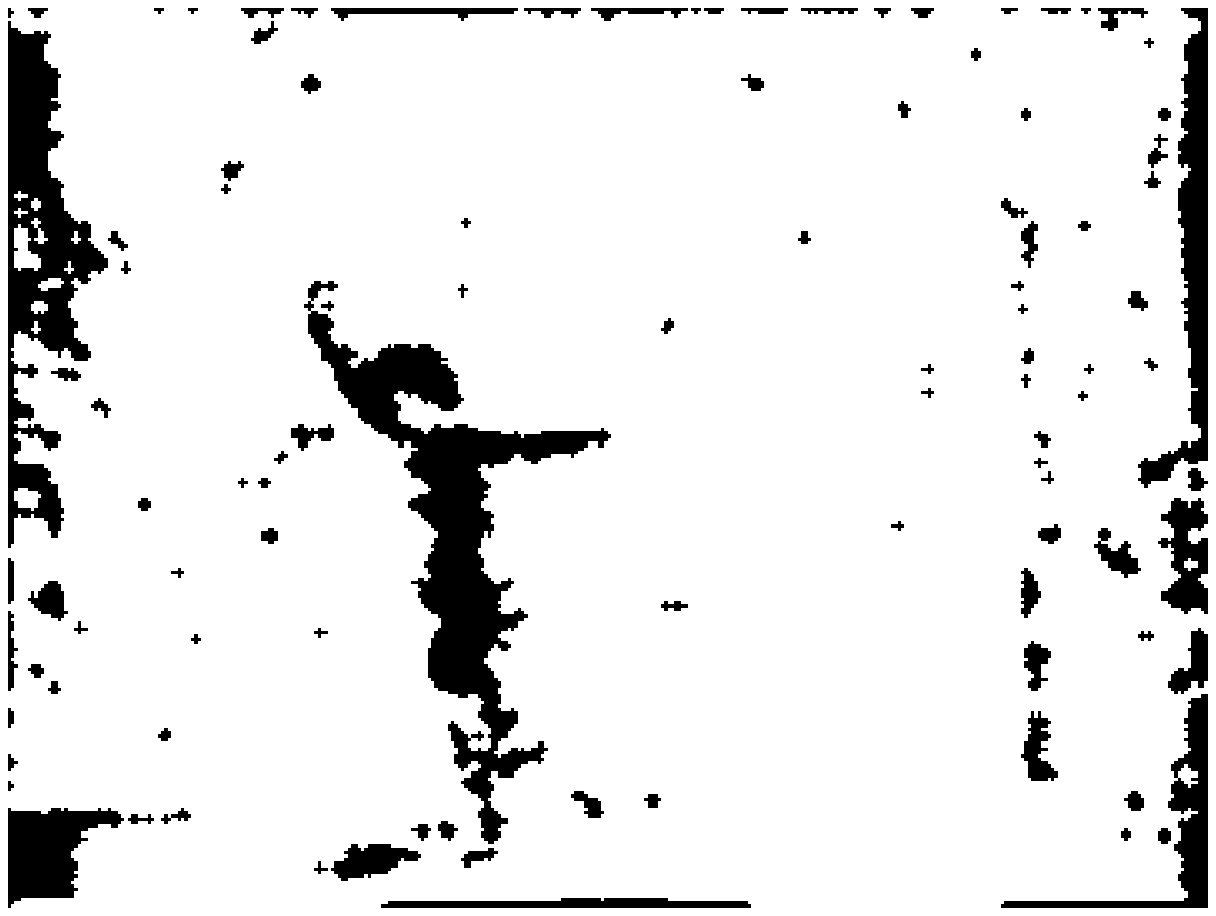} &
\epsfxsize=1in\epsfysize=0.67in\epsfbox{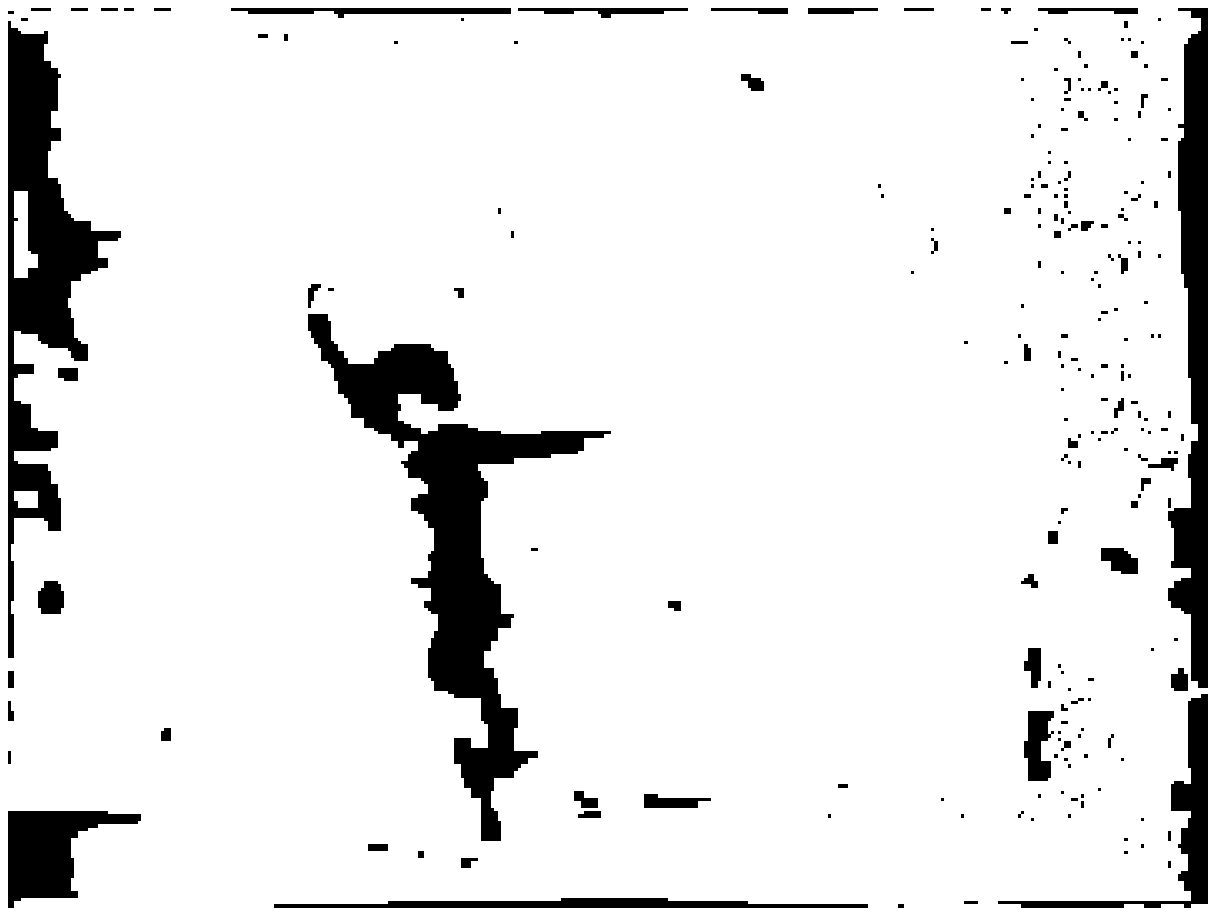} \\ \hline
\epsfxsize=1in\epsfysize=0.67in\epsfbox{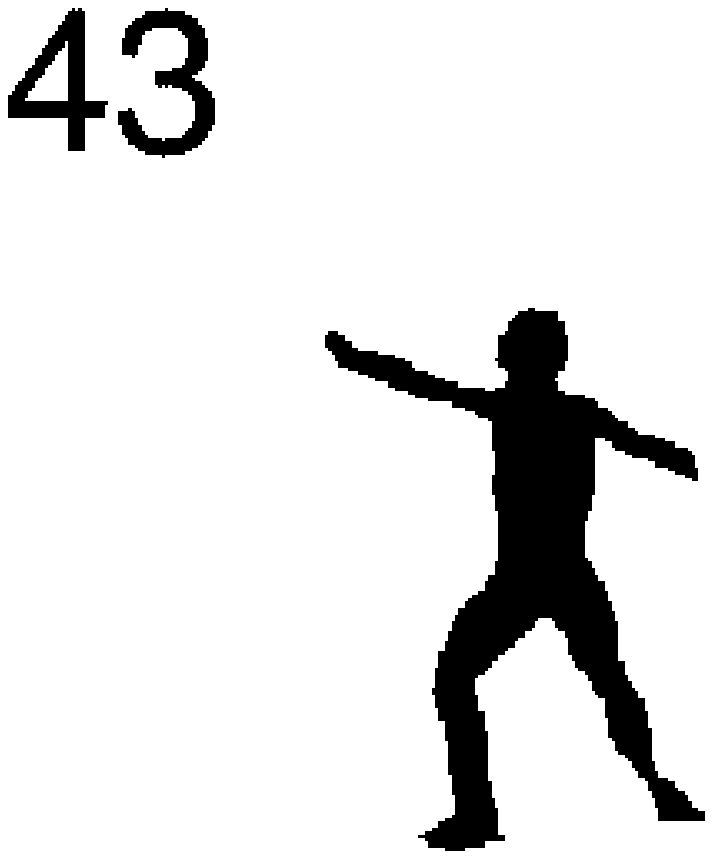} &
\epsfxsize=1in\epsfysize=0.67in\epsfbox{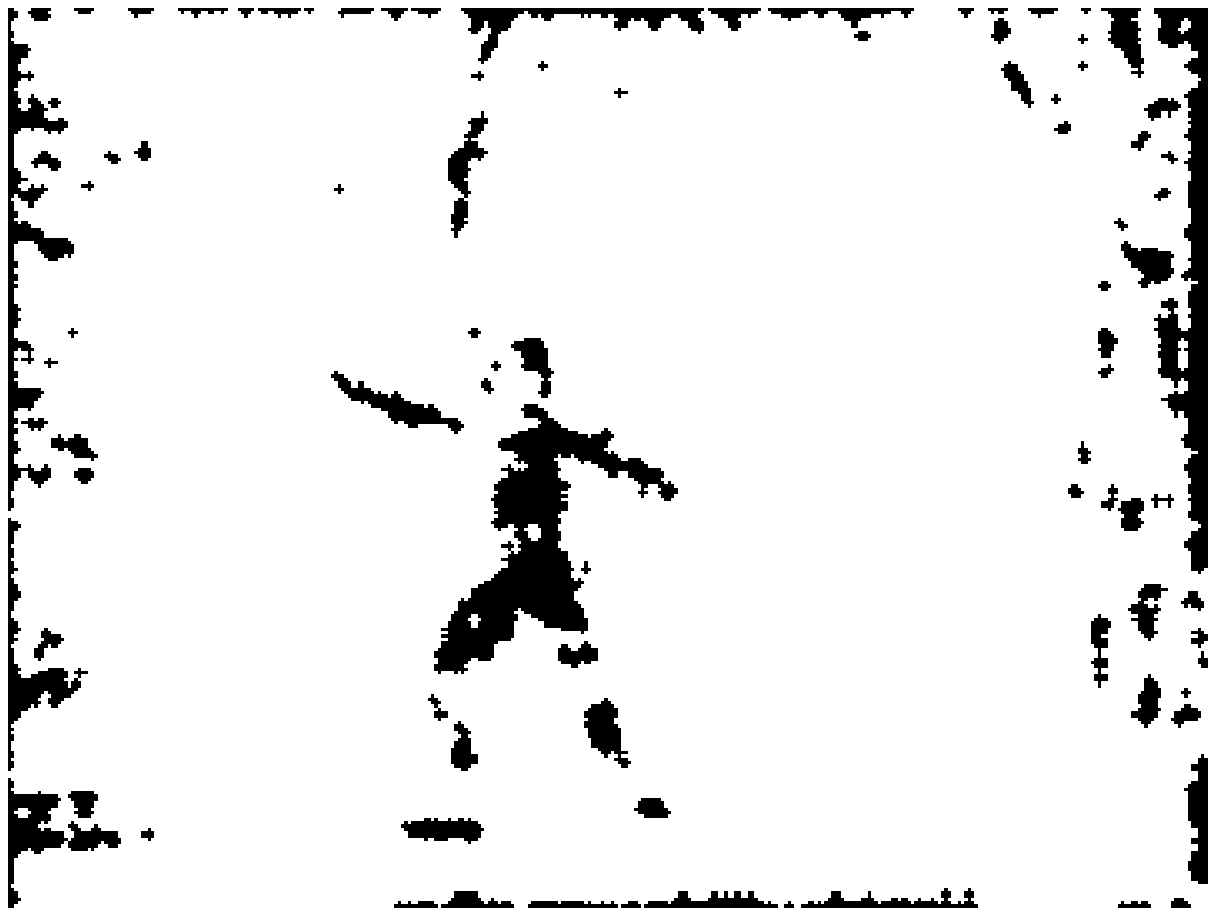} &
\epsfxsize=1in\epsfysize=0.67in\epsfbox{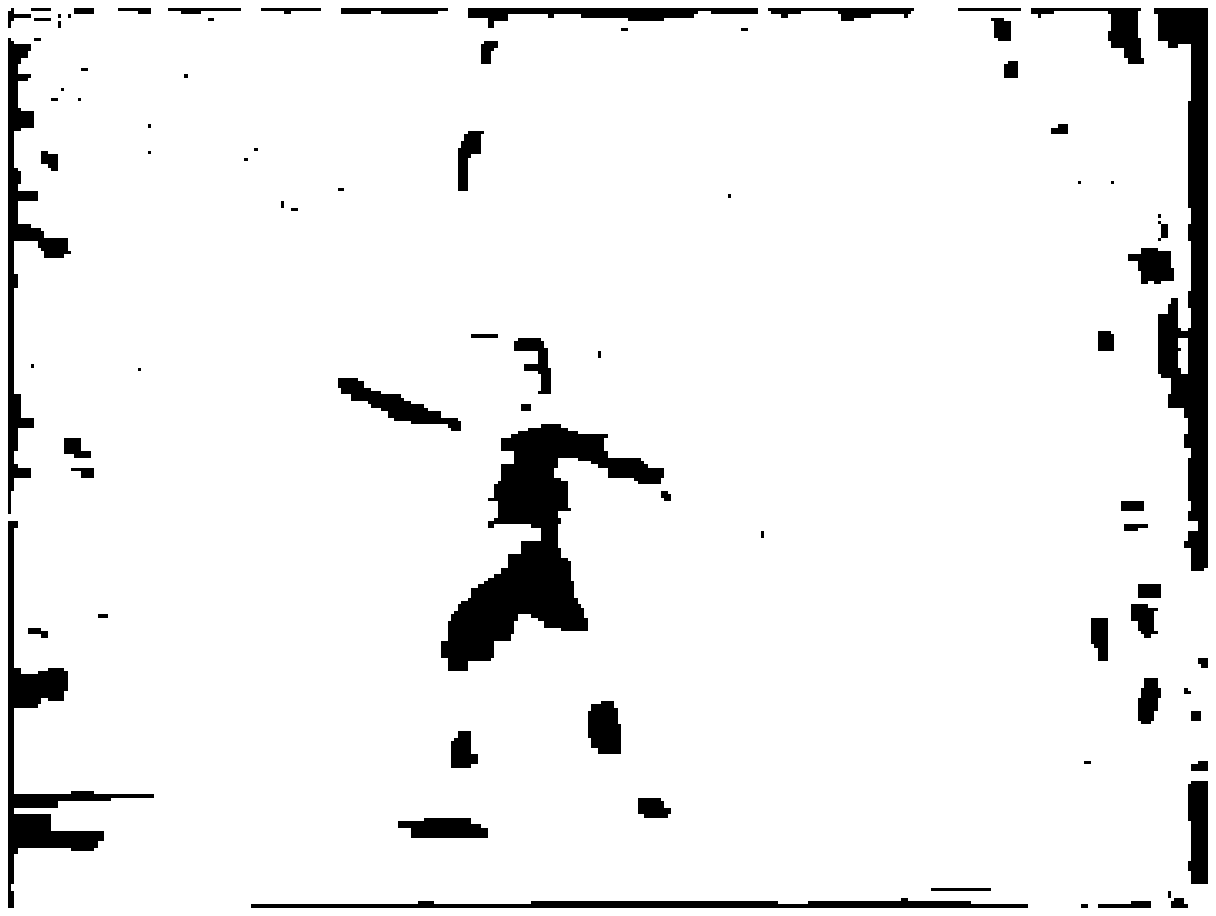} \\ \hline
\epsfxsize=1in\epsfysize=0.67in\epsfbox{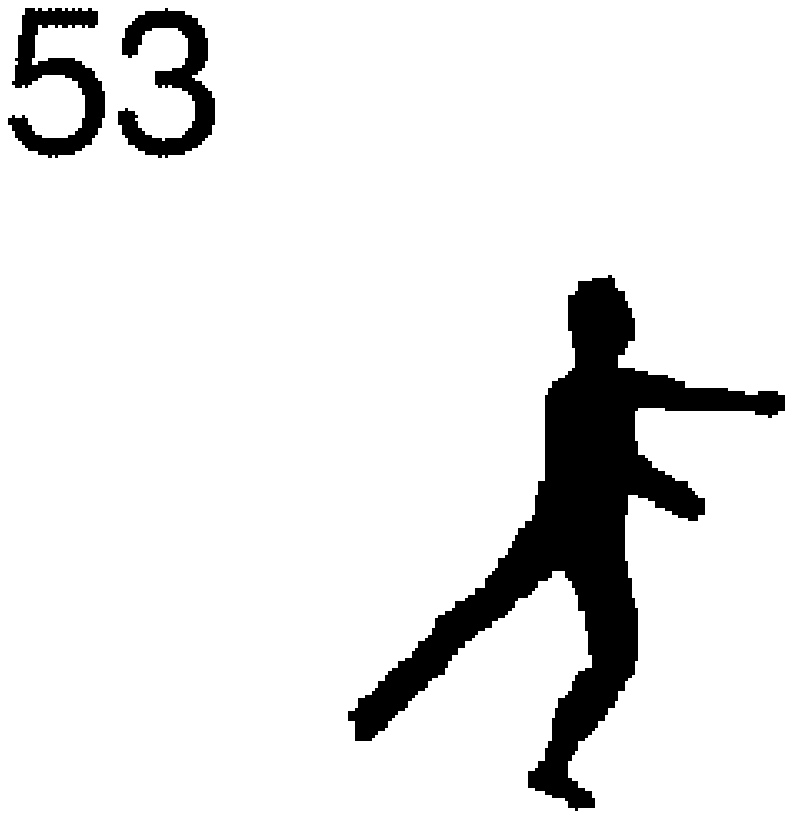} &
\epsfxsize=1in\epsfysize=0.67in\epsfbox{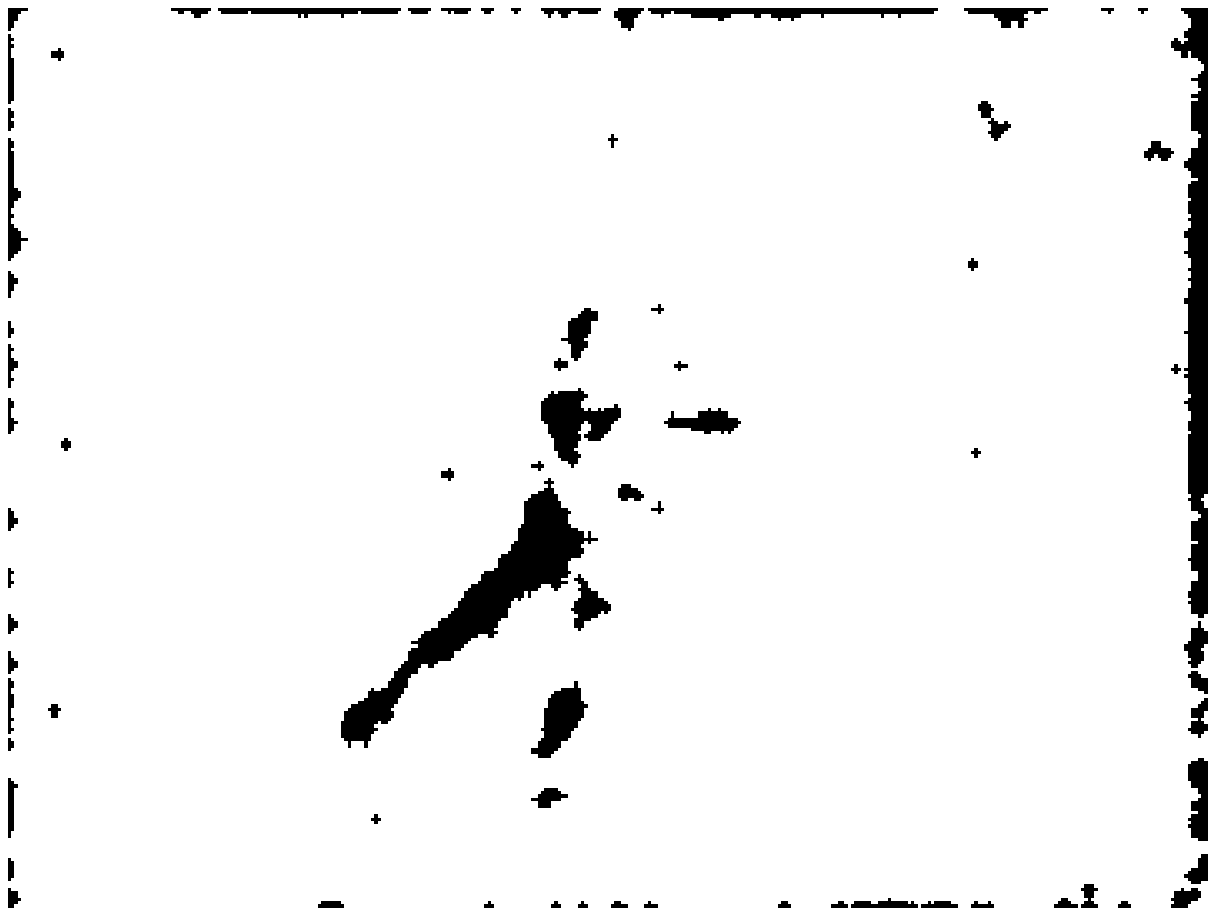} &
\epsfxsize=1in\epsfysize=0.67in\epsfbox{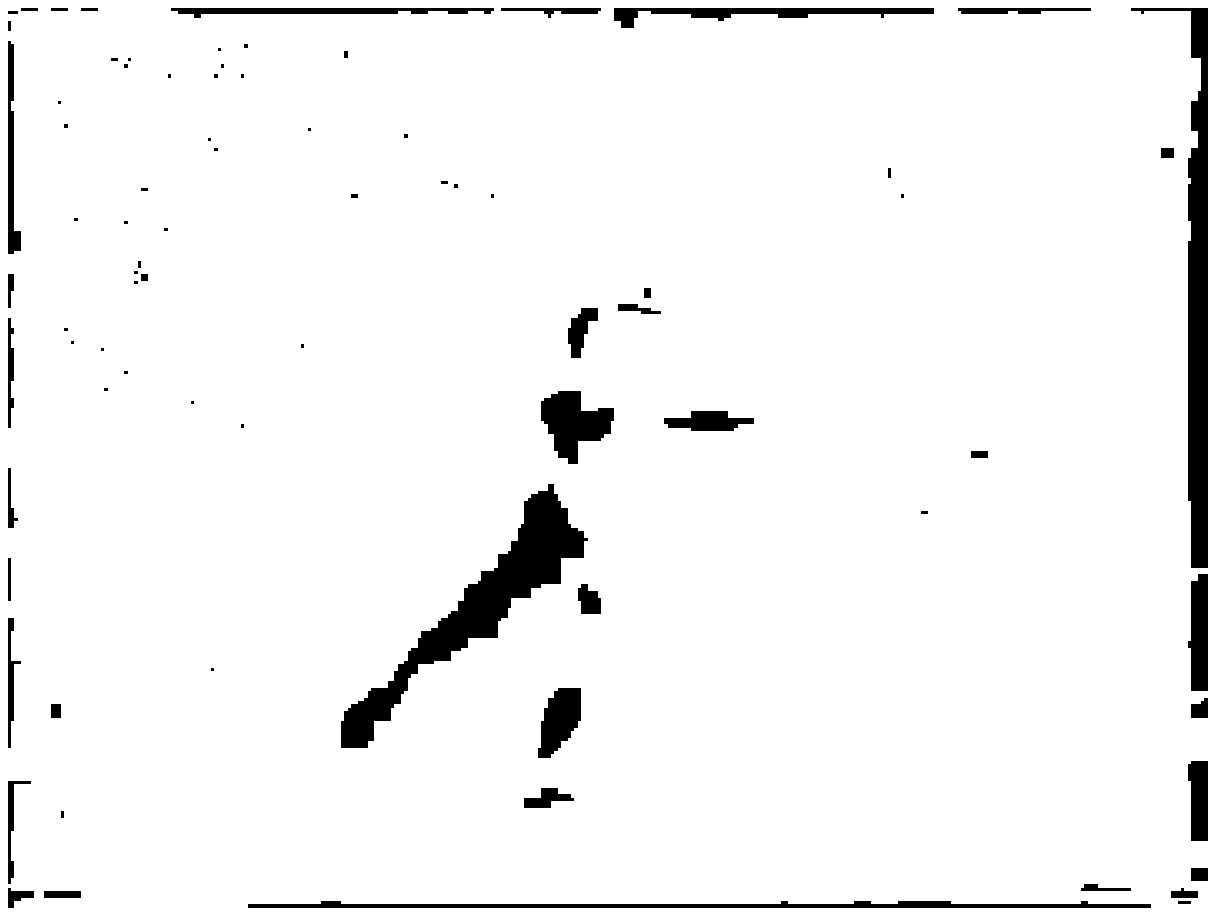} \\ \hline
\epsfxsize=1in\epsfysize=0.67in\epsfbox{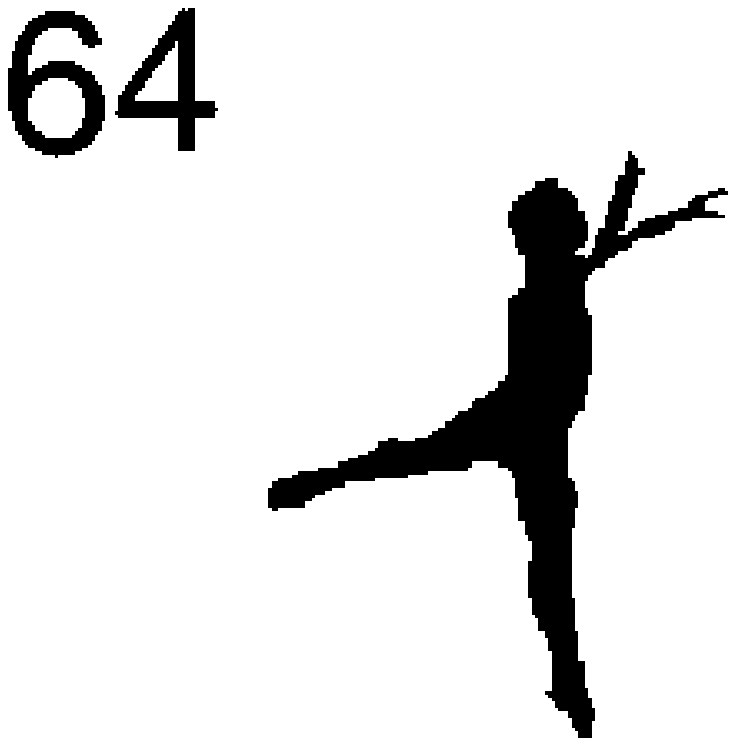} &
\epsfxsize=1in\epsfysize=0.67in\epsfbox{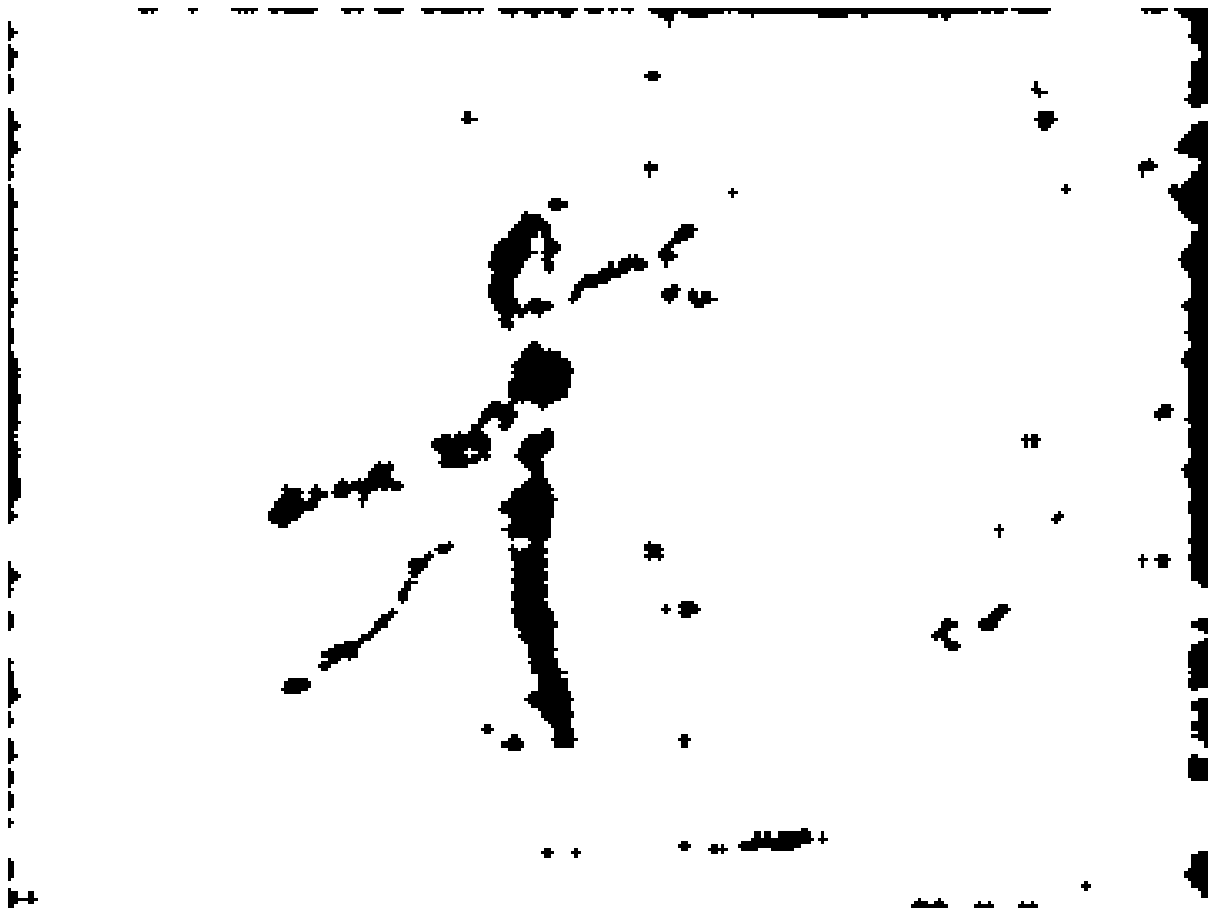} &
\epsfxsize=1in\epsfysize=0.67in\epsfbox{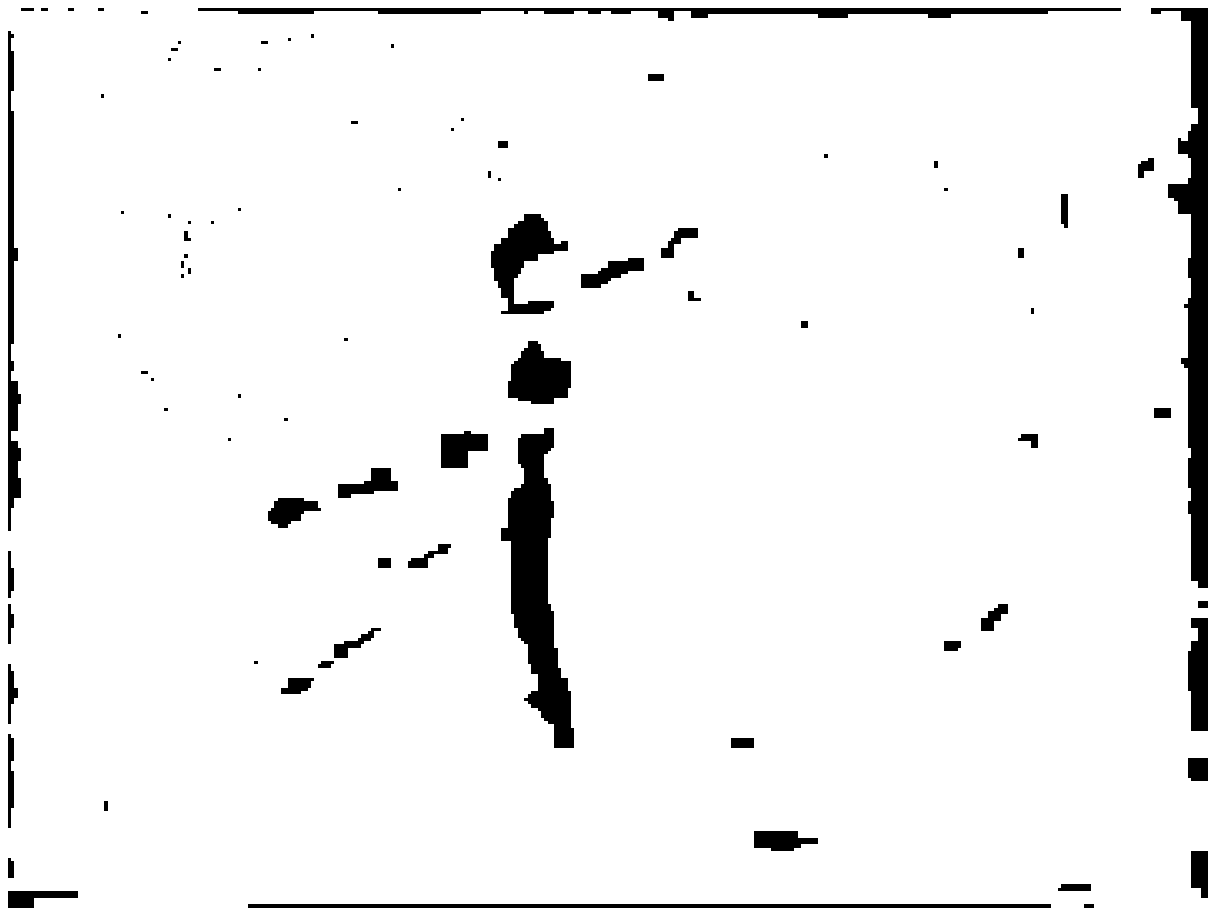} \\ \hline
\epsfxsize=1in\epsfysize=0.67in\epsfbox{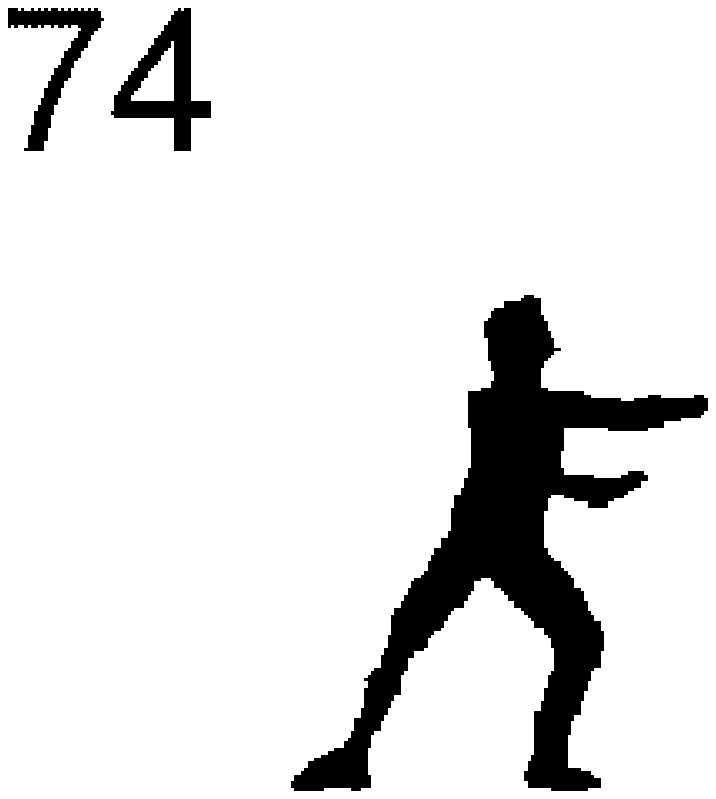} &
\epsfxsize=1in\epsfysize=0.67in\epsfbox{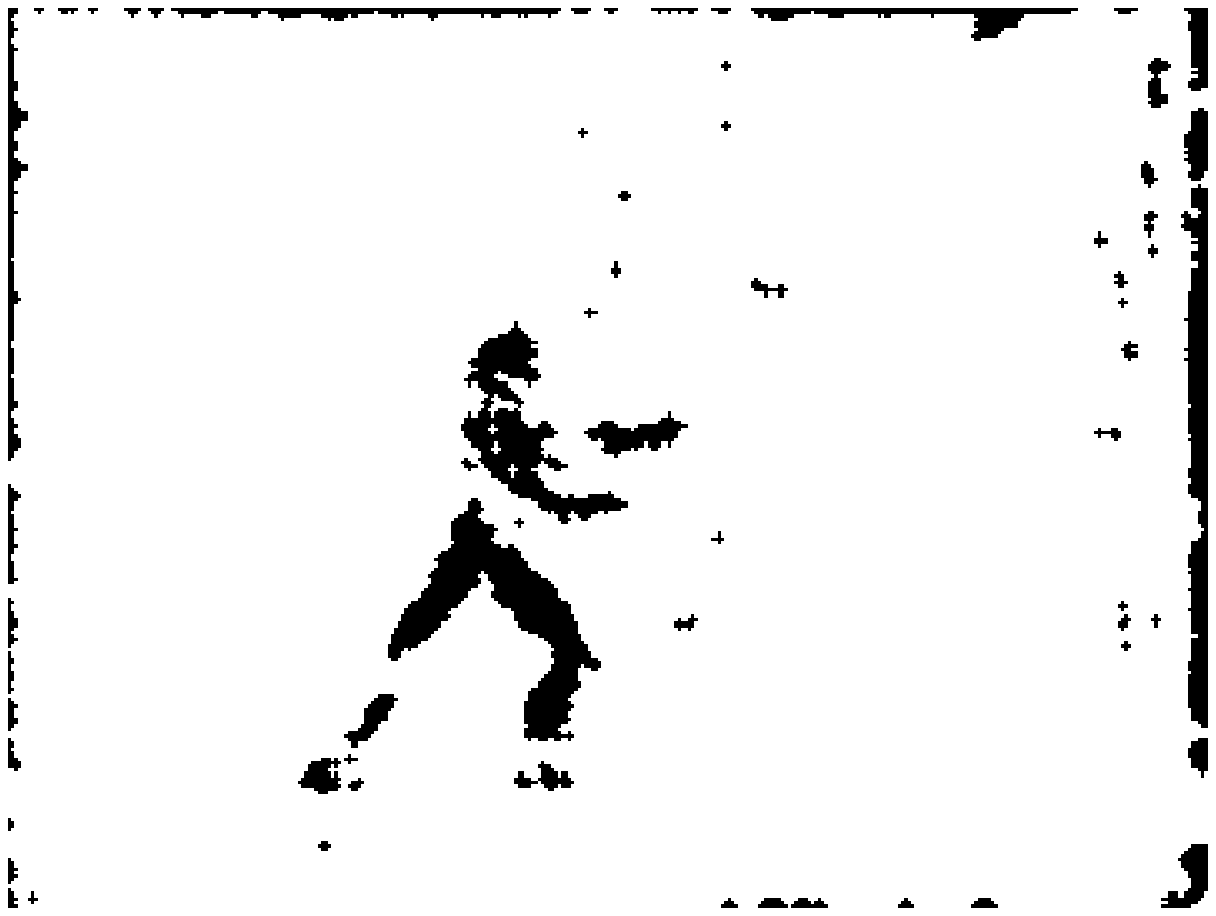} &
\epsfxsize=1in\epsfysize=0.67in\epsfbox{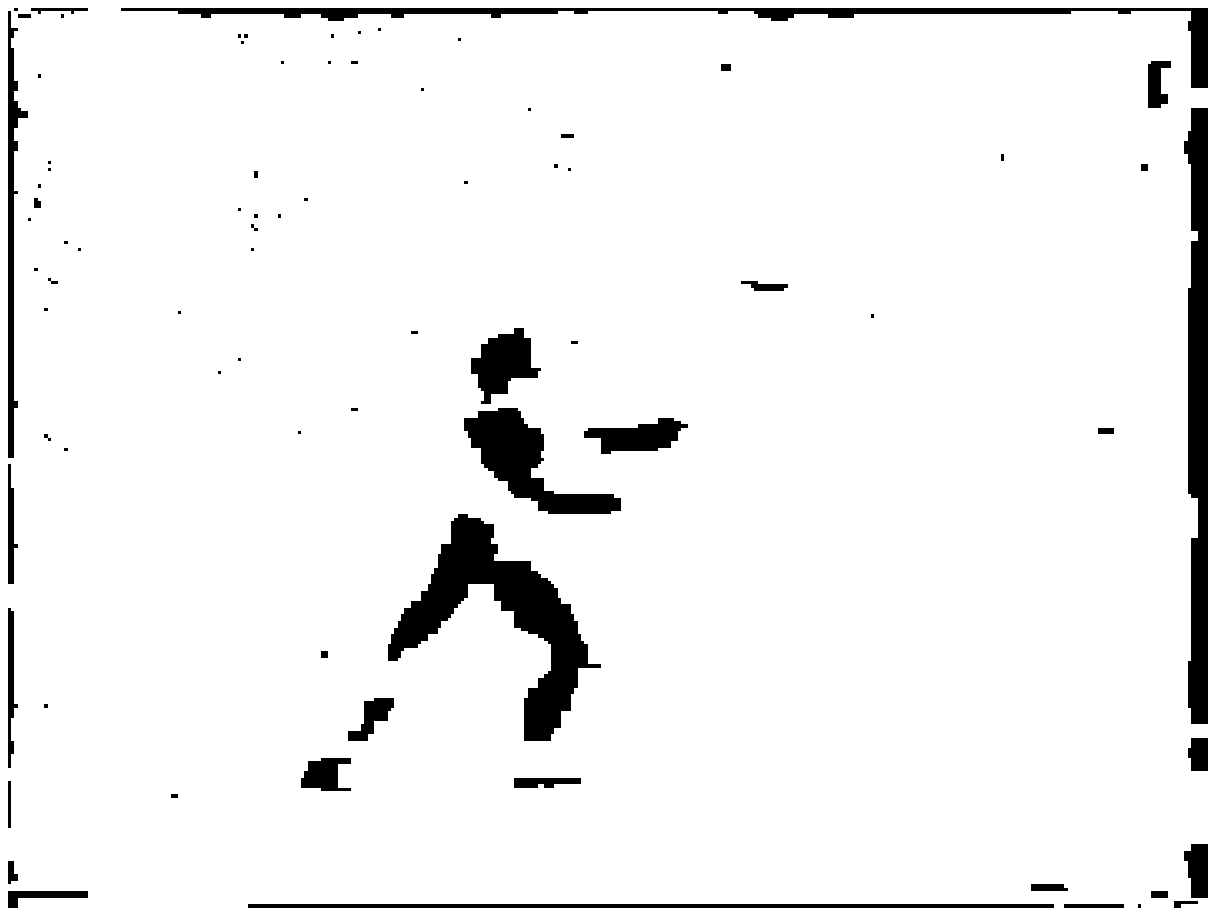} \\ \hline
\epsfxsize=1in\epsfysize=0.67in\epsfbox{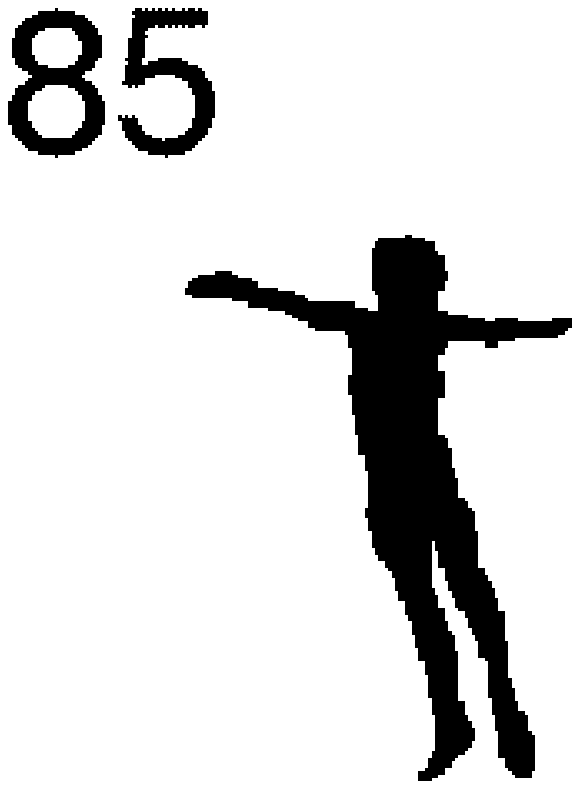} &
\epsfxsize=1in\epsfysize=0.67in\epsfbox{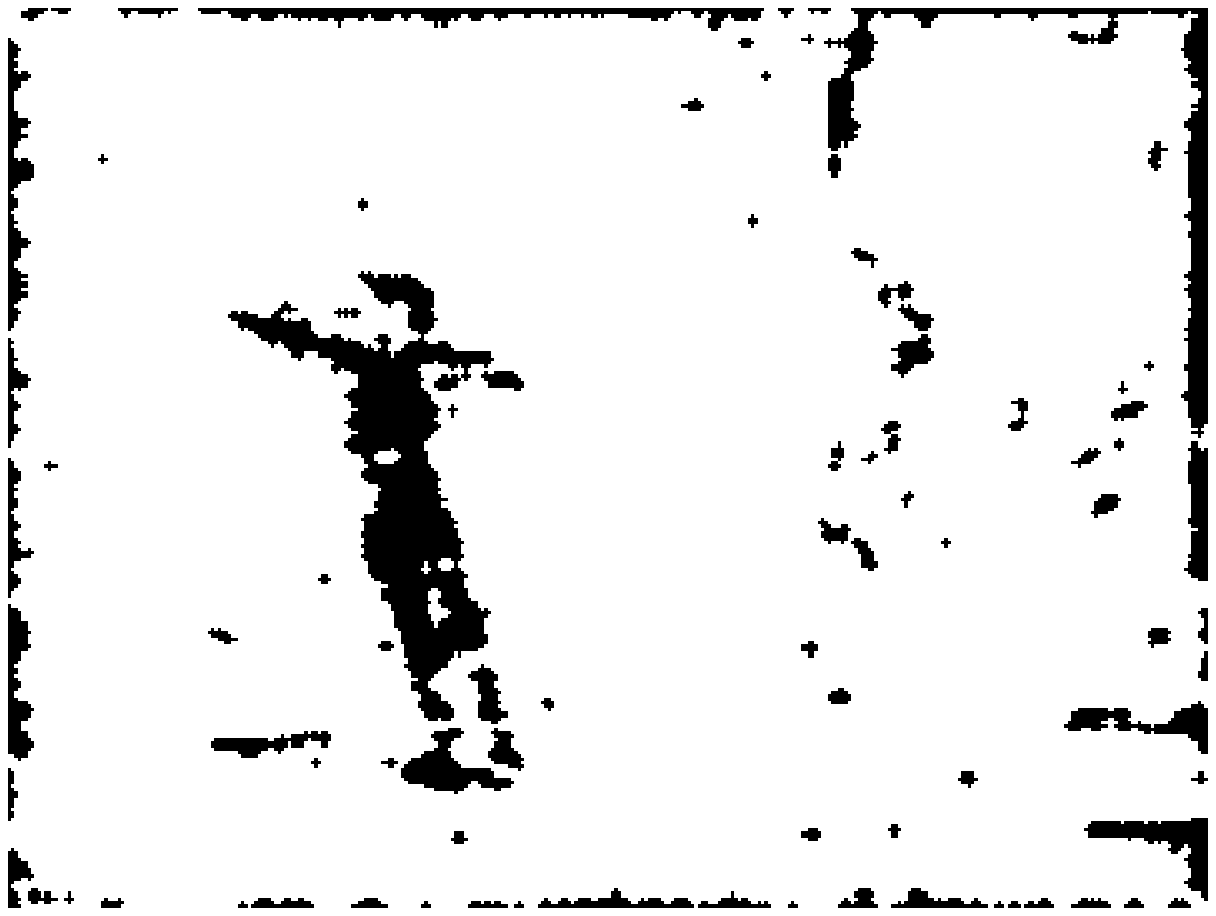} &
\epsfxsize=1in\epsfysize=0.67in\epsfbox{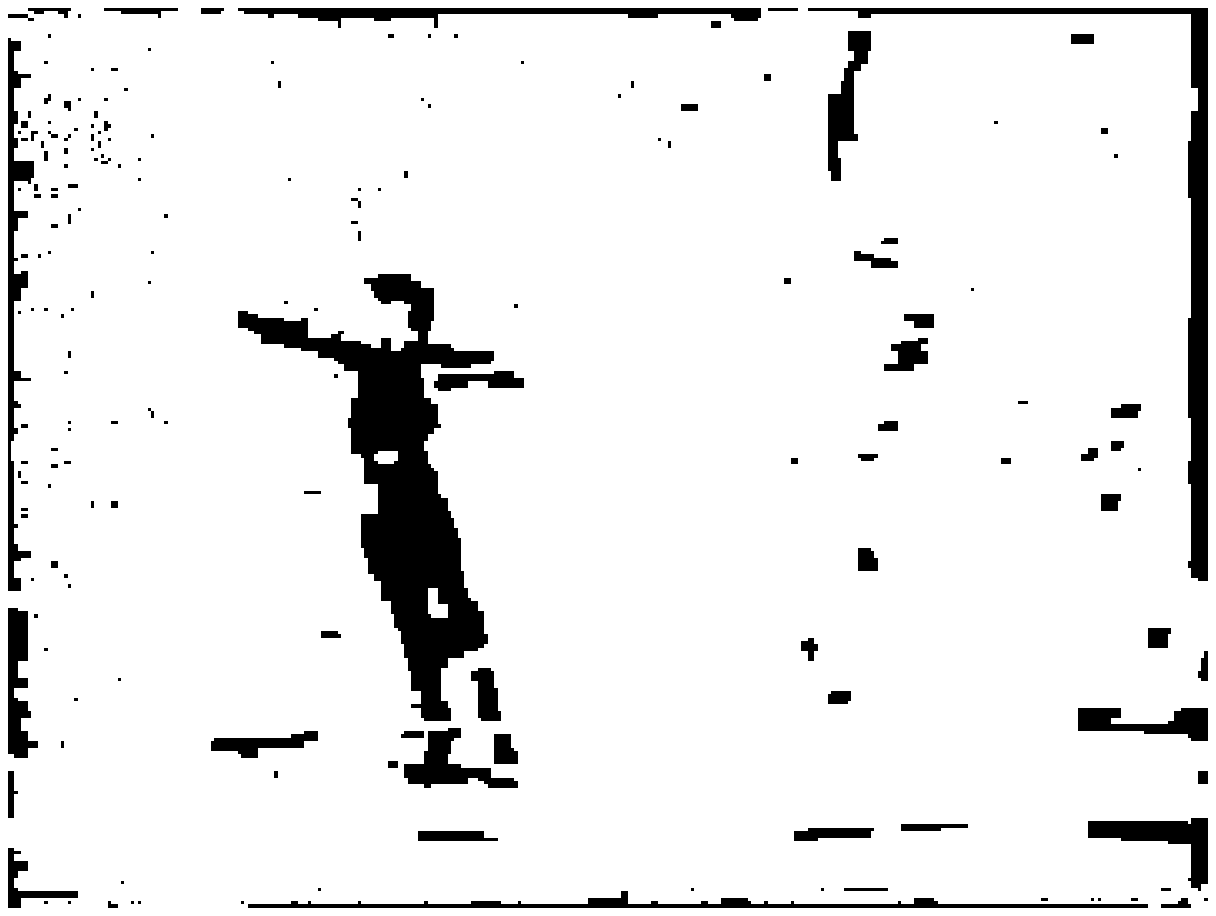} \\ \hline
\epsfxsize=1in\epsfysize=0.67in\epsfbox{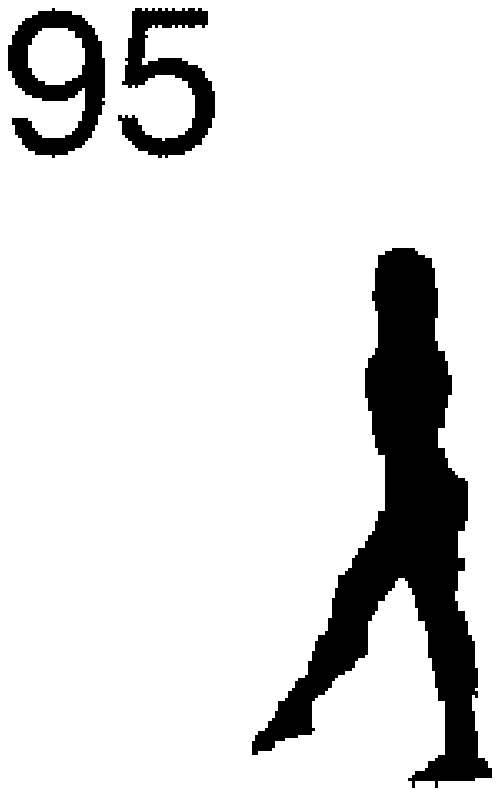} &
\epsfxsize=1in\epsfysize=0.67in\epsfbox{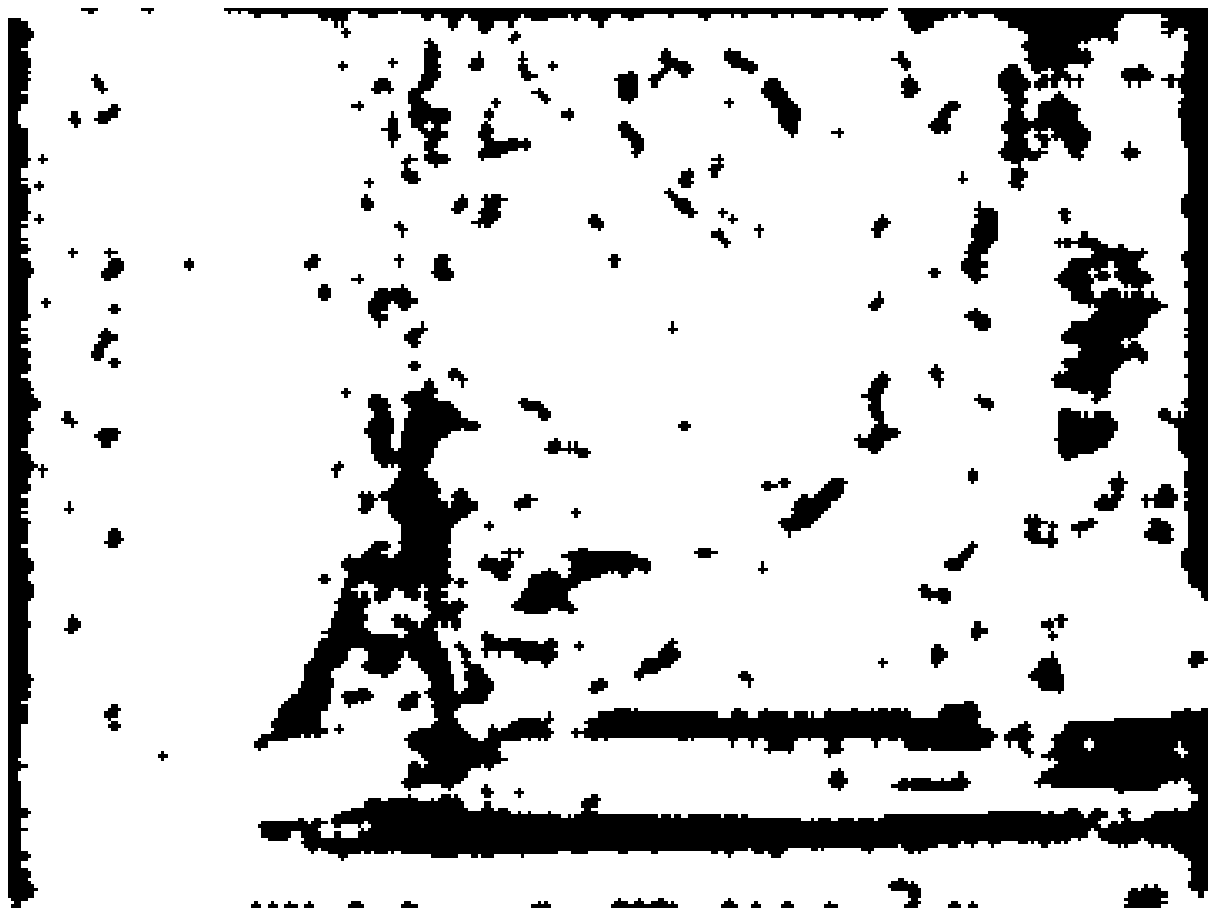} &
\epsfxsize=1in\epsfysize=0.67in\epsfbox{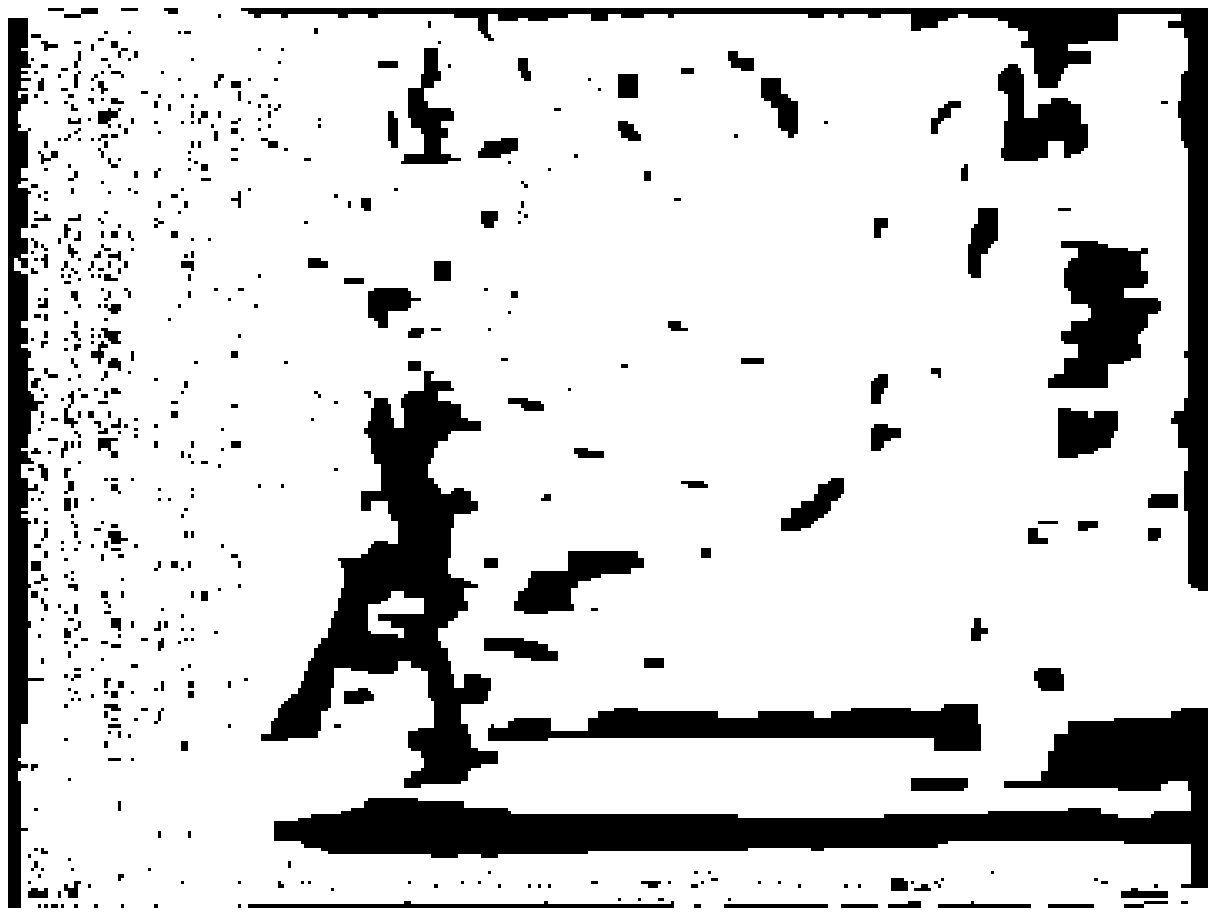} \\ \hline
(Truth) & (Morph.) & (Graph) \\ \hline
\end{tabular}
\end{center}
\caption{A sequence of frames from the {\em Dancer} clip.  The
dancer's costume blends with the background in many places.}
\label{fig-baryseq}
\end{figure}

Although the numeric differences appear small, their psychological
importance can be large, with the numbers unable to tell the entire
story.  The presence or absence of a body part such as a forearm may
alter the error by as little as 0.04 or so, while a one-pixel shift
distributed evenly around the entire the segmentation boundary can
alter the error by 0.12 or more.  To illustrate what the error values
cannot, Figures~\ref{fig-hedvigseq}--\ref{fig-baryseq} show sequences
of frames from each clip, spread evenly across time.  (Frames where
the subject is entering or exiting the screen are not shown.)

Compared side by side, the graph based result (right column) typically
appears ``cleaner'' than the morphological result, and adheres more
faithfully to the contours of the ground truth.  The {\em Indoor} clip
best displays the advantages of the graph algorithm, showing a
smoother boundary and less frequent inclusion of background.  Although
the graph algorithm does better on all three clips, {\em Outdoor} is
mostly too easy to show up the differences, and {\em Dancer} is too
hard.  The errors evident in the {\em Dancer} clip demonstrate that
any contrast-based algorithm will fail where the input is deceptive
(due to low contrast between foreground and background, heavy
shadowing, reflections, noise, etc.).  High quality input remains
crucial, regardless of the algorithm applied.

\begin{table}
\begin{center}
\caption{Summary of foreground segmentation results (connected components error criterion).}
\label{tbl-videos}
\begin{tabular}{|l|c|c|}
\hline
Clip & Morph & Graph \\
\hline
1. Outdoor&  0.164  & 0.161 \\
{\em Params:} & $\tau = 20.3, r = 2$ & $\tau = 16.2, \alpha = 0.94$ \\
\hline
2. Indoor&  0.154  & 0.133 \\
{\em Params:}& $\tau =  5.21, r = 1$  & $\tau =  4.87, \alpha = 0.81$ \\
\hline
3. Dancer&  0.541  & 0.532 \\
{\em Params:} & $\tau =  2.26, r = 1$  & $\tau =  2.15, \alpha = 0.97$ \\
\hline
\end{tabular}
\end{center}
\end{table}


\Section{Related Work and Conclusions}
\label{conc-sect}

Graph cuts can produce a cleaner foreground segmentation based upon
frame-by-frame comparisons with a background model.  This result may
not surprise those who have been following the use of graph-based
methods for other applications.  This paper combines advances from
several different research threads that have not previously been
applied to the specific problem of foreground segmentation.  Given the
status of foreground segmentation as the precursor to a host of other
applications, any advance which can improve the segmentation quality
may have wide-ranging effects.  The experimental results show that the
borrowed techniques produce excellent results, as they have in other
fields.

\SubSection{Previous Work}

Although the use of a 2-way cut for foreground segmentation is novel,
other kinds of graph-based methods have received considerable
attention recently for segmentation applications.  In particular,
methods based upon the {\em minimum normalized cut} (n-cut) have
achieved notable success in general segmentation problems
\cite{shi:ncut}.  General segmentation (with the goal of subdividing
any image into coherent regions without {\em a priori} knowledge of
its contents) is a much more difficult problem than foreground
segmentation with a static background, as explored herein.  Not
surprisingly therefore, n-cut algorithms for general segmentation
differ from this paper's approach, typically employing a
fully-connected graph requiring approximation methods to solve
\cite{shi:ncut98}.  By contrast, the algorithm described herein uses a
graph that represents only local connections among pixels.  Because
the number of links remains linear in the number of nodes, this
local-only graph can be solved easily and exactly.  Furthermore, the
advantages of the normalized cut over the standard cut do not apply in
the local-only case, because the energy of a cut relates only
indirectly to the number of nodes on either side.

A more direct connection to the current work may be drawn from
research on image correspondance for stereo vision and visual
correspondence \cite{olga:motion,olga:stereo}.  Again, the stereopsis
problem is more difficult than foreground segmentation, requiring a
selection among multiple hypothesized displacements at each pixel.
The work mentioned above therefore employs algorithms to approximate a
multiway cut on a graph representing the image.  The algorithm herein
embodies a special case of such a situation, where the existence of
only two categories to distinguish (foreground and background) allows
the use of an exact 2-way cut solution.

In the processing of the video frames prior to the thresholding step,
this work follows current the state of the art.  In particular, it
builds probability models for the distribution of measured color
values at each pixel \cite{karmann:1990}.  It further employs
techniques to eliminate interference from shadows; similar measures
were recently described elsewhere \cite{horprasert:shadow}.  The
authors of the latter work note that their segmentation process runs
in real time, which is a challenge for the new algorithm due to the
graph cut step.  Theoretical bounds on this step are $O(n^2 \log n)$,
although for some problem classes the actual performance can be
quadratic or better \cite{goldberg:mincut}.  Empirically, it appears
that real time processing is still possible at lowered resolution, and
this stricture should ease with time as processor speeds increase.

\SubSection{Final Thoughts}

Using graph cuts for foreground segmentation produces cleaner and more
accurate results than the currently prevailing approach based upon
morphological operations.  The graph-based technique appears better at
overcoming the effects of noise by aggregating information from a
local neighborhood around each pixel, while remaining true to the
underlying data.  On test using synthetic data, it cut the error rate
by at least a third over the current methods for the noisiest input,
and by a greater factor for the less noisy cases.  On real data, the
method significantly reduces error over current methods, although it
cannot magically cure problems associated with low quality input data.
The one disadvantage of the graph-based method is its speed;
empirically it runs more slowly or at lower resolution than the
morphological operations.

Given the range of applications that use background subtraction,
adoption of the new technique seems likely to provide significant
benefits in a number of areas.  In addition to current uses for
background subtraction, the higher fidelity of the graph-cut method
may open up new applications not hitherto feasible because they
require highly reliable input.  In any case, graph cuts for foreground
segmentation deserve a place in the research scientist's bag of tools.

\bibliographystyle{plain}
{\small\bibliography{bgsub-arxiv}}

\end{document}